\pdfoutput=1
\documentclass{article}

\PassOptionsToPackage{numbers,compress,sort}{natbib}
\usepackage{natbib}

\usepackage{wrapfig}
\usepackage[utf8]{inputenc} %
\usepackage[T1]{fontenc}    %
\usepackage{url}            %
\usepackage{booktabs}       %
\usepackage{longtable}
\usepackage{amsfonts}       %
\usepackage{nicefrac}       %
\usepackage{xspace}         %
\usepackage[
  protrusion=true,
  expansion=true,
  final,
  factor=1100,
  stretch=20,
  shrink=20
]{microtype}
\usepackage{amsmath,amssymb}
\usepackage{graphicx}
\usepackage[rightcaption,raggedright]{sidecap}
\usepackage{subcaption}
\usepackage{multirow}
\usepackage{todonotes}
\usepackage{float}
\usepackage{adjustbox}
\usepackage{placeins}
\usepackage{pdflscape}
\usepackage{titletoc}      %

\usepackage[a4paper,margin=1in]{geometry}

\usepackage[table]{xcolor}  %

\definecolor{medalGold}{HTML}{FFE89C}    %
\definecolor{medalSilver}{HTML}{FFD2A6}  %
\definecolor{medalBronze}{HTML}{E8C49C}  %

\definecolor{gold}{RGB}{218,165,32}
\definecolor{silver}{RGB}{160,160,160}
\definecolor{bronze}{RGB}{205,127,50}

\definecolor{priorlight}{RGB}{247,248,250} %
\definecolor{priorAccent}{HTML}{4A3FB7}    %

\usepackage[colorlinks=true,linkcolor=black,urlcolor=blue]{hyperref}
\usepackage{cleveref}
\usepackage[most]{tcolorbox}

\tcbset{
  colback=priorlight,
  colframe=black!15,
  boxrule=0.5pt,
  arc=6pt,
  left=16pt,right=16pt,top=14pt,bottom=14pt,
  enhanced,
}

\newcommand{\field}[2]{%
  \par\noindent
  {\small\color{black!55}\textsc{#1}}\quad #2\par\addvspace{2pt}}

\usepackage{tikz}
\usetikzlibrary{
  arrows.meta,
  positioning,
  fit,
  shapes.geometric,
  shapes.symbols,
  shapes.misc,
  shapes.multipart,
  matrix,
  backgrounds,
  calc,
  decorations.pathreplacing,
  decorations.markings
}

\pgfdeclarelayer{midground}
\pgfsetlayers{background,midground,main}

\colorlet{groupCyan}{cyan!15}
\colorlet{groupGreen}{green!15}
\colorlet{groupOrange}{orange!15}
\colorlet{groupPurple}{violet!12}
\definecolor{darkgreen}{RGB}{0,100,0}

\definecolor{groupCyanEdge}{RGB}{ 30, 110, 135}
\definecolor{groupGreenEdge}{RGB}{ 60, 120,  55}
\definecolor{groupOrangeEdge}{RGB}{180,  95,  20}
\definecolor{groupPurpleEdge}{RGB}{ 95,  60, 140}
\definecolor{groupGray}{RGB}{238, 238, 240}
\definecolor{groupGrayEdge}{RGB}{ 90,  90,  95}

\usepackage{caption}
\newcommand{\ourmodel}{\mbox{TabPFN-3.5}\xspace}
\newcommand{\ourmodelfast}{\mbox{TabPFN-3.5-Fast}\xspace}
\newcommand{\ourmodelplus}{\mbox{TabPFN-3.5-Plus}\xspace}
\newcommand{\ourmodelthinking}{\mbox{TabPFN-3.5-Thinking}\xspace}
\newcommand{\ourmodelthree}{\mbox{TabPFN-3}\xspace}
\newcommand{\ourmodelthreeplus}{\mbox{TabPFN-3-Plus}\xspace}

\newcommand{\RelArena}{RelArena-$\alpha$\xspace}

\begin{document}

\vspace*{-1cm}
\begin{tcolorbox}

\vspace{-0.35em}\hspace*{-0.2cm}\includegraphics[width=0.2\linewidth]{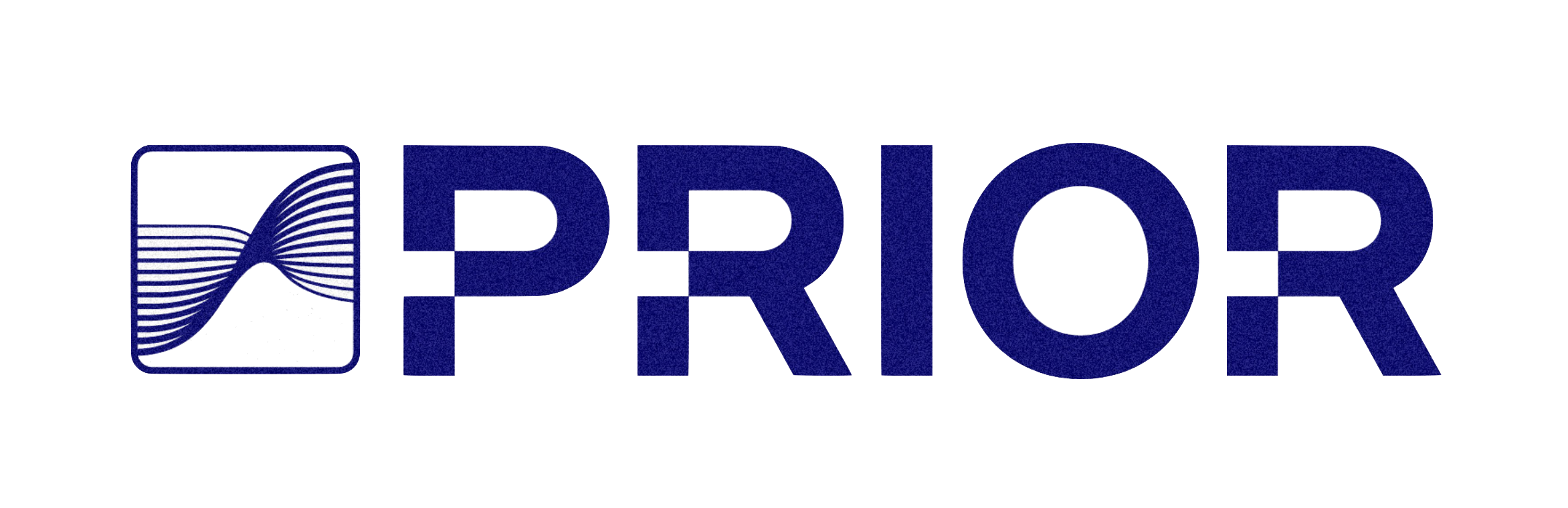}

{\LARGE\bfseries TabPFN-3.5: Technical Report\par}

\vspace{0.35em}
{\large \hyperref[app:contributors]{Prior Labs Team}} (see Appendix \ref{app:contributors} for the list of contributors)

\vspace{1.5em}
We introduce \textbf{\ourmodel{}}, our new flagship Tabular Foundation Model. It significantly outperforms its predecessor, \ourmodelthree{} \citep{grinsztajn2026tabpfn3technicalreport}, and all existing baselines across a broad range of tabular problems. \ourmodel{} sets a new state of the art on standard tabular prediction in TabArena \citep{erickson2025tabarena}, and extends it to the data practitioners encounter in practice: non-i.i.d.\ data with temporal or grouped splits \citep{purucker2026beyond}, tables with strings \citep{blayer2026strable}, text and images \citep{arazi2026multabenchbenchmarkingmultimodaltabular}, high-cardinality categorical features, and wide tables with many features \citep{purucker2026beyond}. These gains carry over to our task-specific harnesses: state of the art on relational data \citep{hayler2026relarena} and stronger time-series forecasting \citep{shchur2025fev}. For faster inference, our variant  \textbf{\ourmodelfast{}} runs up to $3\times$ faster than \ourmodelthree{} while keeping most of the accuracy gains. In addition, we upgrade \textbf{\ourmodelplus{}}, expanding our multimodal capabilities with advanced text and date handling alongside proprietary inference optimizations. Finally, we release a new version of our Thinking mode, \textbf{\ourmodelthinking{}}, which scales inference-time computation to push the state of the art further. It benefits from our stronger base model and from inference-time improvements that make it up to $12\times$ faster than \ourmodelthree{}-Thinking.

\vspace{.5em}

\vspace{0.5em}

\field{Date}{September 14th 2026}
\field{License}{Code: Apache 2.0; weights: TABPFN-3.5 License v1.0 (see \Cref{sec:license})}
\field{Docs}{\url{https://docs.priorlabs.ai}}

\end{tcolorbox}

\vspace{0.5cm}

\begin{table}[H]
  \centering
  \newsavebox{\overviewtablebox}
  \begin{lrbox}{\overviewtablebox}\small%
  \begin{tabular}{ll@{ of }rrr}
    \toprule
    \multicolumn{5}{c}{\textbf{\ourmodel{}: state of the art on all 7 benchmarks}} \\
    \midrule
    \multicolumn{3}{c}{} & \multicolumn{2}{c}{\textbf{Win rate against best}} \\
    \cmidrule(lr){4-5}
    \textbf{Benchmark} & \multicolumn{2}{c}{\textbf{Rank}} & \multicolumn{1}{c}{\textbf{other TFM}} & \multicolumn{1}{c}{\textbf{non-TFM}} \\
    \midrule
    TabArena & \textcolor{gold}{\textbf{1}} & 89 & \cellcolor[HTML]{F1F9C9}\makebox[4.4em][l]{\scriptsize\color{gray}TabFM+}57\% & \cellcolor[HTML]{78B696}\makebox[4.4em][l]{\scriptsize\color{gray}RealMLP}96\% \\
    BeyondArena & \textcolor{gold}{\textbf{1}} & 29 & \cellcolor[HTML]{9ED4A6}\makebox[4.4em][l]{\scriptsize\color{gray}TabICLv2}83\% & \cellcolor[HTML]{AFDDAA}\makebox[4.4em][l]{\scriptsize\color{gray}RealMLP}79\% \\
    STRABLE & \textcolor{gold}{\textbf{1}} & 197 & \cellcolor[HTML]{76B294}\makebox[4.4em][l]{\scriptsize\color{gray}TabICLv2}98\% & \cellcolor[HTML]{75B093}\makebox[4.4em][l]{\scriptsize\color{gray}XGBoost}98\% \\
    MulTaBench & \textcolor{gold}{\textbf{1}} & 14 & \cellcolor[HTML]{81C69E}\makebox[4.4em][l]{\scriptsize\color{gray}TabICLv2}90\% & \cellcolor[HTML]{7BBA99}\makebox[4.4em][l]{\scriptsize\color{gray}TabM}95\% \\
    \RelArena{} & \textcolor{gold}{\textbf{1}} & 12 & \cellcolor[HTML]{F0F9C7}\makebox[4.4em][l]{\scriptsize\color{gray}RT-PluRel}57\% & \cellcolor[HTML]{C9E8AD}\makebox[4.4em][l]{\scriptsize\color{gray}KurveRSC}71\% \\
    TALENT & \textcolor{gold}{\textbf{1}} & 37 & \cellcolor[HTML]{A9DAA9}\makebox[4.4em][l]{\scriptsize\color{gray}TabICLv2}80\% & \cellcolor[HTML]{80C49D}\makebox[4.4em][l]{\scriptsize\color{gray}CatBoost}91\% \\
    ScoringBench & \textcolor{gold}{\textbf{1}} & 52 & \cellcolor[HTML]{B2DEAA}\makebox[4.4em][l]{\scriptsize\color{gray}EXAONE}78\% & \cellcolor[HTML]{7FC39D}\makebox[4.4em][l]{\scriptsize\color{gray}RealMLP}91\% \\
    \midrule
    Mean               & \multicolumn{2}{l}{\textcolor{gold}{\textbf{1}}} & \cellcolor[HTML]{B3DEAA}78\% & \cellcolor[HTML]{85C8A0}89\% \\
    \bottomrule
  \end{tabular}%
  \end{lrbox}
  \resizebox{\dimexpr\wd\overviewtablebox+(\linewidth-\wd\overviewtablebox)/3\relax}{!}{\usebox{\overviewtablebox}}
  \caption{\textbf{\ourmodel{} against the strongest non-TabPFN competitor on each benchmark.} Rank is the position of the best \ourmodel{} member (\ourmodelthinking{} on the first four benchmarks, TabPFN-Rel (3.5) on \RelArena{}, \ourmodel{} on TALENT and ScoringBench) among the methods on the benchmark's leaderboard. Win rates are the share of splits won against the best other Tabular Foundation Model (TFM) and the best non-foundation model, datasets weighted equally. Details in \Cref{sec:app:family_winrates}.}
  \label{tab:family_winrates_tiny_no_hindsight_named}
\end{table}

\newpage
{
  \setcounter{tocdepth}{2}
  \tableofcontents
}
\newpage

\section{Introduction}
\label{sec:introduction}

Tabular Foundation Models (TFMs) have replaced gradient-boosted trees, for decades the default choice for tabular prediction that drives decision making across science and industry \citep{henry2015targeted,khandani2010consumer,carvalho2019systematic,baldi2014searching,shwartz2022tabular,grinsztajn2022tree}, as the strongest predictors on standard i.i.d.\ benchmarks: on TabArena~\citep{erickson2025tabarena}, TFMs dominate the leaderboard, and the strongest default TFM outperforms AutoML systems that tune and ensemble models for hours \citep{erickson2025tabarena,autogluon_tabular}. Successive TabPFN releases~\citep{hollmann2022tabpfnv1,Hollmann2025tabpfnv2,TabPFN-2.5,grinsztajn2026tabpfn3technicalreport} established and extended this paradigm, from a thousand clean numerical rows \citep{hollmann2022tabpfnv1} to a million rows with heterogeneous features (categorical and numerical), missing values, and many-class targets \citep{Hollmann2025tabpfnv2,TabPFN-2.5,grinsztajn2026tabpfn3technicalreport}. Around these models, an active ecosystem has grown across time-series forecasting \citep{hoo2024tabpfn_ts,hoo2025tables_to_time}, causal inference \citep{robertson_dopfn,balazadeh_causalpfn,feuerriegel_causalfm}, graph learning \citep{Hayler2025GraphsTablesZeroShot,eremeev2025graphpfnpriordatafittedgraph}, relational data \citep{hayler2026relarena}, interpretability \citep{rundel2024interpretable,olsen2026computingconditionalshapleyvalues}, survival analysis \citep{qi2026survivalpfn,seletkov2026survivalincontextpriorfittedincontext}, Bayesian optimization \citep{Yu2025GITBO,rogers2026zeroshotbo}, and reinforcement learning \citep{Schiff2025TabPFNRL}, with hundreds of published applications and millions of downloads; the \ourmodelthree{} report \citep{grinsztajn2026tabpfn3technicalreport} surveys both. The field is maturing: within the last year, TabICLv2~\citep{qu2026tabiclv2}, TabFM \citep{tabfm2026}, EXAONE-Tabular \citep{eo2026exaonetabular}, TabDPT-Turbo \citep{hosseinzadeh2026tabdpt}, 
and SAP-RPT-1~\citep{spinaci2026contexttab} have been released as competitive TFMs, and benchmarks have broadened from numerical i.i.d.\ data to non-i.i.d.\ \citep{rubachev_tabred,purucker2026beyond}, string-valued \citep{blayer2026strable}, multimodal~\citep{arazi2026multabenchbenchmarkingmultimodaltabular}, and relational \citep{hayler2026relarena} data, and to calibrated predictive distributions \citep{landsgesell2026scoringbenchbenchmarkevaluatingtabular}. These new benchmarks also show where TFMs still fall short. \citet{purucker2026beyond} show that TFMs excel on clean i.i.d.\ data, yet even the best of them fall behind tuned conventional models on grouped or temporal splits, on tables with many rows or many features, and on tables with text or high-cardinality features. This is the data practitioners face every day. And the TFMs that have pushed accuracy furthest have also grown larger and slower, which makes them harder to deploy.

To address these open problems, we introduce \ourmodel{}. It extends the lead of TFMs from clean i.i.d.\ benchmarks to all of these regimes, on which it is now competitive with heavily tuned gradient-boosted trees and MLPs. We observe these gains across benchmarks: \ourmodel{} sets a new state of the art on TabArena \citep{erickson2025tabarena}, BeyondArena \citep{purucker2026beyond}, STRABLE \citep{blayer2026strable} for tabular data with strings, MulTaBench \citep{arazi2026multabenchbenchmarkingmultimodaltabular} for multimodal tabular learning with text and images, ScoringBench \citep{landsgesell2026scoringbenchbenchmarkevaluatingtabular} for predictive distributions, and, through TabPFN-Rel, RelArena \citep{hayler2026relarena} for relational data. At the same time, it retains what made \ourmodelthree{} practical: training sets of up to one million rows, many-class classification, calibrated predictive distributions from a single forward pass, and cached inference in which the training rows are processed once and test rows are then scored in well under a second. \ourmodel{}'s gains come from revised per-cell encodings, doubling the model width while keeping the cache size unchanged, a single checkpoint trained jointly for classification and regression, and a synthetic prior tuned towards high-cardinality, wide, and grouped data. \Cref{sec:experimental_results} quantifies these gains.

We release two checkpoints compatible with the open-source \href{https://github.com/PriorLabs/TabPFN}{\texttt{tabpfn}} package: \ourmodel{} as the core model and \ourmodelfast{}, an alpha checkpoint for latency-sensitive applications that runs up to $6\times$ faster than \ourmodel{} (and up to $140\times$ faster than TabFM)
with a small drop in accuracy while still matching the previous state of the art on these benchmarks. 
The core model was designed to serve as the foundation of \ourmodelplus{}, which expands our multimodal capabilities for text and date processing alongside proprietary inference optimizations, as well as \ourmodelthinking{}, which also scales inference-time computation to deliver peak performance. \ourmodelthinking{} is our most accurate model across all use cases and further strengthens grouped and temporal modeling. \ourmodelplus{}, \ourmodelfast{}, and \ourmodelthinking{} are available through the \href{https://platform.priorlabs.ai}{Prior Labs API}.

The remainder of this report presents the experimental results (\Cref{sec:experimental_results}); describes the architecture, preprocessing, inference characteristics, and synthetic prior of \ourmodel{}, as well as \ourmodelplus{} and its Thinking mode (\Cref{sec:tabpfn35}); and details licensing (\Cref{sec:license}). For installation and usage, see \href{https://docs.priorlabs.ai/}{our documentation}.

\begin{figure}[!t]
  \centering

  \begin{subtable}[b]{0.47\textwidth}
    \centering
    \small
    \begin{tabular}{lrrrr}
      \toprule
      \multirow{2}{*}{Model} & \multirow{2}{*}{Rows} & \multirow{2}{*}{Features}
        & \multicolumn{2}{c}{Parameters} \\
      \cmidrule(lr){4-5}
      & & & Clf. & Reg. \\
      \midrule
      TabPFN-v1  & $1{,}000$       & $100$      & $26$\,M & ---       \\
      TabPFN-v2  & $10{,}000$      & $500$      & $7$\,M  & $11$\,M \\
      TabPFN-2.5 & $100{,}000$     & $2{,}000$  & $11$\,M & $10$\,M \\
      TabPFN-2.6 & $100{,}000$     & $2{,}000$  & $11$\,M & $13$\,M \\
      TabPFN-3 & $1{,}000{,}000$ & $2{,}000$  & $53$\,M  & $58$\,M \\
     \midrule
      \ourmodel{} & $1{,}000{,}000$ & $6{,}000$  & \multicolumn{2}{c}{$220$\,M} \\
      \ourmodelfast{} & $1{,}000{,}000$ & $6{,}000$  & \multicolumn{2}{c}{$84$\,M} \\
    \bottomrule
    \end{tabular}
    \vspace{1.8cm}
    \caption{\textbf{Overview of TabPFN releases}, with the largest recommended row and feature counts, and the models' parameter counts for each release. These are recommended limits for strong performance, not computational limits. Increasing the number of estimators allows support for up to 20K features.%
    }
    \label{tab:tabpfn-variants}
  \end{subtable}
  \hfill
  \begin{subfigure}[b]{0.47\textwidth}
    \centering
    \includegraphics[width=0.8\linewidth]{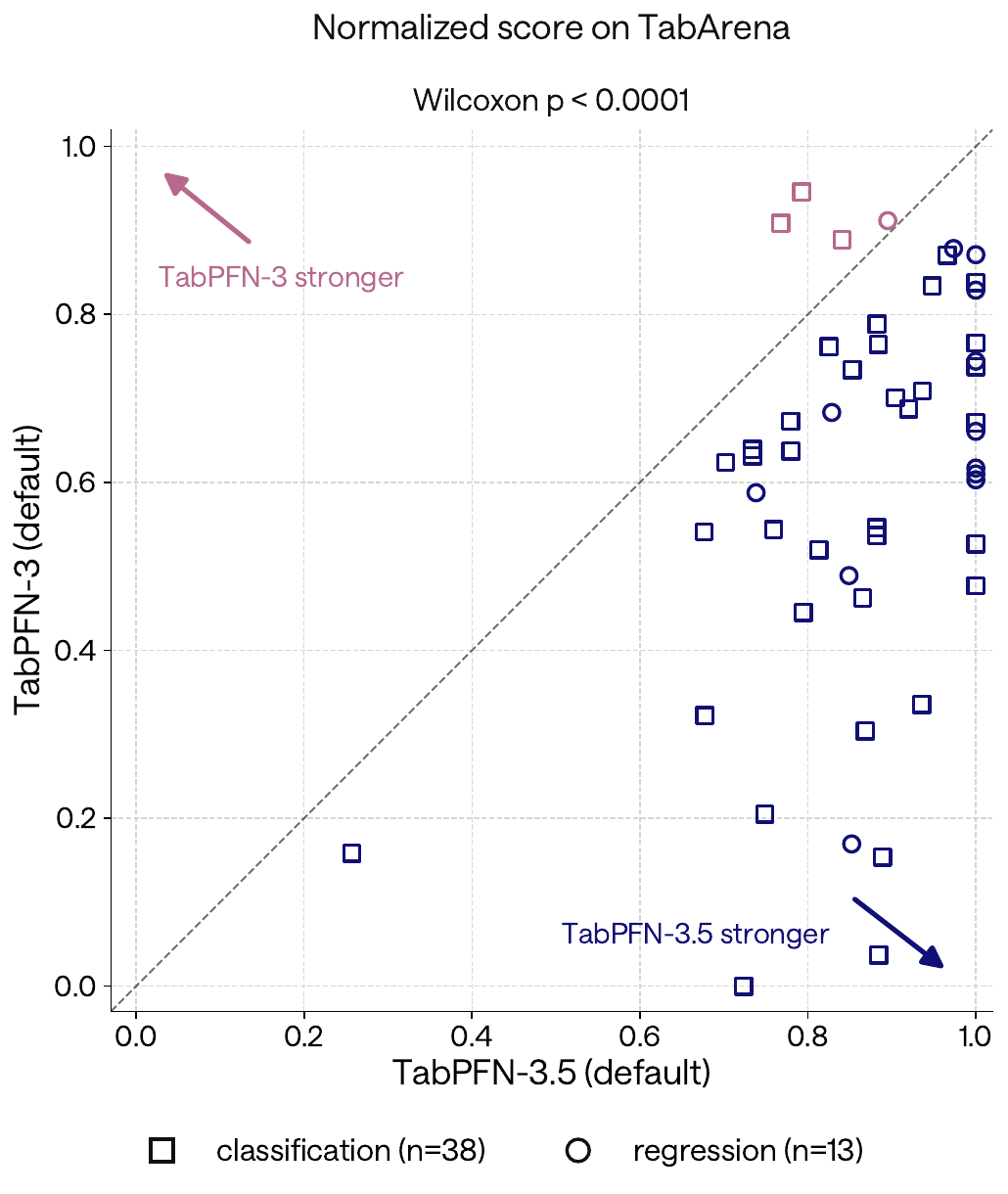}
    \caption{\textbf{\ourmodel{} substantially improves on \ourmodelthree.} We report
    the per-dataset scores on TabArena \citep{erickson2025tabarena}.
    The normalized score is based on a per-fold min-max normalization as explained in~\citep{grinsztajn2026tabpfn3technicalreport}.}
    \label{fig:tabarena-per-dataset-vs-tabpfn-3}
  \end{subfigure}
  
\caption{\textbf{Evolution and performance of the TabPFN model family.}
  (\subref{tab:tabpfn-variants}) Supported input shapes and model size across
  releases. \ourmodel{} raises the recommended feature limit and can simultaneously handle 1M data points and 6k features. %
  (\subref{fig:tabarena-per-dataset-vs-tabpfn-3}) Per-dataset scores on TabArena.
  Points below the diagonal indicate stronger \ourmodel{} performance; a
  Wilcoxon signed-rank test confirms the improvement is significant
  ($p < 0.0001$).}
  \label{fig:table-and-figure}
\end{figure}

\section{Experimental Results}
\label{sec:experimental_results}

\ourmodel{} improves performance across a wide range of settings:
  \begin{itemize}
    \item \textbf{Standard tabular benchmarks.} On TabArena, \ourmodel{} ranks first, and together with \ourmodelfast{} and \ourmodelthinking{} dominates the Elo-versus-time Pareto frontier (\Cref{sec:tabarena}). It also ranks first on the 300-dataset TALENT benchmark (\Cref{sec:talent}).
    \item \textbf{Diverse and non-i.i.d.\ data.} On BeyondArena, which spans grouped and temporal splits, tiny to million-row tables, and text and high-cardinality features, \ourmodel{} ranks first overall, and for every subset of non-large ($<$100K rows) datasets (\Cref{sec:beyondarena}).
    \item \textbf{Multimodal data.} \ourmodel{} ranks first on STRABLE, where tables carry string fields such as names, addresses and free-text notes (\Cref{sec:strable}), and on MulTaBench, where tabular
  columns are paired with text and image fields (\Cref{sec:multabench}). \ourmodelplus{}, which processes text columns natively, and \ourmodelthinking{} extend the lead further on both benchmarks (\Cref{sec:tabpfn3plus}).
    \item \textbf{Predictive distributions.} Beyond point predictions, \ourmodel{} produces better predictive distributions, ranking first on ScoringBench, which scores the full predicted distribution of regression models
  with proper scoring rules (\Cref{sec:scoringbench}).
    \item \textbf{Task-specific harnesses.} The gains carry over to task-specific harnesses: TabPFN-Rel built on \ourmodelplus{} ranks first on \RelArena for relational data (\Cref{sec:relarena}), and the general \ourmodel{}
  checkpoint, with no time-series finetuning, beats the time-series-specific TabPFN-TS-3 on fev-bench (\Cref{sec:fevbench}).
\end{itemize}

Together, \ourmodel{}, \ourmodelfast{} and \ourmodelthinking{} improve on \ourmodelthree{} at every time budget. \ourmodel{} itself runs up to $2\times$ slower than \ourmodelthree{} on large training sets because of its doubled model width;
\ourmodelfast{} is up to $3\times$ faster than \ourmodelthree{} while still about 150 Elo points ahead on TabArena. See \Cref{sec:inference} for inference-speed measurements and details on how our \ourmodelplus offering makes \ourmodel and \ourmodelfast even faster.

\subsection{TabArena}
\label{sec:tabarena}

TabArena \citep{erickson2025tabarena} has become increasingly competitive with the arrival of strong tabular foundation models such as TabFM~\citep{tabfm2026} and EXAONE-Tabular \citep{eo2026exaonetabular}. Indeed, every method in the current top ten is either a tabular foundation model in its default configuration or a highly optimized AutoML system. %

\begin{figure}
    \centering
    \includegraphics[width=\linewidth]{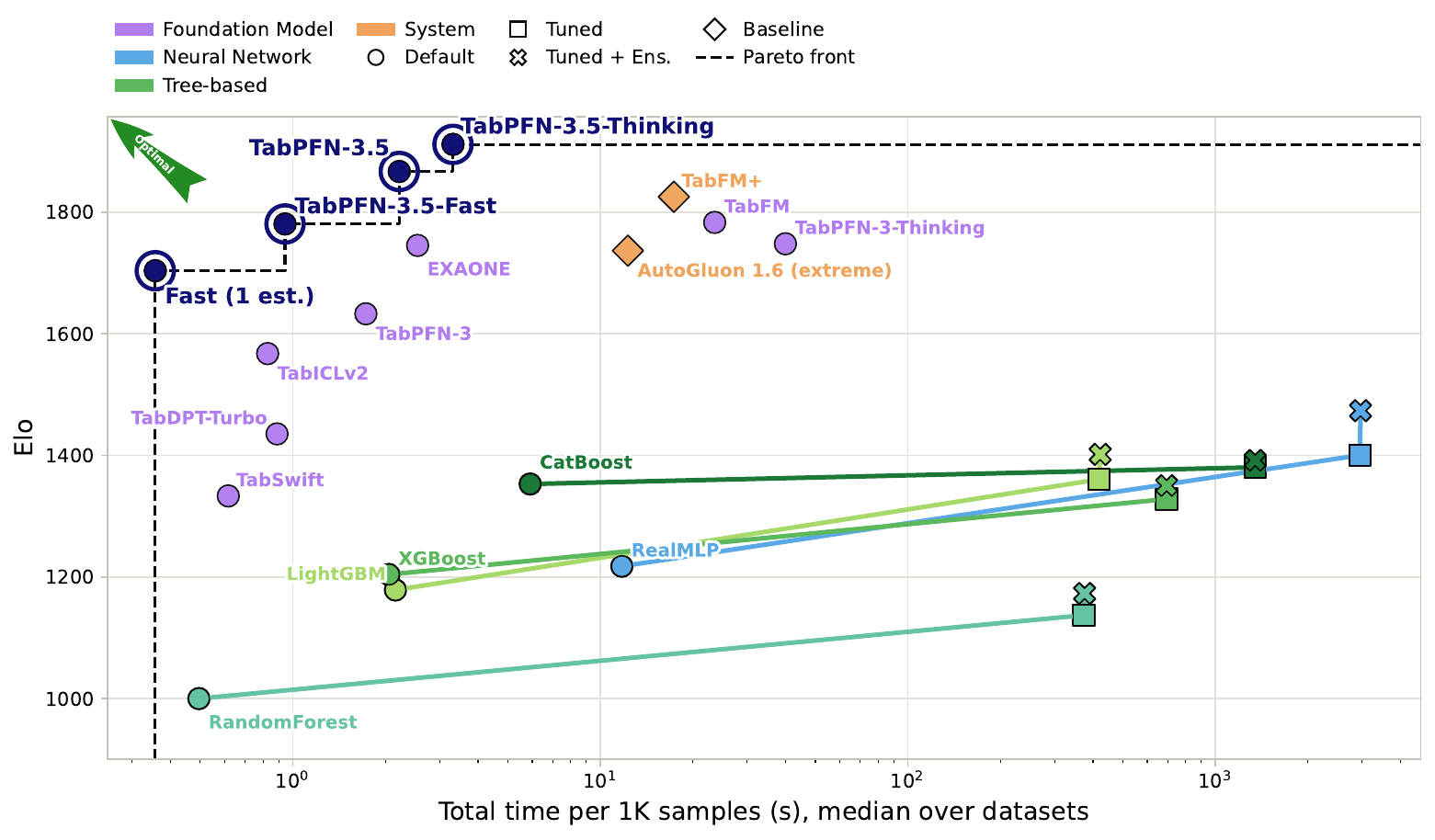}
    \caption{\textbf{TabArena Elo against total time.} Total time is the fit time per 1{,}000 training rows plus the prediction time per 1{,}000 test rows, median over the 51 datasets, on a log scale; \Cref{sec:app:tabarena_timing} describes the measurement.}
    \label{fig:ta_pareto_elo}
\end{figure}

\Cref{fig:ta_pareto_elo} shows the Pareto frontier for Elo and total fit-and-predict time; \Cref{fig:ta_pareto_improv} shows the corresponding improvability frontier. Against a strong set of recent baselines, \ourmodelthinking{}, \ourmodel{} and \ourmodelfast{} form the Pareto frontier: at every time budget one of them reaches the highest Elo. \ourmodel{} beats TabFM by 84 Elo points using one-tenth of its total time per sample, and outperforms AutoGluon~1.6 extreme~\citep{autogluon_tabular} by 130 Elo points in a fifth of the time. \Cref{sec:app:tabarena_timing} describes how the time axis is measured.

\subsection{BeyondArena}
\label{sec:beyondarena}

BeyondArena~\citep{purucker2026beyond} tests whether tabular models generalize beyond the small, IID settings emphasized by standard benchmarks. It includes datasets with different training and test distributions, spanning grouped splits (predicting on data from a subgroup unseen during training, e.g.\ a new country or a new hospital), and temporal splits (predicting on test samples observed later in time than the training samples). BeyondArena also contains tiny to million-row datasets; low and high-dimensional tables; and text and high-cardinality features. Earlier TFMs performed strongly on small IID datasets but remained behind tuned conventional models in several of these harder regimes.

\begin{figure}[t]
    \centering
    \includegraphics[width=\linewidth]{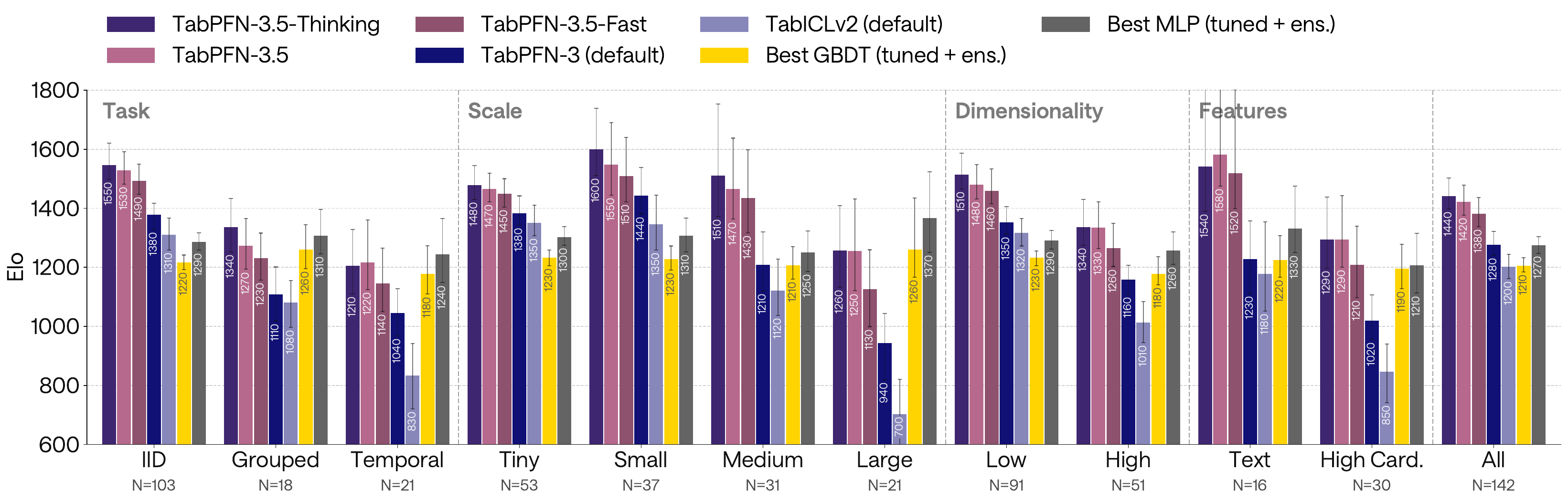}
    \caption{\textbf{Performance on BeyondArena core.} Elo scores compare \ourmodel with \ourmodelthree, TabICLv2, and the strongest tuned and ensembled GBDT and MLP baselines, overall and by data slice. Error bars show bootstrap confidence intervals. The labels of the Elo scores are rounded to the nearest multiple of 10. The group identifier is exposed to the TabPFN models on label-per-sample grouped datasets; see \Cref{sec:app:beyondarena:no_group_id}.}
    \label{fig:ba-core}
\end{figure}

\ourmodel establishes a new state of the art on the full BeyondArena core (\Cref{fig:ba-core}). It ranks first overall and improves over \ourmodelthree in every slice, with particularly strong results on IID, high-dimensional, high-cardinality, and text data. The confidence intervals are wide for some of the smaller slices, so their exact ordering should be interpreted cautiously. Tuned and ensembled MLPs retain the highest performance on grouped, temporal, and large datasets, but \ourmodel substantially narrows these gaps. The temporal (and to a lesser extent the grouped) subset of BeyondArena is confounded with size. When restricted to BeyondArena non-large (datasets with up to 100K rows), TabPFN-3.5 matches the best baselines for grouped and temporal datasets. See Appendix \ref{sec:app:beyondarena:no_group_id} for more details on how we evaluate models on grouped datasets, which differs slightly from the current BeyondArena procedure.

In Appendix \ref{sec:BA_TFMs}, we compare to more recent TFMs that are absent from the official BeyondArena results and cannot run on the full benchmark. This comparison confirms its lead on the evaluated subset.

Finally, we also show the results of TabPFN-3.5-Thinking and TabPFN-3.5-Fast on BeyondArena in Figure \ref{fig:ba-core}. TabPFN-3.5-Thinking improves on TabPFN-3.5 by 20 Elo points, while TabPFN-3.5-Fast is 3x faster and scores only 40 Elo points lower.

\begin{figure}[t]
    \centering
    \includegraphics[width=\linewidth]{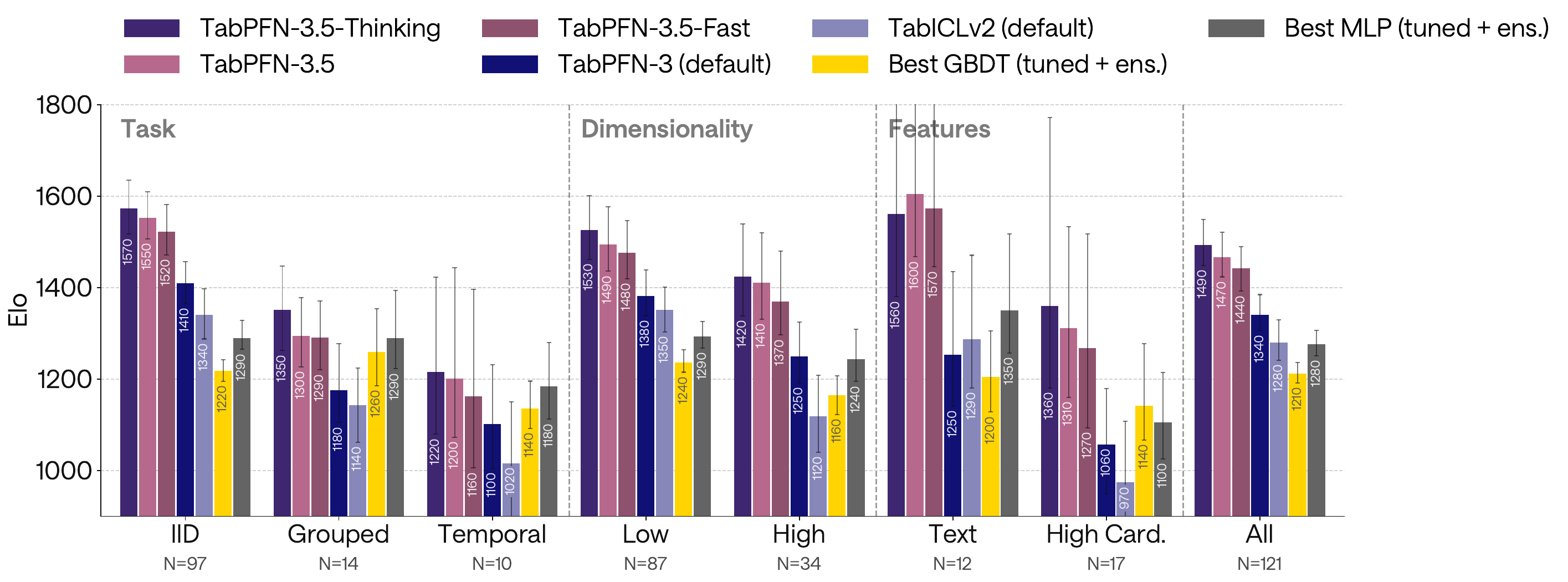}
    \caption{\textbf{BeyondArena without large tables.} \ourmodel leads overall and across the reported data regimes, while matching the strongest baseline on grouped data. Error bars show bootstrap confidence intervals. The labels of the Elo scores are rounded to the nearest multiple of 10. The group identifier is exposed to the TabPFN models on label-per-sample grouped datasets; see \Cref{sec:app:beyondarena:no_group_id}.}
    \label{fig:ba-groupid-nolarge}
\end{figure}

\subsection{STRABLE}
\label{sec:strable}

Enterprise tables frequently contain product names, addresses, job titles, and identifiers alongside numeric columns.
STRABLE~\cite{blayer2026strable} is the largest benchmark of tabular learning with strings, comprising 108 datasets drawn from diverse application fields. We compare against the results published in the official paper, without re-running baselines; consequently, models released after the paper are absent from the comparison. Since TF-IDF was the strongest string encoding for STRABLE, we use it for models without native text handling, including \ourmodel{} and \ourmodelfast; models with native text handling \cite{prokhorenkova2018catboost, spinaci2026contexttab, arazi_tabstar_2025}, as well as \ourmodelplus and \ourmodelthinking{}, receive the raw table. 

\begin{figure}[t]
    \centering
    \begin{subfigure}[t]{0.49\linewidth}
        \centering
        \includegraphics[width=\linewidth]{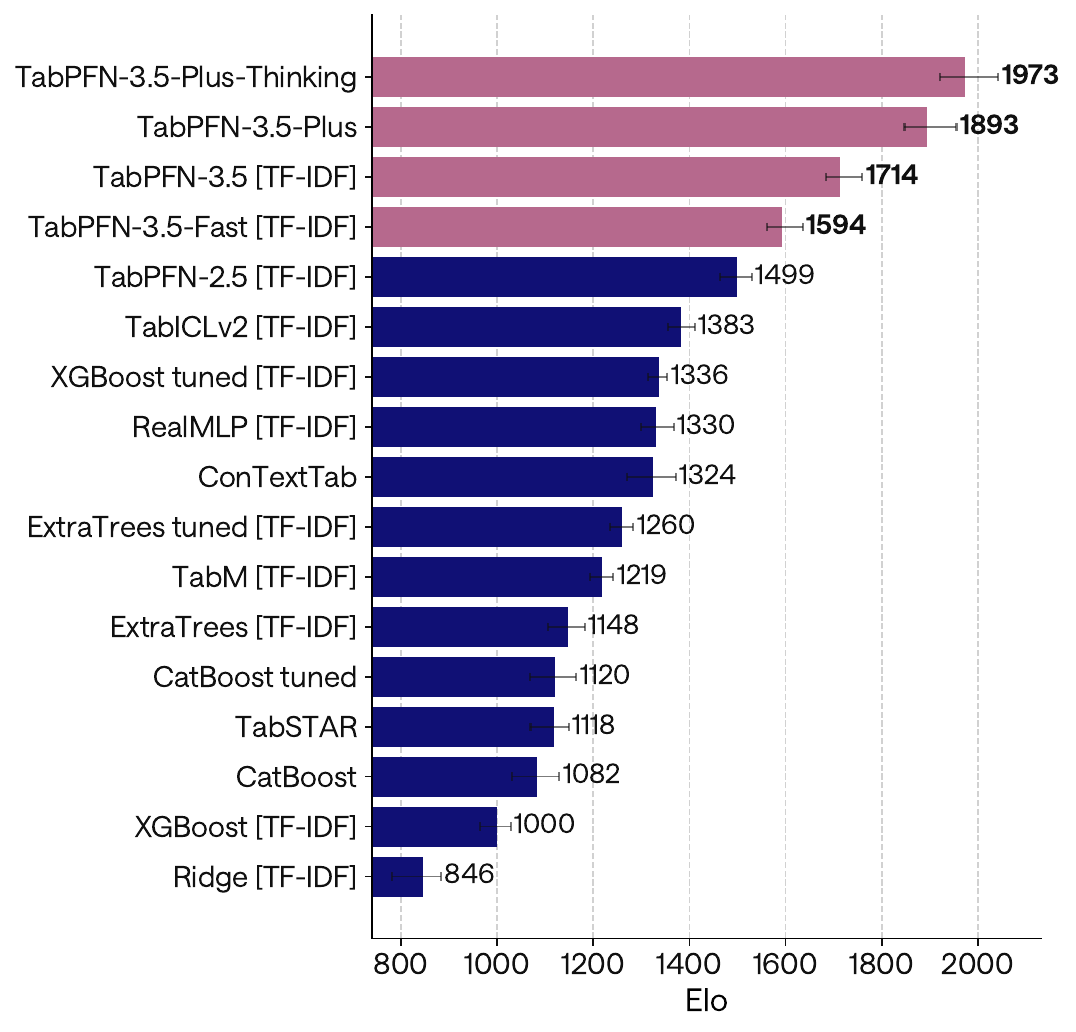}
        \caption{STRABLE with TF-IDF: 108 datasets, 3 folds each.}
        \label{fig:strable-elo-tfidf}
    \end{subfigure}
    \hfill
    \begin{subfigure}[t]{0.49\linewidth}
        \centering
        \includegraphics[width=\linewidth]{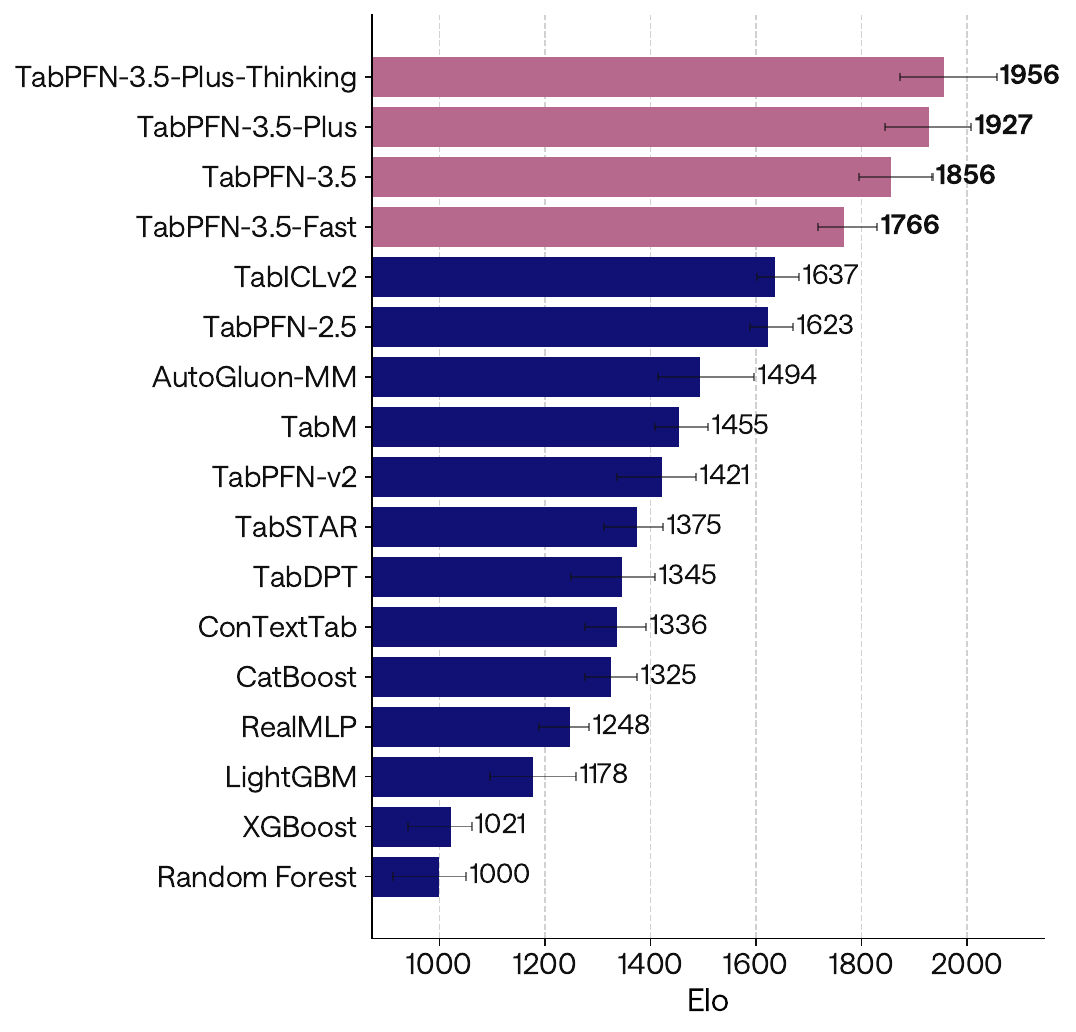}
        \caption{MulTaBench: 40 datasets, 5 folds each.}
        \label{fig:multabench-elo}
    \end{subfigure}
    \caption{\textbf{Elo on STRABLE and MulTaBench}, over the results published with each benchmark. The \ourmodel{} family significantly outperforms all published baselines on both.}
    \label{fig:strable-multabench-elo}
\end{figure}

Figure~\ref{fig:strable-elo-tfidf} shows that the \ourmodel{} family dominates the benchmark, with all variants significantly outperforming prior baselines. Appendix~\ref{sec:app-strable} shows the leaderboard including variants using the embedding model \textit{e5-small-v2} \cite{wang2022text} for string encoding.

\subsection{MulTaBench}
\label{sec:multabench}

In many high-impact domains, tabular learning problems are inherently multimodal, combining structured numerical features with unstructured inputs like free-text and images. MulTaBench~\cite{arazi2026multabenchbenchmarkingmultimodaltabular} benchmarks this setting across 40 datasets: 20 with an image column and 20 with free-text columns. We compare against the results published in the official paper without re-running baselines; consequently, models released after the paper are absent from the comparison. In the official benchmark, baselines lacking native text support use frozen \textit{e5-small-v2}~\cite{wang2022text} embeddings, and baselines lacking image support rely on frozen \textit{DINOv3-small}~\cite{simeoni2025dinov3} embeddings. Both of them are then reduced to 30 PCA \cite{abdi2010principal} components. To ensure a fair comparison, we evaluate \ourmodel{} and \ourmodelfast using the same preprocessing. 

Figure~\ref{fig:multabench-elo} also includes \ourmodelplus{} and \ourmodelthinking{}, which process the raw text columns directly (Section~\ref{sec:tabpfn3plus}) while relying on the frozen embeddings for image features. As shown in the figure, the \ourmodel{} family significantly outperforms all prior published methods. Appendix~\ref{sec:app-multabench} details the text-only half of the benchmark, where the performance gap between our models and the baselines widens further, thanks to our improved text support.

\subsection{\RelArena}
\label{sec:relarena}

\begin{figure}[h]
    \centering
    \includegraphics[width=\linewidth]{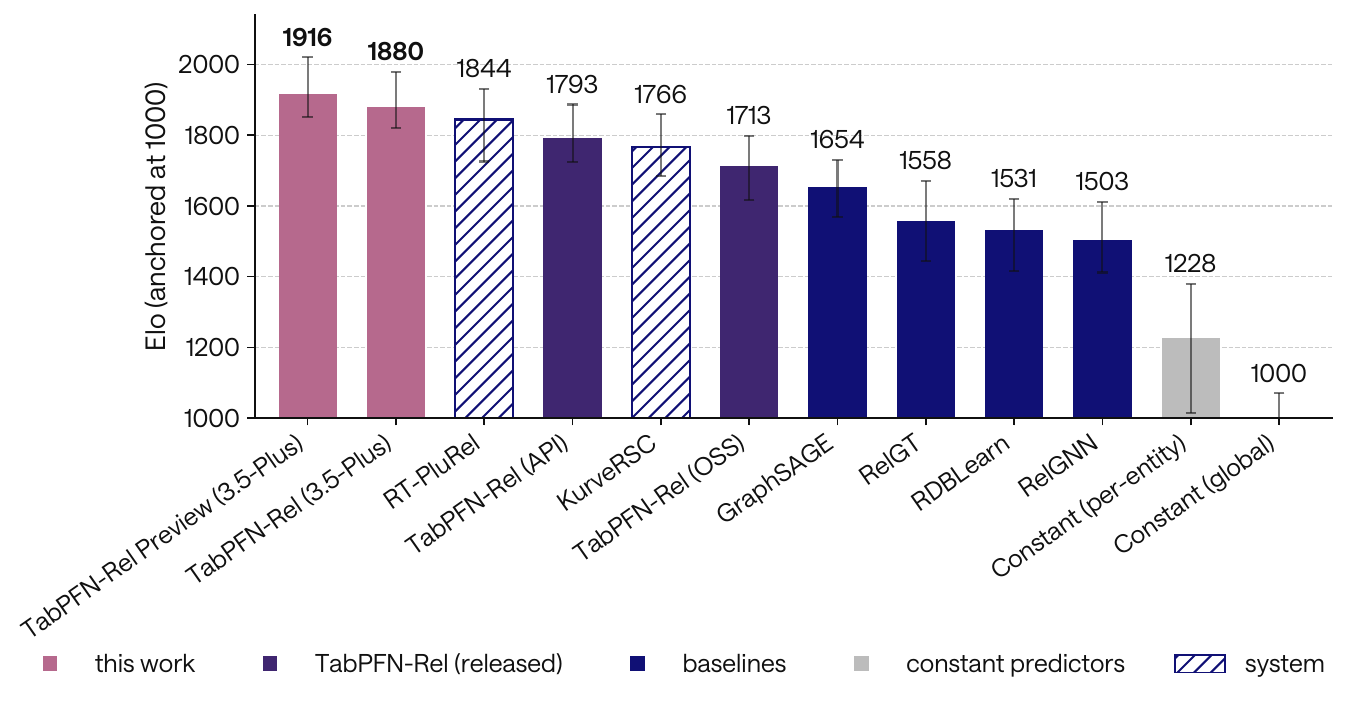}
    \caption{\textbf{Elo scores on the model + system leaderboard over 21 \RelArena tasks:} Entries marked as systems (hatched) comply with \RelArena's data states and evaluation regime but not with its standardized tuning regime (see \citet{hayler2026relarena} for further details). We observe that updating the base model from \ourmodelthreeplus to \ourmodelplus improves performance by nearly 90 Elo points. We also preview our latest (internal) version of TabPFN-Rel, co-developed with \ourmodel, which yields an additional 36 Elo-point improvement.}
    \label{fig:relarena}
\end{figure}

As observed in previous work~\citep{grinsztajn2026tabpfn3technicalreport}, improvements in the base model tend to translate directly into stronger downstream performance on relational tasks through the TabPFN-Rel harness~\citep{hayler2026relarena, grinsztajn2026tabpfn3technicalreport}. \Cref{fig:relarena} illustrates this on the novel \RelArena benchmark~\cite{hayler2026relarena}, which enables reproducible and comparable evaluation on RelBench v1 entity-level tasks~\citep{relbenchv1}. Using TabPFN-3.5-Plus, TabPFN-Rel ranks first on both the models and model+systems leaderboards of \RelArena, outperforming the RT-PluRel system~\citep{rt, rt-j, plurel} while requiring significantly less runtime.

The comparison includes RT-PluRel~\citep{rt,plurel}, KurveRSC~\citep{madrigal2026kurversc},
GraphSAGE~\citep{graphsage,relbenchv1}, RelGT~\citep{relgt}, RDBLearn~\citep{rdblearn}, and
RelGNN~\citep{relgnn}, alongside the released TabPFN-Rel variants~\citep{hayler2026relarena,grinsztajn2026tabpfn3technicalreport}
and the global and per-entity constant predictors.

Building on these results, we continue to co-develop the TabPFN-Rel harness alongside the TabPFN base models. We additionally preview the latest internal version of the harness, which, when combined with TabPFN-3.5-Plus, improves performance by a further 36 Elo points, as also shown in \Cref{fig:relarena}. We co-developed the newer versions of the TabPFN-Rel harness with TabPFN-3.5-Plus. Results with open-source checkpoints and with Thinking mode are work in progress.

\subsection{Inference benchmarks}
\label{sec:inference}

\begin{figure}[t]
  \centering
  \includegraphics[width=\linewidth]{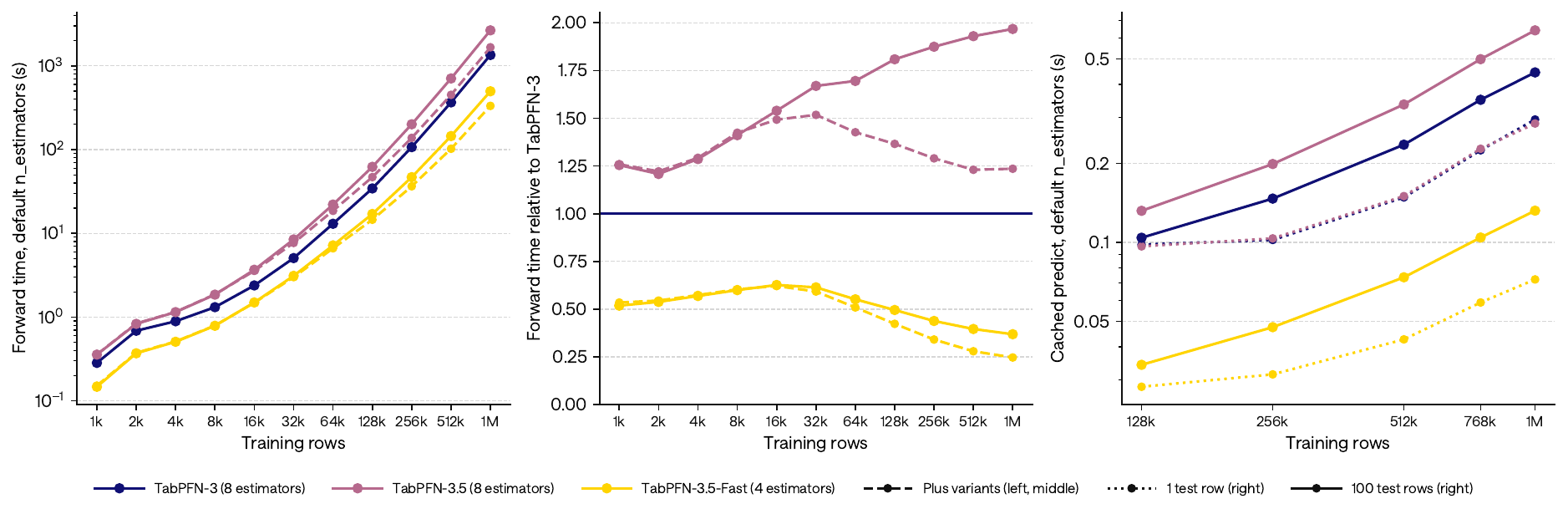}
\caption{Inference speed at 100 columns and 1{,}024 test rows absolute (left) and relative to \ourmodelthree{} (middle). Dashed lines are the same checkpoints behind TabPFN-Plus. Right: predicting 1 or
100 test rows against a KV cache built over the training rows. On a single test row, \ourmodel{} matches the speed of \ourmodelthree{}.}
  \label{fig:inference}
\end{figure}

We benchmark our new models \ourmodel{} and \ourmodelfast{} against \ourmodelthree{} on one NVIDIA RTX
PRO 6000 Blackwell (96\,GB). 
We run a single estimator, measure the model forward time, and multiply by the checkpoint's default number of
estimators (8 for \ourmodelthree{} and \ourmodel{}, 4 for \ourmodelfast{}), see \Cref{fig:inference}. As the number of training rows grows from 1k to 1M, \ourmodel{} runs up to $2\times$ slower than \ourmodelthree{} due to the increased number of attention heads. Our optimized \ourmodelplus{}, however, reduces the model's latency and only has $1.25\times$ the latency of \ourmodelthree{} due to faster FP8 attention \citep{kubler2026attention}.
\ourmodelfast{} is up to 3 times faster than \ourmodelthree{} and the optimized Plus variant can run in a quarter of the time. All scale quadratically in the number of training rows for datasets larger than 128k rows.

Like our previous models, \ourmodel{} also supports cached inference for sub-second predictions. 
The expensive computations on the training rows are done only once and only the relevant tensors are cached.
The models cache the per-block inducing states produced by the feature
distribution embedder, the keys and values of the ICL self-attention blocks, and the inputs to the many-class decoder. For the latter, \ourmodelthree{} caches the final ICL layer's train embeddings. 
We optimized this and \ourmodel{} directly caches the input keys of the many-class decoder, reducing cache size as well as computation.
The only additional new structure that needs caching for \ourmodel{} is for the ECDF features, for which we use a bucketed version to decouple its cache size from the number of training rows.

Overall, the size of the cache remains largely unchanged between \ourmodelthree{} and \ourmodel{} even though the parameter count roughly quadrupled and prediction quality improved substantially. This is crucial for online inference, where a single test example is scored against the training context. This online use-case is bound by the memory bandwidth of moving the cache (and the weights). As we see in \Cref{fig:inference} for a single test row, the cached predict times between \ourmodel{} and \ourmodelthree{} align exactly, whereas \ourmodelfast{} is around three times faster.

\section{Model Design and Inference}
\label{sec:tabpfn35}

This section describes the changes from \ourmodelthree{} to \ourmodel{}: revised per-cell encodings, an architecture optimized for joint multitask training, and a doubled model width (\Cref{sec:arch_overview}), simpler preprocessing (\Cref{sec:preprocessing}), and a synthetic prior tuned to high-cardinality, wide, and grouped data (\Cref{sec:prior}). 
\Cref{sec:tabpfn3plus} introduces \ourmodelplus{} and its Thinking mode.

\subsection{Architecture}
\label{sec:arch_overview}

The architecture of \ourmodel{} retains 
\ourmodelthree’s~\citep{grinsztajn2026tabpfn3technicalreport} overall design while revising several key components, summarized below.

\begin{figure}[htbp]
    \centering
    \includegraphics[width=1.0\linewidth]{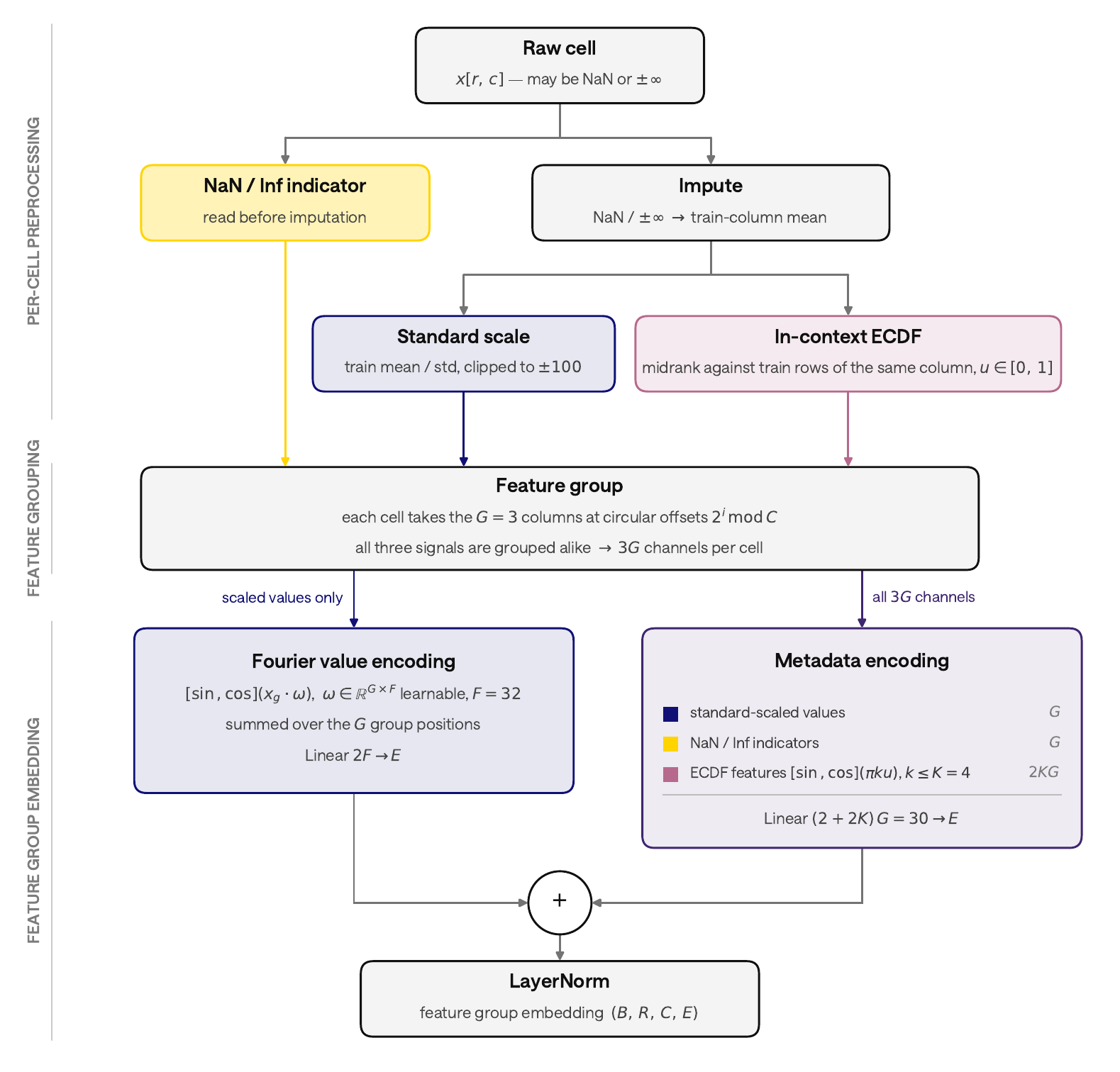}
    \caption{\textbf{Per-cell encoding in \ourmodel{}.} B is the batch size, R and C the numbers of rows and columns, E = 128 the cell embedding width; each group spans G = 3 columns, the Fourier encoder uses F = 32 learned frequencies, and the ECDF encoder K = 4 harmonics. Preprocessing (top) records a missing/infinite
    indicator before imputing with the training-column mean, then derives two signals from the imputed
    value: a standard-scaled value and an in-context ECDF rank $u \in [0,1]$ obtained by midranking
    against the training rows of the same column. Feature grouping (middle) collects the $G = 3$
    columns at circular offsets $2^j \bmod C$ for all three signals, giving $3G$ channels per cell.
    Group embedding (bottom) encodes the scaled values with learnable Fourier features summed over the
    $G$ group positions, while a linear metadata encoder reads all $3G$ channels: the scaled values
    passed through, the missing/infinite indicators, and the ECDF ranks expanded into low-frequency
    sine and cosine terms. The two encodings are summed and layer-normalised into the feature-group
    embedding.}
    \label{fig:cell-encoding}
\end{figure}

\textbf{Per-cell Encodings.} Cells of the table are embedded in three stages: per-cell preprocessing, feature grouping, and group embedding (\Cref{fig:cell-encoding}). We inherit the missing/infiniteness indicators and the feature grouping from \ourmodelthree{} and add two more signals: first, the standardized values are encoded with Fourier features, inspired by TabFM~\citep{tabfm2026}, rather than projected only linearly: each value is multiplied by a bank of learned frequencies and passed through sine and cosine, which captures small differences better than a linear projection of the raw number. This is particularly advantageous for modeling ordinal-encoded categorical variables with high cardinality.
Second, we introduce in-context \emph{Empirical Cumulative Distribution Function (ECDF)} features: each cell is ranked against the cell values of the training rows of its own column, and the resulting $u \in [0,1]$ is expanded into low-frequency sine and cosine terms. This places a value in its column's distribution rather than on its raw scale: midranks are unchanged by any strictly increasing transformation of a column, so a skewed feature and its logarithm yield identical ECDF terms, and a heavy tail that standard scaling crushes against the $\pm 100$ clip is spread evenly over $[0,1]$. As in \ourmodelthree{}, the standardized values are also passed through, so the exact magnitudes survive alongside the two new encodings. The target embedding is unchanged from \ourmodelthree{}: a learned embedding per class and a linear projection for regression targets.

\textbf{Multitask Model.} We modified our model to be trained jointly on classification and regression, with the task type supplied as an input. Only the label encoder and the output head are task-specific; the cell encoder, per-column distribution embedder, feature aggregator and in-context transformer are shared. Classification is read out by attention over the one-hot training labels, regression by a predicted distribution over fixed bins. One run therefore yields a single checkpoint covering both tasks, reducing training cost and simplifying the inference stack. To keep training stable across the two tasks and the larger model width, we also added some new normalization layers throughout the network. Specifically, we added QK-norm (an RMSNorm on queries and keys) in the attention blocks, a LayerNorm after the input encoding, and an RMSNorm before the task-specific head.

\textbf{Increased Capacity.}
We widen the in-context transformer from 512 to 1024 dimensions compared to \ourmodelthree{} by using eight CLS aggregation tokens per row instead of four. The attention head count doubles with it, from 8 to 16, so the per-head dimension stays at 64, and test rows still attend through a single 64-dimensional key/value head. 
Doubling the dimension quadruples the parameter count in the linear layers, which dominate the overall parameter count. However, for large datasets (many rows), the inference time only doubles, as it is linear in the width rather than the parameter count. 
Furthermore, the grouped-query attention for the test rows ensures that the size of the KV cache, and thus the cached-prediction time remains largely unchanged compared to \ourmodelthree{}, at significantly improved accuracy, see~\Cref{sec:inference}.

\subsection{Preprocessing}
\label{sec:preprocessing}
Unlike previous TabPFN versions, \ourmodel{} does not rely on additional feature transformations such as quantile transformations or robust scaling, and does not augment the dataset with singular value decomposition (SVD) components. The improved cell encodings, described above and shown in \Cref{fig:cell-encoding}, provide a sufficiently expressive representation of each cell for the model to handle diverse datasets natively. 
We still use multiple estimators to form a final prediction: by default, \ourmodel{} aggregates eight and \ourmodelfast{} four estimators that operate on distinct permutations of the dataset.

\subsection{Synthetic Prior}
\label{sec:prior}
Our synthetic data generation pipeline closely follows TabPFN-3 \citep{grinsztajn2026tabpfn3technicalreport}. We make some general improvements to diversity and scalability, alongside taking inspiration from the TabICLv2 prior \citep{qu2026tabiclv2}. Furthermore, we put a specific emphasis on generating more realistic datasets that exhibit high-cardinality categorical variables in conjunction with our encoding improvements detailed in \Cref{sec:arch_overview}. Moreover, we adjusted our prior to generate more realistic datasets with large feature counts and grouped data, where the test set is from a different group than the train set. \Cref{fig:ba-groupid-nolarge} shows the effect on the grouped, high-feature-count, and high-cardinality slices of BeyondArena.

\subsection{\ourmodelplus and Thinking mode}
\label{sec:tabpfn3plus}

On top of \ourmodel{}, which we release open-source, our API and enterprise deployments provide access
to \ourmodelplus and \ourmodelthinking. These variants are fully compatible with the open-source \ourmodel{} interface and can be used as a drop-in replacement, while offering additional capabilities.

While our open-source release now supports preprocessing of dates and text features,  \textbf{\ourmodelplus{}} further improves on multimodal data.
Furthermore, \ourmodelplus{} provides latency-optimized inference, most notably through FP8 attention that accelerates inference on datasets with many rows \citep{kubler2026attention}.

\textbf{\ourmodelthinking} applies additional inference-time computation on top of
\ourmodelplus to push prediction quality further. Thinking mode composes with native multimodal
support, so a single call can handle mixed numerical, categorical, date, and text columns under the same
inference-time-compute regime. We emphasize that our Thinking mode achieves this strong performance
while relying only on \ourmodel{}, without using LLMs, real data, internet search, or any other model.

These models substantially improve on \ourmodelthree-Plus and \ourmodelthree-Thinking \citep{grinsztajn2026tabpfn3technicalreport}. First, both benefit directly from the stronger base model. Second, we have significantly improved their implementation. Notably, \ourmodelthinking is twelve times faster than \ourmodelthree-Thinking, and \ourmodelplus now comes with inference speed optimizations. The implementation details of \ourmodelplus{} and \ourmodelthinking{} are proprietary and are intentionally not described in this report.

\ourmodelplus and \ourmodelthinking are available through our API and through enterprise deployments including on-prem, Virtual Private Cloud (VPC) deployment on AWS SageMaker and Azure AI Foundry, and SAP AI Core; see \Cref{sec:platform_availability} and \Cref{sec:license} for deployment options, licensing and access. Benchmark results are reported in \Cref{sec:experimental_results}.

\section{License and Availability}
\label{sec:license}

Our open-source release has two parts under two licenses: the code, and the model weights.

\paragraph{Code.} Our open-source code, including the \texttt{tabpfn} package\footnote{\url{https://github.com/PriorLabs/TabPFN}} that runs the models and the Python client SDK\footnote{\url{https://github.com/PriorLabs/tabpfn-client}}, is released under the Apache License 2.0.

\paragraph{Model weights.} We release the \ourmodel{} and \ourmodelfast{} weights under the TABPFN-3.5 License v1.0, designed to be permissive for academic use, research, and evaluation in commercial settings. The license explicitly allows testing, evaluation, and internal benchmarking, so an organization can download the models and run preliminary assessments on its own datasets without a commercial agreement. The key restriction is that the models, their derivatives, and their outputs cannot be used for commercial or production purposes. This includes, but is not limited to, revenue-generating products, competitive benchmarking for procurement decisions, client deliverables, and using model outputs as inputs to internal commercial decision-making. The full TABPFN-3.5 License v1.0 text is available at \url{https://huggingface.co/Prior-Labs/tabpfn_3_5/blob/main/LICENSE}.

\paragraph{Commercial use.} For production use, we offer a Commercial Enterprise License, available for our managed API, SAP AI Core, Virtual Private Cloud deployments (at the time of publication: AWS SageMaker \& Azure AI Foundry), and on-prem or other custom deployment modes across other software platforms such as Databricks. The Commercial Enterprise License provides access to our proprietary high-speed inference engine, dedicated support, integration tooling, additional internal models, and the \ourmodelplus{} and \ourmodelthinking{} variants, which are not available as part of the open-source release. The managed API runs on our optimized GPU infrastructure and is the recommended option for users without dedicated local GPUs; it is accessible via the Python client SDK (\texttt{pip install tabpfn-client}) or a standard REST API. For commercial licensing inquiries, please contact \href{mailto:sales@priorlabs.ai}{sales@priorlabs.ai}.

\subsection{Distribution}
\label{sec:platform_availability}
\ourmodel{} and \ourmodelfast{} are available through the open-source PyPI distribution for evaluation and non-commercial use, and through a managed API for commercial workloads. The models are listed on the AWS SageMaker Marketplace\footnote{\url{https://aws.amazon.com/marketplace/pp/prodview-chfhncrdzlb3s}}, the Azure AI Foundry Model Catalog\footnote{\url{https://ai.azure.com/catalog/models/TabPFN-2.5}} and SAP AI Core with full support for batch and real-time inference on classification and regression tasks; \ourmodel{} will become available on these marketplaces with this report. A reference integration for Databricks is available through the Databricks Industry Solutions repository\footnote{\url{https://github.com/databricks-industry-solutions/tabpfn-databricks}}. See \Cref{sec:license} for license terms, commercial-use scope, and whom to contact about production deployment.

\bibliographystyle{unsrtnat}
\bibliography{references}

\begin{thebibliography}{64}
\providecommand{\natexlab}[1]{#1}
\providecommand{\url}[1]{\texttt{#1}}
\expandafter\ifx\csname urlstyle\endcsname\relax
  \providecommand{\doi}[1]{doi: #1}\else
  \providecommand{\doi}{doi: \begingroup \urlstyle{rm}\Url}\fi

\bibitem[Grinsztajn et~al.(2026)Grinsztajn, Flöge, Key, Birkel, Jund, Roof,
  Manium, Hoo, Bühler, Garg, Safaric, Robertson, Jäger, Alessi, Hayler,
  Moroshan, Purucker, Singer, Arazi, Siems, Metzen, Grab, Erickson, Guo,
  Kalfon, Bing, Salinas, Cornu, Wehrhahn, Kriuchkova, Kaya, Sidhoum, Salmon,
  Chen, Hulsebos, LeCun, Müller, Schölkopf, Gambhir, Hollmann, and
  Hutter]{grinsztajn2026tabpfn3technicalreport}
Léo Grinsztajn, Klemens Flöge, Oscar Key, Felix Birkel, Philipp Jund, Brendan
  Roof, Mihir Manium, Shi~Bin Hoo, Magnus Bühler, Anurag Garg, Dominik
  Safaric, Jake Robertson, Benjamin Jäger, Simone Alessi, Adrian Hayler,
  Vladyslav Moroshan, Lennart Purucker, Philipp Singer, Alan Arazi, Julien
  Siems, Jan~Hendrik Metzen, Georg Grab, Nick Erickson, Siyuan Guo, Eliott
  Kalfon, Simon Bing, David Salinas, Clara Cornu, Lilly~Charlotte Wehrhahn,
  Diana Kriuchkova, Kursat Kaya, Lydia Sidhoum, Marie Salmon, Jerry Chen,
  Madelon Hulsebos, Yann LeCun, Samuel Müller, Bernhard Schölkopf, Sauraj
  Gambhir, Noah Hollmann, and Frank Hutter.
\newblock {TabPFN-3}: Technical report, 2026.
\newblock URL \url{https://arxiv.org/abs/2605.13986}.

\bibitem[Erickson et~al.(2025)Erickson, Purucker, Tschalzev, Holzm{\"u}ller,
  Desai, Salinas, and Hutter]{erickson2025tabarena}
Nick Erickson, Lennart Purucker, Andrej Tschalzev, David Holzm{\"u}ller,
  Prateek Desai, David Salinas, and Frank Hutter.
\newblock {TabArena}: A living benchmark for machine learning on tabular data.
\newblock In D.~Belgrave, C.~Zhang, H.~Lin, R.~Pascanu, P.~Koniusz,
  M.~Ghassemi, and N.~Chen, editors, \emph{Advances in Neural Information
  Processing Systems}, volume~38, pages 17285--17350. Curran Associates, Inc.,
  2025.
\newblock \doi{10.52202/085713-0519}.
\newblock URL
  \url{https://proceedings.neurips.cc/paper_files/paper/2025/file/1697e3fb412da11dc9488249f9e7bbc9-Paper-Datasets_and_Benchmarks_Track.pdf}.

\bibitem[Purucker et~al.(2026)Purucker, Tschalzev, Erickson, Blayer,
  Holzm{\"u}ller, Arazi, Pfefferle, Tajjar, Varoquaux, and
  Hutter]{purucker2026beyond}
Lennart Purucker, Andrej Tschalzev, Nick Erickson, Gioia Blayer, David
  Holzm{\"u}ller, Alan Arazi, Alexander Pfefferle, Mustafa Tajjar, Ga{\"e}l
  Varoquaux, and Frank Hutter.
\newblock Beyond {IID}: How general are tabular foundation models, really?
\newblock \emph{arXiv preprint arXiv:2606.30410}, 2026.

\bibitem[Blayer et~al.(2026)Blayer, Kim, Lefebvre, Purucker, Arazi, Shapira,
  Reichart, Hutter, Le~Morvan, Holzm{\"u}ller, and
  Varoquaux]{blayer2026strable}
Gioia Blayer, Myung~Jun Kim, F{\'e}lix Lefebvre, Lennart Purucker, Alan Arazi,
  Eilam Shapira, Roi Reichart, Frank Hutter, Marine Le~Morvan, David
  Holzm{\"u}ller, and Ga{\"e}l Varoquaux.
\newblock {STRABLE}: Benchmarking tabular machine learning with strings.
\newblock \emph{arXiv preprint arXiv:2605.12292}, 2026.

\bibitem[Arazi et~al.(2026)Arazi, Shapira, Grunblat, Ventura, Hoffer, Blayer,
  Holzmüller, Purucker, Varoquaux, Hutter, and
  Reichart]{arazi2026multabenchbenchmarkingmultimodaltabular}
Alan Arazi, Eilam Shapira, Shoham Grunblat, Mor Ventura, Elad Hoffer, Gioia
  Blayer, David Holzmüller, Lennart Purucker, Gaël Varoquaux, Frank Hutter,
  and Roi Reichart.
\newblock Multabench: Benchmarking multimodal tabular learning with text and
  image, 2026.
\newblock URL \url{https://arxiv.org/abs/2605.10616}.

\bibitem[Hayler et~al.(2026)Hayler, Fl{\"o}ge, Arazi, Ranjan, Leskovec, Birkel,
  Roof, Garg, Collins, Sidhoum, K{\"u}bler, Guo, Key, Metzen, Grace, Salinas,
  Cahu, Bing, J{\"a}ger, {\c{C}}elik, Manium, Monteiro, Robertson, Chen,
  Kalfon, Pereda, Wehrhahn, Safaric, Schroeder, Grab, Kriuchkova, Cornu,
  Singer, Erickson, Balazadeh, Salmon, Alessi, Kaya, Jund, Grinsztajn, LeCun,
  Sch{\"o}lkopf, Hulsebos, Purucker, Gambhir, Hutter, and
  Hollmann]{hayler2026relarena}
Adrian Hayler, Klemens Fl{\"o}ge, Alan Arazi, Rishabh Ranjan, Jure Leskovec,
  Felix Birkel, Brendan Roof, Anurag Garg, Kristina Collins, Lydia Sidhoum,
  Jonas K{\"u}bler, Siyuan Guo, Oscar Key, Jan~Hendrik Metzen, Rylee Grace,
  David Salinas, Arthur Cahu, Simon Bing, Benjamin J{\"a}ger, Tuana
  {\c{C}}elik, Mihir Manium, Vitor Monteiro, Jake Robertson, Jerry Chen, Eliott
  Kalfon, Tom{\'a}s Pereda, Lilly Wehrhahn, Dominik Safaric, Tobias Schroeder,
  Georg Grab, Diana Kriuchkova, Clara Cornu, Philipp Singer, Nick Erickson,
  Vahid Balazadeh, Marie Salmon, Simone Alessi, K{\"u}r{\c{s}}at Kaya, Philipp
  Jund, L{\'e}o Grinsztajn, Yann LeCun, Bernhard Sch{\"o}lkopf, Madelon
  Hulsebos, Lennart Purucker, Sauraj Gambhir, Frank Hutter, and Noah Hollmann.
\newblock Advancing open and reproducible relational learning:
  {RelArena}-$\alpha$, {TabPFN-Rel} and {RPI}.
\newblock \emph{arXiv preprint arXiv:2608.16319}, 2026.
\newblock URL \url{https://arxiv.org/abs/2608.16319}.

\bibitem[Shchur et~al.(2025)Shchur, Ansari, Turkmen, Stella, Erickson, Guerron,
  Bohlke-Schneider, and Wang]{shchur2025fev}
Oleksandr Shchur, Abdul~Fatir Ansari, Caner Turkmen, Lorenzo Stella, Nick
  Erickson, Pablo Guerron, Michael Bohlke-Schneider, and Yuyang Wang.
\newblock fev-bench: A realistic benchmark for time series forecasting.
\newblock \emph{arXiv preprint arXiv:2509.26468}, 2025.

\bibitem[Henry et~al.(2015)Henry, Hager, Pronovost, and
  Saria]{henry2015targeted}
Katharine~E Henry, David~N Hager, Peter~J Pronovost, and Suchi Saria.
\newblock A targeted real-time early warning score (trewscore) for septic
  shock.
\newblock \emph{Science Translational Medicine}, 7\penalty0 (299):\penalty0
  299ra122, 2015.

\bibitem[Khandani et~al.(2010)Khandani, Kim, and Lo]{khandani2010consumer}
Amir~E Khandani, Adlar~J Kim, and Andrew~W Lo.
\newblock Consumer credit-risk models via machine-learning algorithms.
\newblock \emph{Journal of Banking \& Finance}, 34\penalty0 (11):\penalty0
  2767--2787, 2010.

\bibitem[Carvalho et~al.(2019)Carvalho, Soares, Vita, Francisco, Basto, and
  Alcal{\'a}]{carvalho2019systematic}
Thyago~P. Carvalho, Fabr{\'\i}zzio A. A. M.~N. Soares, Roberto Vita, Roberto
  da~P. Francisco, Jo{\~a}o~P. Basto, and Symone G.~S. Alcal{\'a}.
\newblock A systematic literature review of machine learning methods applied to
  predictive maintenance.
\newblock \emph{Computers \& Industrial Engineering}, 137:\penalty0 106024,
  2019.

\bibitem[Baldi et~al.(2014)Baldi, Sadowski, and Whiteson]{baldi2014searching}
Pierre Baldi, Peter Sadowski, and Daniel Whiteson.
\newblock Searching for exotic particles in high-energy physics with deep
  learning.
\newblock \emph{Nature Communications}, 5\penalty0 (1):\penalty0 4308, 2014.

\bibitem[Shwartz-Ziv and Armon(2022)]{shwartz2022tabular}
Ravid Shwartz-Ziv and Amitai Armon.
\newblock Tabular data: Deep learning is not all you need.
\newblock \emph{Information Fusion}, 81:\penalty0 84--90, 2022.

\bibitem[Grinsztajn et~al.(2022)Grinsztajn, Oyallon, and
  Varoquaux]{grinsztajn2022tree}
L{\'e}o Grinsztajn, Edouard Oyallon, and Ga{\"e}l Varoquaux.
\newblock Why do tree-based models still outperform deep learning on typical
  tabular data?
\newblock \emph{Advances in Neural Information Processing Systems},
  35:\penalty0 507--520, 2022.

\bibitem[Erickson et~al.(2020)Erickson, Mueller, Shirkov, Zhang, Larroy, Li,
  and Smola]{autogluon_tabular}
Nick Erickson, Jonas Mueller, Alexander Shirkov, Hang Zhang, Pedro Larroy,
  Mu~Li, and Alexander Smola.
\newblock Autogluon-tabular: Robust and accurate automl for structured data.
\newblock \emph{arXiv preprint arXiv:2003.06505}, 2020.

\bibitem[Hollmann et~al.(2023)Hollmann, M{\"u}ller, Eggensperger, and
  Hutter]{hollmann2022tabpfnv1}
Noah Hollmann, Samuel M{\"u}ller, Katharina Eggensperger, and Frank Hutter.
\newblock {TabPFN}: A transformer that solves small tabular classification
  problems in a second.
\newblock In \emph{The Eleventh International Conference on Learning
  Representations ({ICLR})}, 2023.
\newblock URL \url{https://openreview.net/forum?id=cp5PvcI6w8_}.

\bibitem[Hollmann et~al.(2025)Hollmann, Müller, Purucker, Krishnakumar,
  Körfer, Hoo, Schirrmeister, and Hutter]{Hollmann2025tabpfnv2}
Noah Hollmann, Samuel Müller, Lennart Purucker, Arjun Krishnakumar, Max
  Körfer, Shi~Bin Hoo, Robin~Tibor Schirrmeister, and Frank Hutter.
\newblock Accurate predictions on small data with a tabular foundation model.
\newblock \emph{Nature}, 637\penalty0 (8045):\penalty0 319--326, 2025.
\newblock ISSN 1476-4687.
\newblock \doi{10.1038/s41586-024-08328-6}.
\newblock URL \url{https://doi.org/10.1038/s41586-024-08328-6}.

\bibitem[Grinsztajn et~al.(2025)Grinsztajn, Fl{\"o}ge, Key, Birkel, Jund, Roof,
  J{\"a}ger, Safaric, Alessi, Hayler, Manium, Yu, Jablonski, Hoo, Garg,
  Robertson, B{\"u}hler, Moroshan, Purucker, Cornu, Wehrhahn, Bonetto,
  Sch{\"o}lkopf, Gambhir, Hollmann, and Hutter]{TabPFN-2.5}
L{\'e}o Grinsztajn, Klemens Fl{\"o}ge, Oscar Key, Felix Birkel, Philipp Jund,
  Brendan Roof, Benjamin J{\"a}ger, Dominik Safaric, Simone Alessi, Adrian
  Hayler, Mihir Manium, Rosen Yu, Felix Jablonski, Shi~Bin Hoo, Anurag Garg,
  Jake Robertson, Magnus B{\"u}hler, Vladyslav Moroshan, Lennart Purucker,
  Clara Cornu, Lilly~Charlotte Wehrhahn, Alessandro Bonetto, Bernhard
  Sch{\"o}lkopf, Sauraj Gambhir, Noah Hollmann, and Frank Hutter.
\newblock {TabPFN-2.5}: Advancing the state of the art in tabular foundation
  models, 2025.
\newblock URL \url{https://arxiv.org/abs/2511.08667}.

\bibitem[Hoo et~al.(2024)Hoo, M{\"u}ller, Salinas, and
  Hutter]{hoo2024tabpfn_ts}
Shi~Bin Hoo, Samuel M{\"u}ller, David Salinas, and Frank Hutter.
\newblock The tabular foundation model {TabPFN} outperforms specialized time
  series forecasting models based on simple features.
\newblock In \emph{NeurIPS Workshop on Time Series in the Age of Large Models},
  2024.

\bibitem[Hoo et~al.(2025)Hoo, M{\"u}ller, Salinas, and
  Hutter]{hoo2025tables_to_time}
Shi~Bin Hoo, Samuel M{\"u}ller, David Salinas, and Frank Hutter.
\newblock From tables to time: Extending {TabPFN-v2} to time series
  forecasting, 2025.
\newblock URL \url{https://arxiv.org/abs/2501.02945}.

\bibitem[Robertson et~al.(2025)Robertson, Reuter, Guo, Hollmann, Hutter, and
  Sch{\"o}lkopf]{robertson_dopfn}
Jake Robertson, Arik Reuter, Siyuan Guo, Noah Hollmann, Frank Hutter, and
  Bernhard Sch{\"o}lkopf.
\newblock {Do-PFN}: In-context learning for causal effect estimation.
\newblock In \emph{Advances in Neural Information Processing Systems},
  volume~38, pages 193605--193642, 2025.
\newblock \doi{10.52202/085713-5814}.
\newblock URL \url{https://arxiv.org/abs/2506.06039}.

\bibitem[Balazadeh et~al.(2025)Balazadeh, Kamkari, Thomas, Li, Ma, Cresswell,
  and Krishnan]{balazadeh_causalpfn}
Vahid Balazadeh, Hamidreza Kamkari, Valentin Thomas, Benson Li, Junwei Ma,
  Jesse~C. Cresswell, and Rahul~G. Krishnan.
\newblock Causalpfn: Amortized causal effect estimation via in-context
  learning, 2025.
\newblock URL \url{https://arxiv.org/abs/2506.07918}.

\bibitem[Ma et~al.(2025{\natexlab{a}})Ma, Frauen, Javurek, and
  Feuerriegel]{feuerriegel_causalfm}
Yuchen Ma, Dennis Frauen, Emil Javurek, and Stefan Feuerriegel.
\newblock Foundation models for causal inference via prior-data fitted
  networks, 2025{\natexlab{a}}.
\newblock URL \url{https://arxiv.org/abs/2506.10914}.

\bibitem[Hayler et~al.(2025)Hayler, Huang, Ceylan, Bronstein, and
  Finkelshtein]{Hayler2025GraphsTablesZeroShot}
Adrian Hayler, Xingyue Huang, {\.I}smail~{\.I}lkan Ceylan, Michael Bronstein,
  and Ben Finkelshtein.
\newblock Bringing graphs to the table: Zero-shot node classification via
  tabular foundation models.
\newblock \emph{arXiv preprint arXiv:2509.07143}, 2025.
\newblock \doi{10.48550/arXiv.2509.07143}.
\newblock URL \url{https://arxiv.org/abs/2509.07143}.

\bibitem[Eremeev et~al.(2025)Eremeev, Platonov, Bazhenov, Babenko, and
  Prokhorenkova]{eremeev2025graphpfnpriordatafittedgraph}
Dmitry Eremeev, Oleg Platonov, Gleb Bazhenov, Artem Babenko, and Liudmila
  Prokhorenkova.
\newblock Graphpfn: A prior-data fitted graph foundation model, 2025.
\newblock URL \url{https://arxiv.org/abs/2509.21489}.

\bibitem[Rundel et~al.(2024)Rundel, Kobialka, von Crailsheim, Feurer, Nagler,
  and R{\"u}gamer]{rundel2024interpretable}
David Rundel, Julius Kobialka, Constantin von Crailsheim, Matthias Feurer,
  Thomas Nagler, and David R{\"u}gamer.
\newblock Interpretable machine learning for {TabPFN}.
\newblock In \emph{World Conference on Explainable Artificial Intelligence},
  pages 465--476. Springer, 2024.

\bibitem[Olsen and
  Christensen(2026)]{olsen2026computingconditionalshapleyvalues}
Lars Henry~Berge Olsen and Dennis Christensen.
\newblock Computing conditional shapley values using tabular foundation models,
  2026.
\newblock URL \url{https://arxiv.org/abs/2602.09489}.

\bibitem[Qi et~al.(2026)Qi, Balazadeh, Cooper, Greiner, and
  Krishnan]{qi2026survivalpfn}
Shi-ang Qi, Vahid Balazadeh, Michael Cooper, Russell Greiner, and Rahul~G
  Krishnan.
\newblock Survivalpfn: Amortizing survival prediction via in-context bayesian
  inference.
\newblock \emph{arXiv preprint arXiv:2605.15488}, 2026.

\bibitem[Seletkov et~al.(2026)Seletkov, Hager, Kaissis, Braren, Rueckert, and
  Rehms]{seletkov2026survivalincontextpriorfittedincontext}
Dmitrii Seletkov, Paul Hager, Georgios Kaissis, Rickmer Braren, Daniel
  Rueckert, and Raphael Rehms.
\newblock Survival in-context: Amortized bayesian survival analysis via
  prior-fitted networks, 2026.
\newblock URL \url{https://arxiv.org/abs/2603.29475}.

\bibitem[Yu et~al.(2025)Yu, Picard, and Ahmed]{Yu2025GITBO}
Rosen Ting-Ying Yu, Cyril Picard, and Faez Ahmed.
\newblock {GIT-BO}: High-dimensional bayesian optimization with tabular
  foundation models.
\newblock \emph{arXiv preprint arXiv:2505.20685}, 2025.
\newblock \doi{10.48550/arXiv.2505.20685}.
\newblock URL \url{https://arxiv.org/abs/2505.20685}.

\bibitem[Rogers and Ponnada(2026)]{rogers2026zeroshotbo}
Theodore Rogers and Srividya Ponnada.
\newblock Zero-shot {Bayesian} optimization with {TabPFN}: Competitive with
  state-of-the-art without per-task training.
\newblock In \emph{AutoML Conference}, 2026.
\newblock URL
  \url{https://www.amazon.science/publications/zero-shot-bayesian-optimization-with-tabpfn-competitive-with-state-of-the-art-without-per-task-training}.

\bibitem[Schiff et~al.(2026)Schiff, Lindenbaum, and Efroni]{Schiff2025TabPFNRL}
David Schiff, Ofir Lindenbaum, and Yonathan Efroni.
\newblock {ICR-RL}: Deep reinforcement learning via in-context regression.
\newblock In \emph{International Conference on Machine Learning ({ICML})},
  2026.
\newblock URL \url{https://arxiv.org/abs/2509.11259}.

\bibitem[Qu et~al.(2026)Qu, Holzm{\"u}ller, Varoquaux, and
  Le~Morvan]{qu2026tabiclv2}
Jingang Qu, David Holzm{\"u}ller, Ga{\"e}l Varoquaux, and Marine Le~Morvan.
\newblock {TabICLv2}: {A} better, faster, scalable, and open tabular foundation
  model.
\newblock In \emph{International Conference on Machine Learning}, 2026.

\bibitem[{Google Research}(2026)]{tabfm2026}
{Google Research}.
\newblock Introducing {TabFM}: A zero-shot foundation model for tabular data.
\newblock Google Research Blog, June 2026.
\newblock URL
  \url{https://research.google/blog/introducing-tabfm-a-zero-shot-foundation-model-for-tabular-data/}.
\newblock Accessed 2026-09-03.

\bibitem[Eo et~al.(2026)Eo, Suh, Cho, Kim, Kim, Nam, and
  Lee]{eo2026exaonetabular}
Moonjung Eo, Min-Kook Suh, Hye-Seung Cho, Jiwon Kim, Seoyoon Kim, Sangjun Nam,
  and Soonyoung Lee.
\newblock {EXAONE} tabular 1.0: Technical report, 2026.
\newblock URL \url{https://arxiv.org/abs/2608.25774}.

\bibitem[Hosseinzadeh et~al.(2026)Hosseinzadeh, Labach, Xue, Han, Thomas, and
  Caterini]{hosseinzadeh2026tabdpt}
Rasa Hosseinzadeh, Alex Labach, Zexin Xue, Shuyi Han, Valentin Thomas, and
  Anthony~L Caterini.
\newblock Tabdpt-turbo: Efficient in-context learning for tabular prediction.
\newblock \emph{arXiv preprint arXiv:2608.01400}, 2026.

\bibitem[Spinaci et~al.(2025)Spinaci, Polewczyk, Schambach, and
  Thelin]{spinaci2026contexttab}
Marco Spinaci, Marek Polewczyk, Maximilian Schambach, and Sam Thelin.
\newblock {ConTextTab}: A semantics-aware tabular in-context learner.
\newblock In \emph{Advances in Neural Information Processing Systems},
  volume~38, pages 163185--163220, 2025.
\newblock \doi{10.52202/085713-4918}.

\bibitem[Rubachev et~al.(2025)Rubachev, Kartashev, Gorishniy, and
  Babenko]{rubachev_tabred}
Ivan Rubachev, Nikolay Kartashev, Yury Gorishniy, and Artem Babenko.
\newblock Tabred: Analyzing pitfalls and filling the gaps in tabular deep
  learning benchmarks.
\newblock In \emph{International Conference on Learning Representations}, 2025.

\bibitem[Landsgesell et~al.(2026)Landsgesell, Knoll, and
  Wenzel]{landsgesell2026scoringbenchbenchmarkevaluatingtabular}
Jonas Landsgesell, Pascal Knoll, and Tizian Wenzel.
\newblock {ScoringBench}: A benchmark for evaluating tabular foundation models
  with proper scoring rules, 2026.
\newblock URL \url{https://arxiv.org/abs/2603.29928}.

\bibitem[Prokhorenkova et~al.(2018)Prokhorenkova, Gusev, Vorobev, Dorogush, and
  Gulin]{prokhorenkova2018catboost}
Liudmila Prokhorenkova, Gleb Gusev, Aleksandr Vorobev, Anna~Veronika Dorogush,
  and Andrey Gulin.
\newblock {CatBoost}: unbiased boosting with categorical features.
\newblock \emph{Advances in Neural Information Processing Systems}, 31, 2018.

\bibitem[Arazi et~al.(2025)Arazi, Shapira, and Reichart]{arazi_tabstar_2025}
Alan Arazi, Eilam Shapira, and Roi Reichart.
\newblock {TabSTAR}: {A} {Tabular} {Foundation} {Model} for {Tabular} {Data}
  with {Text} {Fields}.
\newblock In \emph{Advances in {Neural} {Information} {Processing} {Systems}},
  volume~38, pages 172108--172161, 2025.
\newblock URL
  \url{https://proceedings.neurips.cc/paper_files/paper/2025/file/faf6e23e198314c7728eaa6ac44ae079-Paper-Conference.pdf}.

\bibitem[Wang et~al.(2022)Wang, Yang, Huang, Jiao, Yang, Jiang, Majumder, and
  Wei]{wang2022text}
Liang Wang, Nan Yang, Xiaolong Huang, Binxing Jiao, Linjun Yang, Daxin Jiang,
  Rangan Majumder, and Furu Wei.
\newblock Text embeddings by weakly-supervised contrastive pre-training.
\newblock \emph{arXiv preprint arXiv:2212.03533}, 2022.

\bibitem[Sim{\'e}oni et~al.(2025)Sim{\'e}oni, Vo, Seitzer, Baldassarre, Oquab,
  Jose, Khalidov, Szafraniec, Yi, Ramamonjisoa, et~al.]{simeoni2025dinov3}
Oriane Sim{\'e}oni, Huy~V Vo, Maximilian Seitzer, Federico Baldassarre, Maxime
  Oquab, Cijo Jose, Vasil Khalidov, Marc Szafraniec, Seungeun Yi, Micha{\"e}l
  Ramamonjisoa, et~al.
\newblock {DINOv3}.
\newblock \emph{arXiv preprint arXiv:2508.10104}, 2025.

\bibitem[Abdi and Williams(2010)]{abdi2010principal}
Herv{\'e} Abdi and Lynne~J Williams.
\newblock Principal component analysis.
\newblock \emph{Wiley interdisciplinary reviews: computational statistics},
  2\penalty0 (4):\penalty0 433--459, 2010.

\bibitem[Robinson et~al.(2024)Robinson, Ranjan, Hu, Huang, Han, Dobles, Fey,
  Lenssen, Yuan, Zhang, He, and Leskovec]{relbenchv1}
Joshua Robinson, Rishabh Ranjan, Weihua Hu, Kexin Huang, Jiaqi Han, Alejandro
  Dobles, Matthias Fey, Jan~E. Lenssen, Yiwen Yuan, Zecheng Zhang, Xinwei He,
  and Jure Leskovec.
\newblock {RelBench}: A benchmark for deep learning on relational databases.
\newblock In \emph{Advances in Neural Information Processing Systems},
  volume~37, pages 21330--21341, 2024.
\newblock \doi{10.52202/079017-0672}.
\newblock URL \url{https://arxiv.org/abs/2407.20060}.

\bibitem[Ranjan et~al.(2026{\natexlab{a}})Ranjan, Hudovernik, Znidar,
  Kanatsoulis, Upendra, Mohammadi, Meyer, Palczewski, Guestrin, and
  Leskovec]{rt}
Rishabh Ranjan, Valter Hudovernik, Mark Znidar, Charilaos Kanatsoulis, Roshan
  Upendra, Mahmoud Mohammadi, Joe Meyer, Tom Palczewski, Carlos Guestrin, and
  Jure Leskovec.
\newblock Relational {Transformer}: {Toward} {Zero}-{Shot} {Foundation}
  {Models} for {Relational} {Data}.
\newblock In \emph{International Conference on Learning Representations
  ({ICLR})}, 2026{\natexlab{a}}.
\newblock URL \url{http://arxiv.org/abs/2510.06377}.

\bibitem[Ranjan et~al.(2026{\natexlab{b}})Ranjan, Kothapalli, Agarwal,
  Kanatsoulis, Upendra, Palczewski, Guestrin, and Leskovec]{rt-j}
Rishabh Ranjan, Vignesh Kothapalli, Harshvardhan Agarwal, Charilaos~I.
  Kanatsoulis, Roshan~Reddy Upendra, Tom Palczewski, Carlos Guestrin, and Jure
  Leskovec.
\newblock Large-{Scale} {Pretraining} unlocks {Few}-{Shot} {Prediction} for
  {Relational} {Data}.
\newblock In \emph{2nd {ICML} Workshop on Foundation Models for Structured
  Data}, 2026{\natexlab{b}}.
\newblock URL \url{https://openreview.net/forum?id=oQINTd9din}.

\bibitem[Kothapalli et~al.(2026)Kothapalli, Ranjan, Hudovernik, Dwivedi,
  Hoffart, Guestrin, and Leskovec]{plurel}
Vignesh Kothapalli, Rishabh Ranjan, Valter Hudovernik, Vijay~Prakash Dwivedi,
  Johannes Hoffart, Carlos Guestrin, and Jure Leskovec.
\newblock {PluRel}: {Synthetic} {Data} unlocks {Scaling} {Laws} for
  {Relational} {Foundation} {Models}.
\newblock In \emph{International Conference on Machine Learning ({ICML})},
  2026.
\newblock URL \url{http://arxiv.org/abs/2602.04029}.

\bibitem[Madrigal(2026)]{madrigal2026kurversc}
Wes Madrigal.
\newblock {KurveRSC}: Validation-guided relational signal compression with a
  downstream learner in the loop.
\newblock Technical report, Kurve AI, September 2026.
\newblock URL
  \url{https://github.com/kurveai/kurversc/blob/main/docs/kurversc-technical-report.pdf}.

\bibitem[Hamilton et~al.(2017)Hamilton, Ying, and Leskovec]{graphsage}
William~L. Hamilton, Rex Ying, and Jure Leskovec.
\newblock Inductive representation learning on large graphs.
\newblock In \emph{Advances in Neural Information Processing Systems},
  volume~30, 2017.
\newblock URL
  \url{https://proceedings.neurips.cc/paper/2017/hash/5dd9db5e033da9c6fb5ba83c7a7ebea9-Abstract.html}.

\bibitem[Dwivedi et~al.(2026)Dwivedi, Jaladi, Shen, Lopez, Kanatsoulis, Puri,
  Fey, and Leskovec]{relgt}
Vijay~Prakash Dwivedi, Sri Jaladi, Yangyi Shen, Federico Lopez, Charilaos
  Kanatsoulis, Rishi Puri, Matthias Fey, and Jure Leskovec.
\newblock Relational graph transformer.
\newblock In \emph{International Conference on Learning Representations
  ({ICLR})}, 2026.
\newblock URL
  \url{https://proceedings.iclr.cc/paper_files/paper/2026/hash/fa1ef43326cfb8a808c0649708187f2a-Abstract-Conference.html}.

\bibitem[Zhang et~al.(2026)Zhang, Xu, Gan, Wipf, and Wang]{rdblearn}
Yanlin Zhang, Linjie Xu, Quan Gan, David Wipf, and Minjie Wang.
\newblock Rdblearn: Simple in-context prediction over relational databases,
  2026.
\newblock URL \url{https://arxiv.org/abs/2602.18495}.

\bibitem[Chen et~al.(2025)Chen, Kanatsoulis, and Leskovec]{relgnn}
Tianlang Chen, Charilaos Kanatsoulis, and Jure Leskovec.
\newblock {RelGNN}: Composite message passing for relational deep learning.
\newblock In \emph{Proceedings of the 42nd International Conference on Machine
  Learning}, volume 267, pages 8296--8312, 2025.
\newblock URL \url{https://proceedings.mlr.press/v267/chen25ad.html}.

\bibitem[Kübler et~al.(2026)Kübler, Jäger, Flöge, Hollmann, and
  Hutter]{kubler2026attention}
Jonas~M. Kübler, Benjamin Jäger, Klemens Flöge, Noah Hollmann, and Frank
  Hutter.
\newblock Attention quantization for tabular foundation models.
\newblock \emph{arXiv preprint arXiv:2609.13031}, 2026.

\bibitem[Hoerl and Kennard(1970)]{hoerl1970ridge}
Arthur~E Hoerl and Robert~W Kennard.
\newblock Ridge regression: Biased estimation for nonorthogonal problems.
\newblock \emph{Technometrics}, 12\penalty0 (1):\penalty0 55--67, 1970.

\bibitem[Geurts et~al.(2006)Geurts, Ernst, and Wehenkel]{geurts2006extremely}
Pierre Geurts, Damien Ernst, and Louis Wehenkel.
\newblock Extremely randomized trees.
\newblock \emph{Machine learning}, 63\penalty0 (1):\penalty0 3--42, 2006.

\bibitem[Chen and Guestrin(2016)]{chen2016xgboost}
Tianqi Chen and Carlos Guestrin.
\newblock Xgboost: A scalable tree boosting system.
\newblock In \emph{Proceedings of the 22nd acm sigkdd international conference
  on knowledge discovery and data mining}, pages 785--794, 2016.

\bibitem[Holzm{\"{u}}ller et~al.(2024)Holzm{\"{u}}ller, Grinsztajn, and
  Steinwart]{holzmuller2024realmlp}
David Holzm{\"{u}}ller, L{\'{e}}o Grinsztajn, and Ingo Steinwart.
\newblock Better by default: Strong pre-tuned mlps and boosted trees on tabular
  data.
\newblock In Amir Globersons, Lester Mackey, Danielle Belgrave, Angela Fan,
  Ulrich Paquet, Jakub~M. Tomczak, and Cheng Zhang, editors, \emph{Advances in
  Neural Information Processing Systems 38: Annual Conference on Neural
  Information Processing Systems 2024, NeurIPS 2024, Vancouver, BC, Canada,
  December 10 - 15, 2024}, 2024.
\newblock URL
  \url{http://papers.nips.cc/paper\_files/paper/2024/hash/2ee1c87245956e3eaa71aaba5f5753eb-Abstract-Conference.html}.

\bibitem[Gorishniy et~al.(2025)Gorishniy, Kotelnikov, and
  Babenko]{gorishniy2024tabm}
Yury Gorishniy, Akim Kotelnikov, and Artem Babenko.
\newblock Tabm: Advancing tabular deep learning with parameter-efficient
  ensembling.
\newblock In \emph{The Thirteenth International Conference on Learning
  Representations}, 2025.
\newblock URL \url{https://openreview.net/forum?id=Sd4wYYOhmY}.

\bibitem[Ma et~al.(2025{\natexlab{b}})Ma, Thomas, Hosseinzadeh, Labach,
  Kamkari, Cresswell, Golestan, Yu, Caterini, and Volkovs]{ma2024tabdpt}
Junwei Ma, Valentin Thomas, Rasa Hosseinzadeh, Alex Labach, Hamidreza Kamkari,
  Jesse~C. Cresswell, Keyvan Golestan, Guangwei Yu, Anthony~L. Caterini, and
  Maksims Volkovs.
\newblock {TabDPT}: Scaling tabular foundation models on real data.
\newblock In \emph{Advances in Neural Information Processing Systems},
  volume~38, 2025{\natexlab{b}}.

\bibitem[Ke et~al.(2017)Ke, Meng, Finley, Wang, Chen, Ma, Ye, and
  Liu]{lightgbm}
Guolin Ke, Qi~Meng, Thomas Finley, Taifeng Wang, Wei Chen, Weidong Ma, Qiwei
  Ye, and Tie-Yan Liu.
\newblock Lightgbm: A highly efficient gradient boosting decision tree.
\newblock In I.~Guyon, U.~V. Luxburg, S.~Bengio, H.~Wallach, R.~Fergus,
  S.~Vishwanathan, and R.~Garnett, editors, \emph{Advances in Neural
  Information Processing Systems 30}, pages 3146--3154. Curran Associates,
  Inc., 2017.
\newblock URL
  \url{http://papers.nips.cc/paper/6907-lightgbm-a-highly-efficient-gradient-boosting-decision-tree.pdf}.

\bibitem[Breiman(2001)]{breiman2001random}
Leo Breiman.
\newblock Random forests.
\newblock \emph{Machine Learning}, 45\penalty0 (1):\penalty0 5--32, 2001.
\newblock URL \url{http://dx.doi.org/10.1023/A%3A1010933404324}.

\bibitem[Tang et~al.(2024)Tang, Fang, Zhou, Yang, Zhong, Hu, Kirchhoff, and
  Karypis]{tang2024autogluon}
Zhiqiang Tang, Haoyang Fang, Su~Zhou, Taojiannan Yang, Zihan Zhong, Cuixiong
  Hu, Katrin Kirchhoff, and George Karypis.
\newblock {AutoGluon-Multimodal} ({AutoMM}): Supercharging multimodal {AutoML}
  with foundation models.
\newblock In \emph{Proceedings of the Third International Conference on
  Automated Machine Learning}, 2024.
\newblock URL \url{https://proceedings.mlr.press/v256/tang24a.html}.

\bibitem[Liu et~al.(2025)Liu, Cai, Zhou, Yin, Zhou, Jiang, and
  Ye]{talent_benchmark_jmlr}
Si-Yang Liu, Hao-Run Cai, Qi-Le Zhou, Huai-Hong Yin, Tao Zhou, Jun-Peng Jiang,
  and Han-Jia Ye.
\newblock Talent: A tabular analytics and learning toolbox.
\newblock \emph{Journal of Machine Learning Research}, 26\penalty0
  (226):\penalty0 1--16, 2025.
\newblock URL \url{http://jmlr.org/papers/v26/25-0512.html}.

\bibitem[Madrigal()]{madrigal_graphreduce}
Wes Madrigal.
\newblock {GraphReduce}.
\newblock \url{https://github.com/wesmadrigal/GraphReduce}.

\end{thebibliography}

\appendix
\section{Contributors}
\label{app:contributors}

Benjamin Jäger led the development of TabPFN-3.5 and is listed first.
The order of the remaining members of the TabPFN-3.5 model development
team, Nick Erickson, Léo Grinsztajn, Felix Birkel and Klemens Flöge, was randomized.
The rest of the Prior Labs team is listed below, appearing in random
order per group.

\paragraph{Research/Model Dev:} Benjamin Jäger, Nick Erickson, Léo Grinsztajn, Felix
Birkel, Klemens Flöge, Lennart Purucker, Arthur Cahu, Vahid Balazadeh,
Jonas Kübler, Alan Arazi, Philipp Jund, Siyuan Guo, Anurag Garg, Jan Hendrik Metzen, Mihir Manium, Jake Robertson, David
Salinas, Andrej Tschalzev, Simon Bing, Adrian Hayler, Tobias Schröder,
Oscar Key.
\paragraph{Engineering/Platform:} Brendan Roof, Georg Grab, Simone
Alessi, Dominik Safaric.
\paragraph{Applied:} Eliott Kalfon, Philipp Singer.
\paragraph{GTM:} Clara Cornu, Diana Kriuchkova, Lilly Wehrhahn, Tuana
Çelik, Vitor Monteiro.
\paragraph{Ops/People/Legal:} Rylee Grace, Adèle Frankel, Kürşat Kaya,
Kristina Collins, Lydia Sidhoum, Marie Salmon, Tomás Pereda, Jerry Chen.
\paragraph{Scientific Advisors:} Madelon Hulsebos, Yann LeCun, Bernhard
Schölkopf.
\paragraph{Founders:} Frank Hutter, Sauraj Gambhir, Noah Hollmann.

\section{Acknowledgements}
\label{app:acknowledgements}

We acknowledge the EuroHPC Joint Undertaking for awarding this project access to the EuroHPC supercomputer LUMI, hosted by CSC (Finland) and the LUMI consortium through a EuroHPC Regular Access call.

We acknowledge EuroHPC JU for awarding the project ID EHPC-AIF-2025LS14-031 access to MareNostrum 5 ACC hosted by the Barcelona Supercomputing Center (BSC), Spain.

\begin{center}
\hfill
\begin{minipage}{0.3\textwidth}
\centering
\includegraphics[width=\textwidth]{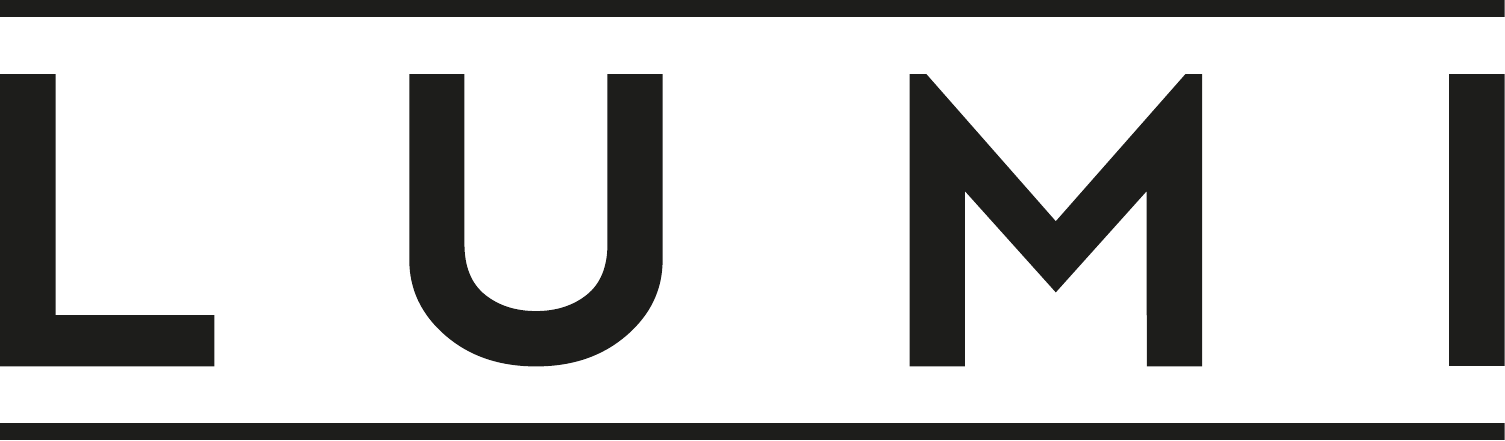}
\end{minipage}
\hfill
\begin{minipage}{0.35\textwidth}
\centering
\includegraphics[width=\textwidth]{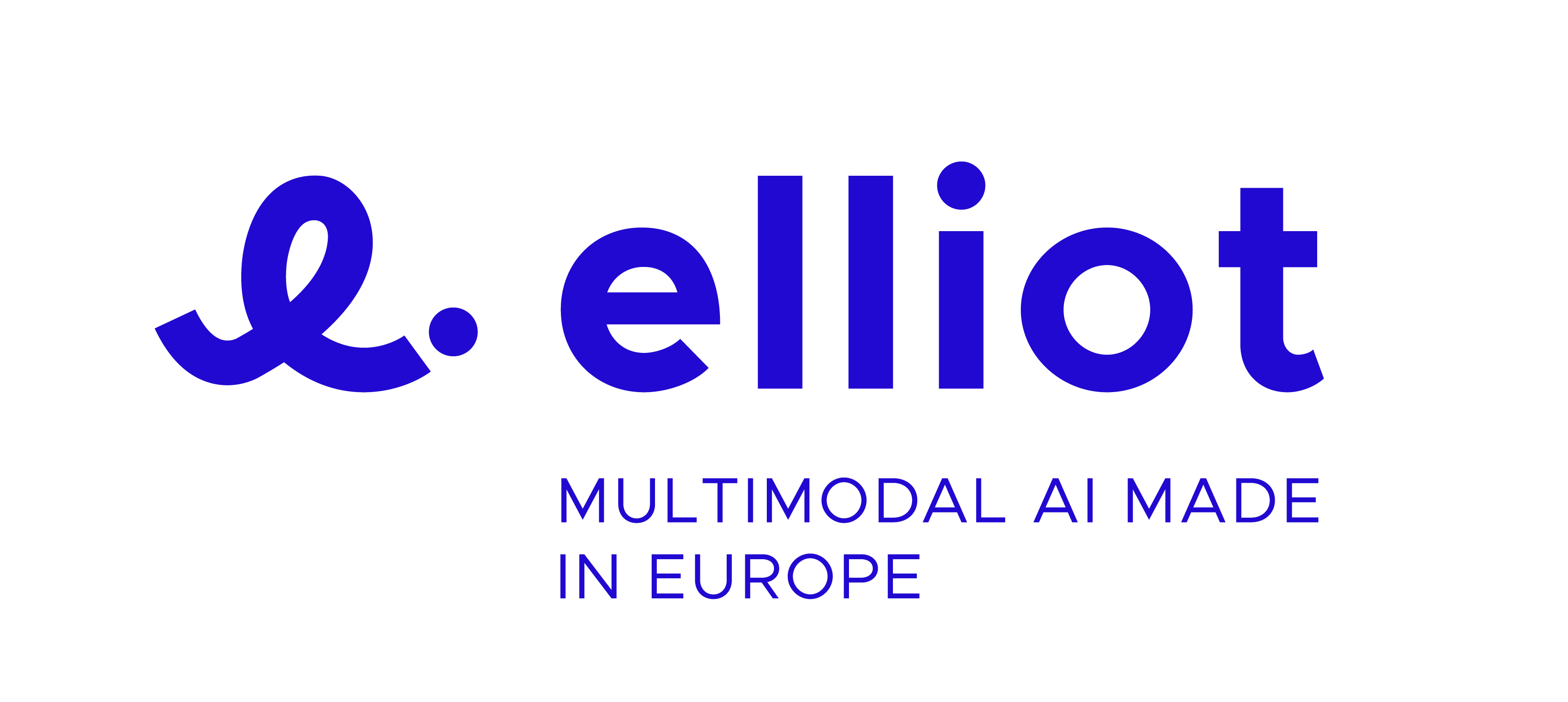}
\end{minipage}
\hfill ~
\\[1em]
\includegraphics[height=1.95cm]{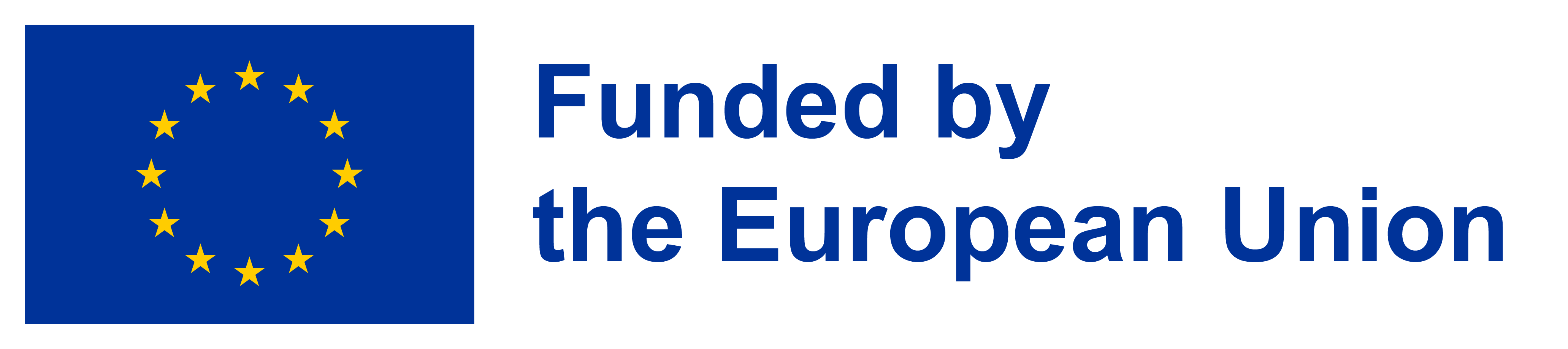}
\end{center}

%
Co-funded by the European Union. Views and opinions expressed are however those of the author(s) only and do not necessarily reflect those of the European Union or the European Commission. Neither the European Union nor the European Commission can be held responsible for them. This work was supported by the European Union’s Horizon Europe research and innovation programme under grant agreement No 101214398 (ELLIOT).

\section{More Experimental Results}\label{app:more_experimental_results}

\subsection{Additional details on TabArena}
\label{sec:app:tabarena}

\subsubsection{Tuning trajectories and pairwise win rates on TabArena}
\label{sec:app:tabarena_trajectories_winrate}

\begin{figure}[H]
  \centering
  \begin{minipage}[t]{0.48\linewidth}
    \centering
    \includegraphics[width=\linewidth]{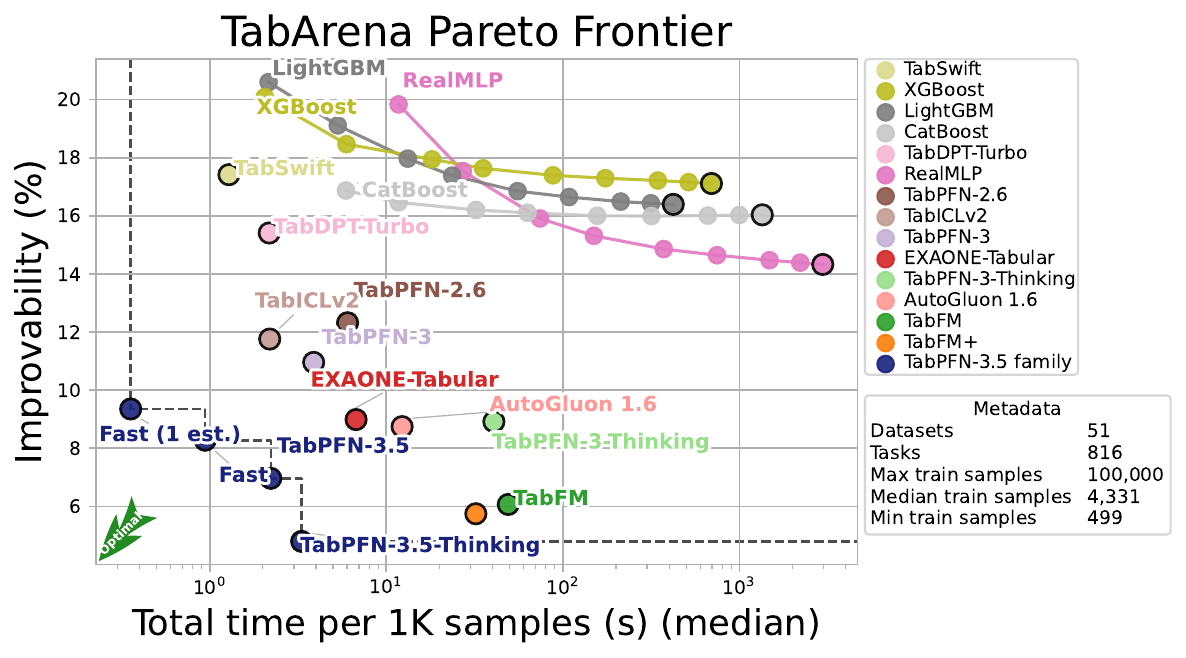}
    \caption{\textbf{Pareto frontier on TabArena}: improvability against the total training plus inference time per 1{,}000 rows, median over the 51 datasets, log scale. Connected points are the tuning trajectories of the classical baselines; single points are default configurations and the \ourmodel{} family. \Cref{sec:app:tabarena_timing} describes the time measurement.}
    \label{fig:tabarena_pareto_trajectories}
  \end{minipage}\hfill
  \begin{minipage}[t]{0.48\linewidth}
    \centering
    \includegraphics[width=\linewidth]{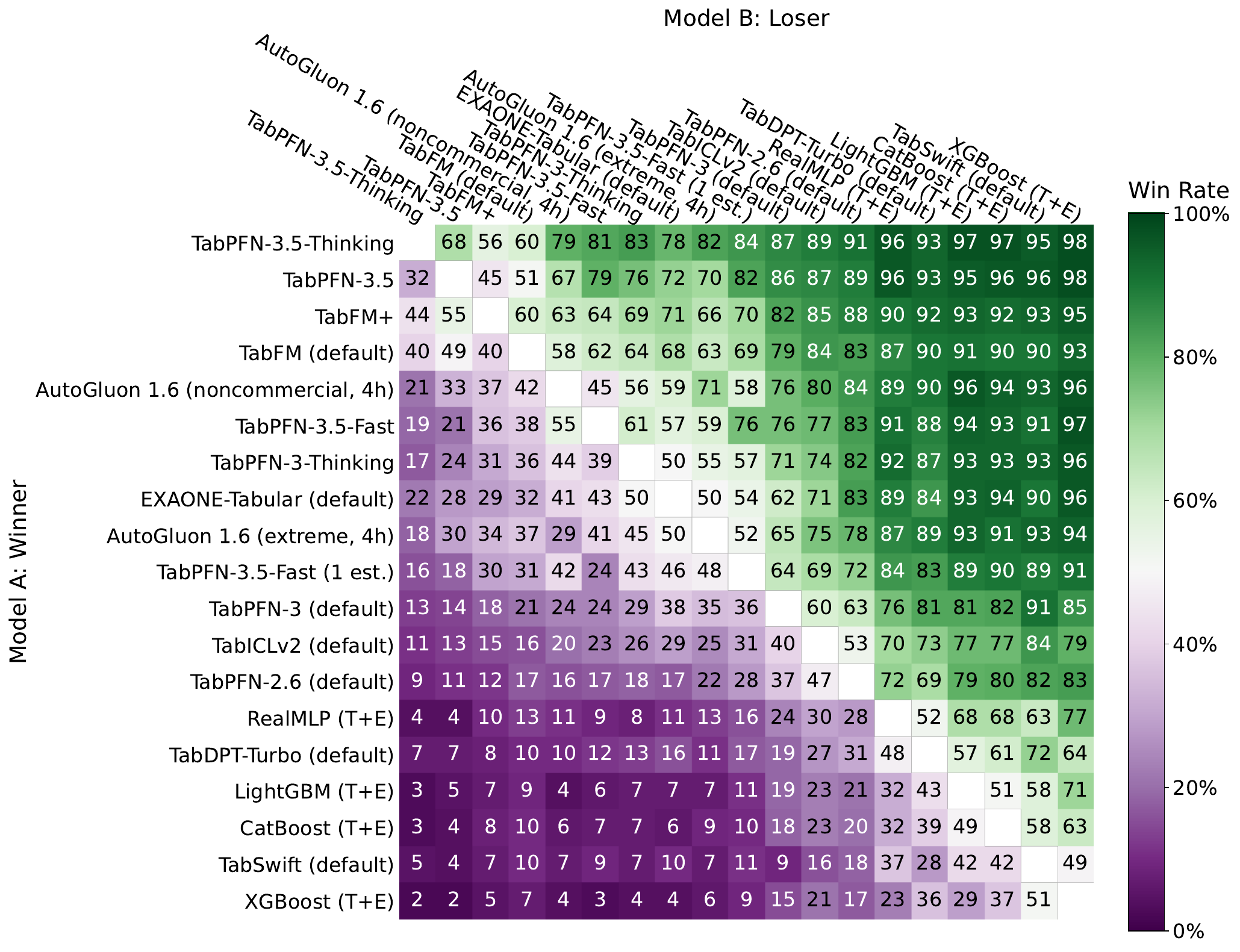}
    \caption{\textbf{Pairwise win rates on TabArena} for the strongest models on TabArena: the share of the 816 tasks on which the row method beats the column method. See \Cref{sec:app:tabarena_leaderboard_tables} for the full results.}
    \label{fig:tabarena_winrate}
  \end{minipage}
\end{figure}

\begin{figure}[H]
  \centering
  \begin{minipage}[t]{0.48\linewidth}
    \centering
    \includegraphics[width=\linewidth]{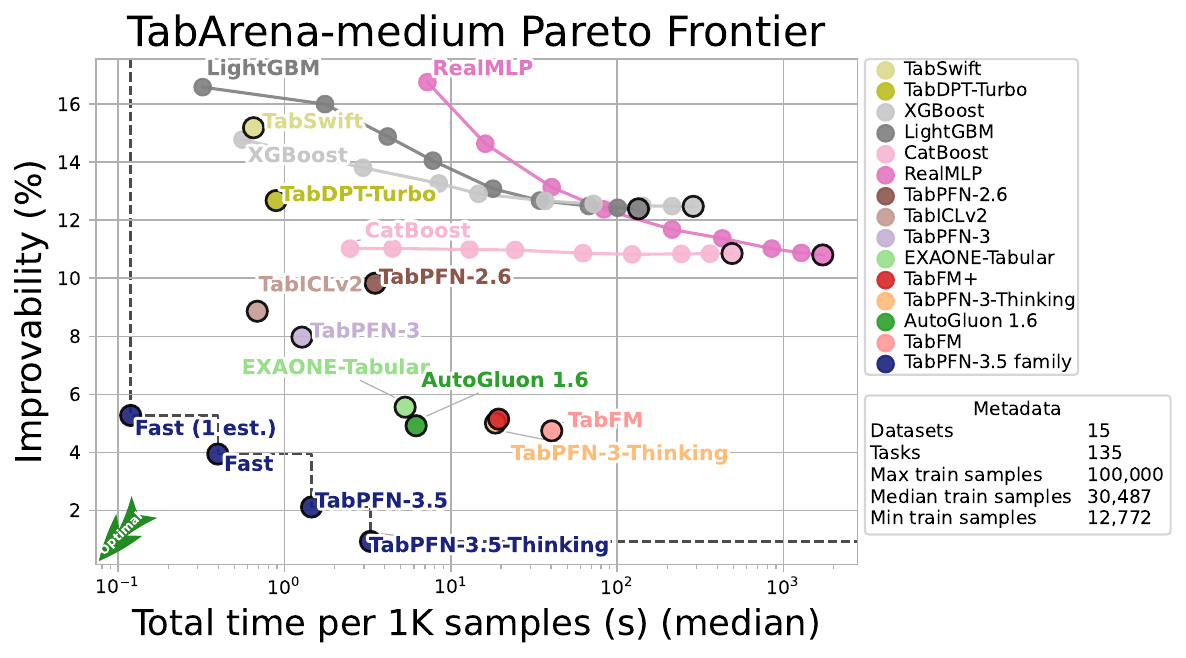}
    \caption{\textbf{Pareto frontier on TabArena-medium} (10k--100k training rows): improvability against the total training plus inference time per 1{,}000 rows, median over the 15 datasets, log scale. Same layout as \Cref{fig:tabarena_pareto_trajectories}.}
    \label{fig:tabarena_pareto_trajectories_medium}
  \end{minipage}\hfill
  \begin{minipage}[t]{0.48\linewidth}
    \centering
    \includegraphics[width=\linewidth]{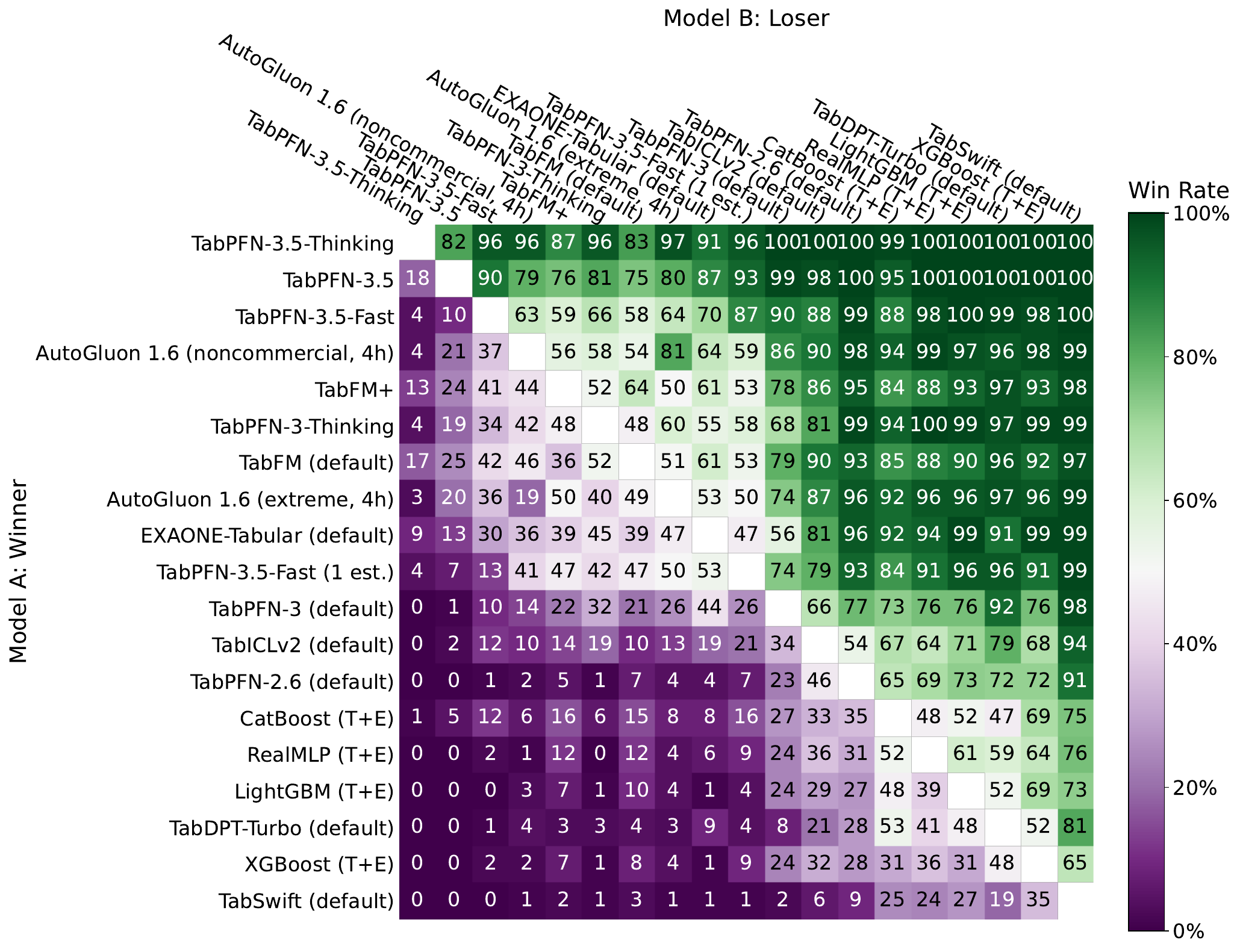}
    \caption{\textbf{Pairwise win rates on TabArena-medium} (10k--100k training rows) for the strongest models on TabArena, over the 135 tasks. See \Cref{sec:app:tabarena_leaderboard_tables} for the full results.}
    \label{fig:tabarena_winrate_medium}
  \end{minipage}
\end{figure}

\Cref{fig:tabarena_pareto_trajectories} places the \ourmodel{} family on TabArena's
improvability-against-time plane together with the tuning trajectories of the classical
baselines: each connected path shows a model's tuned-and-ensembled portfolio as the number of
random configurations grows from one to all, left to right, with the default configuration as
the left-most point. Improvability is how much a model would gain by switching to the best
model on each dataset. \Cref{fig:tabarena_winrate} gives the pairwise win rates of the same
methods over the 816 tasks. \Cref{fig:tabarena_pareto_trajectories_medium,fig:tabarena_winrate_medium}
repeat both views on TabArena-medium, the 15 datasets with 10k--100k training rows.

\subsubsection{TabArena leaderboard tables}
\label{sec:app:tabarena_leaderboard_tables}

We present the leaderboard tables for
\hyperref[tab:tabarena_table]{TabArena},
\hyperref[tab:tabarena_small_table]{TabArena-small},
\hyperref[tab:tabarena_medium_table]{TabArena-medium},
\hyperref[tab:tabarena_classification_table]{TabArena-classification}, and
\hyperref[tab:tabarena_regression_table]{TabArena-regression} below. Elo, win counts and
improvability are computed over the full TabArena field; the tables show the methods of the
report's figures. Train and predict times are medians over the datasets of each view of the
time per 1{,}000 rows, measured as described in \Cref{sec:app:tabarena_timing}.

In all five views, \ourmodelthinking{} ranks first in Elo. \ourmodel{} ranks second in every
view except TabArena-small, where TabFM+ places between the two, and \ourmodelfast{} reaches
the Elo of TabFM on the full benchmark at a fraction of its train and predict time.

\begin{table}[H]
\centering
\caption{\textbf{TabArena leaderboard using all 51 datasets with 816 total tasks.}}
\label{tab:tabarena_table}
\begin{tabular}{llccrr}
\toprule
\textbf{Model} & \textbf{Elo ($\uparrow$)} & \textbf{\#wins ($\uparrow$)} & \textbf{Improva-} & \textbf{Train time} & \textbf{Predict time} \\
 &  &  & \textbf{bility ($\downarrow$)} & \textbf{per 1K [s]} & \textbf{per 1K [s]} \\
\midrule
TabPFN-3.5-Thinking & \textcolor{gold}{\textbf{1910${}_{-82,+135}$}} & \textcolor{gold}{\textbf{13.6}} & \textcolor{gold}{\textbf{4.8\%}} & 2.00 & 1.47 \\
TabPFN-3.5 & \textcolor{silver}{\textbf{1866${}_{-65,+87}$}} & 3.3 & 7.0\% & 1.83 & 0.44 \\
TabFM+ & \textcolor{bronze}{\textbf{1823${}_{-81,+104}$}} & \textcolor{silver}{\textbf{8.4}} & \textcolor{silver}{\textbf{5.8\%}} & 9.24 & 7.63 \\
AutoGluon 1.6 (noncommercial, 4h) & 1786${}_{-57,+97}$ & 0.8 & 8.2\% & 14.48 & 0.82 \\
TabPFN-3.5-Fast & 1780${}_{-59,+83}$ & 0.7 & 8.3\% & 0.78 & 0.16 \\
TabFM (D) & 1780${}_{-95,+99}$ & \textcolor{bronze}{\textbf{8.0}} & \textcolor{bronze}{\textbf{6.1\%}} & 17.11 & 7.00 \\
TabPFN-3-Thinking & 1748${}_{-50,+72}$ & 1.0 & 8.9\% & 36.84 & 2.43 \\
EXAONE-Tabular (D) & 1747${}_{-54,+70}$ & 2.3 & 9.0\% & 1.94 & 0.59 \\
AutoGluon 1.6 (extreme, 4h) & 1737${}_{-51,+86}$ & 0.7 & 8.8\% & 11.58 & 0.61 \\
TabPFN-3.5-Fast (1 est.) & 1703${}_{-54,+66}$ & 1.0 & 9.4\% & 0.30 & 0.05 \\
TabPFN-3 (D) & 1631${}_{-48,+66}$ & 0.2 & 11.0\% & 1.33 & 0.38 \\
TabPFN-2.6 (D) & 1579${}_{-43,+57}$ & 0.1 & 12.3\% & 5.48 & 0.55 \\
TabICLv2 (D) & 1565${}_{-54,+64}$ & 0.3 & 11.8\% & 0.66 & 0.13 \\
RealMLP (T+E) & 1474${}_{-41,+43}$ & 0.2 & 14.3\% & 2950.72 & 11.97 \\
TabDPT-Turbo (D) & 1434${}_{-45,+49}$ & 0.3 & 15.4\% & 0.65 & 0.18 \\
LightGBM (T+E) & 1404${}_{-28,+28}$ & 0.0 & 16.4\% & 417.05 & 2.64 \\
RealMLP (T) & 1401${}_{-47,+38}$ & 0.1 & 15.6\% & 2950.72 & 0.66 \\
CatBoost (T+E) & 1394${}_{-33,+33}$ & 0.1 & 16.1\% & 1346.21 & 0.34 \\
CatBoost (T) & 1382${}_{-33,+30}$ & 0.0 & 16.3\% & 1346.21 & 0.04 \\
LightGBM (T) & 1362${}_{-26,+25}$ & 0.0 & 17.1\% & 417.05 & 0.33 \\
CatBoost (D) & 1355${}_{-39,+39}$ & 0.0 & 16.9\% & 5.88 & 0.02 \\
XGBoost (T+E) & 1352${}_{-30,+28}$ & 0.0 & 17.1\% & 693.49 & 1.69 \\
TabSwift (D) & 1333${}_{-48,+60}$ & 0.0 & 17.4\% & 0.55 & 0.07 \\
XGBoost (T) & 1329${}_{-28,+29}$ & 0.0 & 17.4\% & 693.49 & 0.31 \\
RealMLP (D) & 1217${}_{-38,+33}$ & 0.1 & 19.8\% & 10.06 & 1.69 \\
XGBoost (D) & 1203${}_{-35,+34}$ & 0.0 & 20.1\% & 1.94 & 0.12 \\
LightGBM (D) & 1178${}_{-29,+32}$ & 0.0 & 20.6\% & 1.96 & 0.14 \\
\bottomrule
\end{tabular}

\end{table}

\begin{table}[H]
\centering
\caption{\textbf{TabArena-small leaderboard on the 36 smallest datasets in TabArena}, with 500--10k training samples, evaluated on the full 681 tasks.}
\label{tab:tabarena_small_table}
\begin{tabular}{llccrr}
\toprule
\textbf{Model} & \textbf{Elo ($\uparrow$)} & \textbf{\#wins ($\uparrow$)} & \textbf{Improva-} & \textbf{Train time} & \textbf{Predict time} \\
 &  &  & \textbf{bility ($\downarrow$)} & \textbf{per 1K [s]} & \textbf{per 1K [s]} \\
\midrule
TabPFN-3.5-Thinking & \textcolor{gold}{\textbf{1824${}_{-68,+119}$}} & \textcolor{bronze}{\textbf{5.3}} & \textcolor{silver}{\textbf{6.4\%}} & 1.52 & 2.14 \\
TabFM+ & \textcolor{silver}{\textbf{1806${}_{-80,+126}$}} & \textcolor{gold}{\textbf{7.5}} & \textcolor{gold}{\textbf{6.0\%}} & 10.68 & 6.54 \\
TabPFN-3.5 & \textcolor{bronze}{\textbf{1796${}_{-57,+92}$}} & 2.0 & 9.0\% & 2.11 & 0.44 \\
TabFM (D) & 1753${}_{-85,+142}$ & \textcolor{silver}{\textbf{6.3}} & \textcolor{bronze}{\textbf{6.6\%}} & 15.08 & 5.38 \\
AutoGluon 1.6 (noncommercial, 4h) & 1734${}_{-54,+72}$ & 0.4 & 9.8\% & 19.64 & 0.92 \\
TabPFN-3.5-Fast & 1723${}_{-51,+94}$ & 0.6 & 10.1\% & 0.89 & 0.16 \\
EXAONE-Tabular (D) & 1708${}_{-59,+71}$ & 1.4 & 10.4\% & 1.85 & 0.55 \\
TabPFN-3-Thinking & 1693${}_{-45,+81}$ & 0.6 & 10.6\% & 51.76 & 2.69 \\
AutoGluon 1.6 (extreme, 4h) & 1689${}_{-47,+55}$ & 0.5 & 10.4\% & 15.56 & 0.63 \\
TabPFN-3.5-Fast (1 est.) & 1656${}_{-55,+82}$ & 0.6 & 11.1\% & 0.45 & 0.05 \\
TabPFN-3 (D) & 1611${}_{-51,+68}$ & 0.2 & 12.2\% & 1.97 & 0.40 \\
TabPFN-2.6 (D) & 1572${}_{-38,+57}$ & 0.1 & 13.4\% & 7.03 & 0.55 \\
TabICLv2 (D) & 1553${}_{-74,+90}$ & 0.3 & 13.0\% & 0.78 & 0.14 \\
RealMLP (T+E) & 1455${}_{-43,+49}$ & 0.2 & 15.8\% & 3771.85 & 21.90 \\
TabDPT-Turbo (D) & 1430${}_{-52,+68}$ & 0.3 & 16.6\% & 0.86 & 0.19 \\
RealMLP (T) & 1382${}_{-42,+46}$ & 0.1 & 17.2\% & 3771.85 & 1.78 \\
LightGBM (T+E) & 1372${}_{-33,+28}$ & 0.0 & 18.1\% & 892.49 & 2.57 \\
TabSwift (D) & 1350${}_{-51,+58}$ & 0.0 & 18.4\% & 0.57 & 0.06 \\
CatBoost (T+E) & 1349${}_{-38,+35}$ & 0.1 & 18.2\% & 1739.17 & 0.40 \\
LightGBM (T) & 1338${}_{-32,+29}$ & 0.0 & 18.5\% & 892.49 & 0.35 \\
CatBoost (T) & 1337${}_{-33,+31}$ & 0.0 & 18.4\% & 1739.17 & 0.05 \\
XGBoost (T+E) & 1311${}_{-35,+33}$ & 0.0 & 19.1\% & 884.48 & 2.37 \\
CatBoost (D) & 1305${}_{-38,+34}$ & 0.0 & 19.3\% & 8.46 & 0.05 \\
XGBoost (T) & 1294${}_{-40,+32}$ & 0.0 & 19.3\% & 884.48 & 0.39 \\
RealMLP (D) & 1212${}_{-43,+37}$ & 0.1 & 21.1\% & 15.69 & 4.69 \\
LightGBM (D) & 1158${}_{-44,+43}$ & 0.0 & 22.3\% & 3.61 & 0.17 \\
XGBoost (D) & 1157${}_{-38,+25}$ & 0.0 & 22.3\% & 3.29 & 0.25 \\
\bottomrule
\end{tabular}

\end{table}

\begin{table}[H]
\centering
\caption{\textbf{TabArena-medium leaderboard on the 15 largest datasets in TabArena}, with 10k--100k training samples, evaluated on the full 135 tasks with 9 splits per dataset.}
\label{tab:tabarena_medium_table}
\begin{tabular}{llccrr}
\toprule
\textbf{Model} & \textbf{Elo ($\uparrow$)} & \textbf{\#wins ($\uparrow$)} & \textbf{Improva-} & \textbf{Train time} & \textbf{Predict time} \\
 &  &  & \textbf{bility ($\downarrow$)} & \textbf{per 1K [s]} & \textbf{per 1K [s]} \\
\midrule
TabPFN-3.5-Thinking & \textcolor{gold}{\textbf{2494${}_{-114,+232}$}} & \textcolor{gold}{\textbf{8.3}} & \textcolor{gold}{\textbf{0.9\%}} & 2.33 & 0.99 \\
TabPFN-3.5 & \textcolor{silver}{\textbf{2293${}_{-92,+225}$}} & \textcolor{bronze}{\textbf{1.2}} & \textcolor{silver}{\textbf{2.1\%}} & 1.06 & 0.40 \\
TabPFN-3.5-Fast & \textcolor{bronze}{\textbf{2117${}_{-105,+213}$}} & 0.1 & \textcolor{bronze}{\textbf{3.9\%}} & 0.30 & 0.10 \\
AutoGluon 1.6 (noncommercial, 4h) & 2097${}_{-78,+140}$ & 0.4 & 4.3\% & 7.01 & 0.63 \\
TabPFN-3-Thinking & 2067${}_{-56,+99}$ & 0.4 & 5.0\% & 15.90 & 2.12 \\
AutoGluon 1.6 (extreme, 4h) & 2022${}_{-64,+105}$ & 0.2 & 4.9\% & 5.21 & 0.52 \\
TabFM+ & 2003${}_{-108,+251}$ & 0.9 & 5.1\% & 3.53 & 13.80 \\
EXAONE-Tabular (D) & 1995${}_{-73,+137}$ & 0.9 & 5.6\% & 3.93 & 1.21 \\
TabFM (D) & 1986${}_{-129,+270}$ & \textcolor{silver}{\textbf{1.7}} & 4.8\% & 26.32 & 13.12 \\
TabPFN-3.5-Fast (1 est.) & 1980${}_{-87,+152}$ & 0.3 & 5.3\% & 0.09 & 0.03 \\
TabPFN-3 (D) & 1788${}_{-129,+194}$ & 0.0 & 8.0\% & 0.58 & 0.21 \\
TabPFN-2.6 (D) & 1691${}_{-61,+108}$ & 0.0 & 9.8\% & 2.76 & 0.70 \\
TabICLv2 (D) & 1689${}_{-103,+168}$ & 0.0 & 8.9\% & 0.41 & 0.12 \\
RealMLP (T+E) & 1608${}_{-95,+91}$ & 0.0 & 10.8\% & 1719.82 & 1.67 \\
CatBoost (T+E) & 1600${}_{-77,+99}$ & 0.0 & 10.9\% & 489.93 & 0.25 \\
CatBoost (T) & 1586${}_{-84,+100}$ & 0.0 & 11.1\% & 489.93 & 0.02 \\
LightGBM (T+E) & 1572${}_{-51,+59}$ & 0.0 & 12.4\% & 131.56 & 2.64 \\
CatBoost (D) & 1572${}_{-109,+114}$ & 0.0 & 11.0\% & 2.47 & 0.01 \\
XGBoost (T+E) & 1534${}_{-61,+82}$ & 0.0 & 12.5\% & 282.13 & 0.56 \\
RealMLP (T) & 1522${}_{-85,+102}$ & 0.0 & 11.9\% & 1719.82 & 0.08 \\
TabDPT-Turbo (D) & 1514${}_{-134,+141}$ & 0.0 & 12.7\% & 0.46 & 0.18 \\
LightGBM (T) & 1490${}_{-49,+60}$ & 0.0 & 13.5\% & 131.56 & 0.13 \\
XGBoost (T) & 1488${}_{-58,+60}$ & 0.0 & 13.0\% & 282.13 & 0.07 \\
XGBoost (D) & 1366${}_{-99,+90}$ & 0.0 & 14.8\% & 0.49 & 0.05 \\
TabSwift (D) & 1337${}_{-103,+110}$ & 0.0 & 15.2\% & 0.42 & 0.13 \\
RealMLP (D) & 1264${}_{-80,+76}$ & 0.0 & 16.8\% & 6.75 & 0.23 \\
LightGBM (D) & 1260${}_{-60,+46}$ & 0.0 & 16.6\% & 0.29 & 0.04 \\
\bottomrule
\end{tabular}

\end{table}

\begin{table}[H]
\centering
\caption{\textbf{TabArena-classification leaderboard on the 38 classification datasets in TabArena}, evaluated on the full 594 tasks.}
\label{tab:tabarena_classification_table}
\begin{tabular}{llccrr}
\toprule
\textbf{Model} & \textbf{Elo ($\uparrow$)} & \textbf{\#wins ($\uparrow$)} & \textbf{Improva-} & \textbf{Train time} & \textbf{Predict time} \\
 &  &  & \textbf{bility ($\downarrow$)} & \textbf{per 1K [s]} & \textbf{per 1K [s]} \\
\midrule
TabPFN-3.5-Thinking & \textcolor{gold}{\textbf{1872${}_{-88,+131}$}} & \textcolor{gold}{\textbf{7.7}} & \textcolor{gold}{\textbf{6.2\%}} & 2.37 & 1.54 \\
TabPFN-3.5 & \textcolor{silver}{\textbf{1845${}_{-71,+106}$}} & 2.0 & 8.9\% & 1.86 & 0.49 \\
TabFM+ & \textcolor{bronze}{\textbf{1803${}_{-89,+117}$}} & \textcolor{bronze}{\textbf{7.3}} & \textcolor{silver}{\textbf{7.2\%}} & 10.68 & 7.80 \\
TabPFN-3.5-Fast & 1784${}_{-69,+102}$ & 0.5 & 10.0\% & 0.78 & 0.16 \\
TabFM (D) & 1768${}_{-114,+121}$ & \textcolor{silver}{\textbf{7.5}} & \textcolor{silver}{\textbf{7.2\%}} & 18.58 & 7.06 \\
AutoGluon 1.6 (noncommercial, 4h) & 1757${}_{-53,+80}$ & 0.2 & 10.2\% & 12.89 & 0.79 \\
EXAONE-Tabular (D) & 1752${}_{-58,+79}$ & 2.1 & 10.8\% & 1.58 & 0.51 \\
TabPFN-3-Thinking & 1744${}_{-51,+88}$ & 0.8 & 11.0\% & 35.78 & 2.42 \\
TabPFN-3.5-Fast (1 est.) & 1710${}_{-53,+97}$ & 0.8 & 11.2\% & 0.32 & 0.05 \\
AutoGluon 1.6 (extreme, 4h) & 1709${}_{-42,+61}$ & 0.2 & 10.9\% & 9.70 & 0.57 \\
TabPFN-3 (D) & 1625${}_{-65,+71}$ & 0.1 & 13.5\% & 1.32 & 0.38 \\
TabPFN-2.6 (D) & 1574${}_{-53,+55}$ & 0.1 & 14.8\% & 5.17 & 0.54 \\
TabICLv2 (D) & 1570${}_{-64,+73}$ & 0.3 & 14.1\% & 0.68 & 0.14 \\
RealMLP (T+E) & 1461${}_{-36,+51}$ & 0.1 & 17.1\% & 2879.46 & 12.48 \\
TabDPT-Turbo (D) & 1419${}_{-54,+71}$ & 0.2 & 18.5\% & 0.69 & 0.19 \\
LightGBM (T+E) & 1411${}_{-30,+44}$ & 0.0 & 18.8\% & 382.09 & 1.49 \\
CatBoost (T+E) & 1395${}_{-42,+47}$ & 0.1 & 18.5\% & 1107.00 & 0.33 \\
RealMLP (T) & 1390${}_{-42,+49}$ & 0.1 & 18.5\% & 2879.46 & 0.60 \\
CatBoost (T) & 1386${}_{-40,+45}$ & 0.0 & 18.7\% & 1107.00 & 0.04 \\
CatBoost (D) & 1372${}_{-44,+43}$ & 0.0 & 19.0\% & 5.41 & 0.03 \\
LightGBM (T) & 1370${}_{-30,+36}$ & 0.0 & 19.5\% & 382.09 & 0.25 \\
XGBoost (T+E) & 1359${}_{-43,+39}$ & 0.0 & 19.6\% & 685.87 & 1.45 \\
TabSwift (D) & 1338${}_{-58,+55}$ & 0.0 & 20.5\% & 0.56 & 0.06 \\
XGBoost (T) & 1334${}_{-38,+35}$ & 0.0 & 19.9\% & 685.87 & 0.21 \\
RealMLP (D) & 1230${}_{-39,+41}$ & 0.1 & 22.7\% & 10.48 & 1.71 \\
XGBoost (D) & 1220${}_{-47,+40}$ & 0.0 & 22.5\% & 1.77 & 0.12 \\
LightGBM (D) & 1182${}_{-41,+49}$ & 0.0 & 23.4\% & 1.79 & 0.12 \\
\bottomrule
\end{tabular}

\end{table}

\begin{table}[H]
\centering
\caption{\textbf{TabArena-regression leaderboard on the 13 regression datasets in TabArena}, evaluated on the full 222 tasks.}
\label{tab:tabarena_regression_table}
\begin{tabular}{llccrr}
\toprule
\textbf{Model} & \textbf{Elo ($\uparrow$)} & \textbf{\#wins ($\uparrow$)} & \textbf{Improva-} & \textbf{Train time} & \textbf{Predict time} \\
 &  &  & \textbf{bility ($\downarrow$)} & \textbf{per 1K [s]} & \textbf{per 1K [s]} \\
\midrule
TabPFN-3.5-Thinking & \textcolor{gold}{\textbf{2250${}_{-135,+241}$}} & \textcolor{gold}{\textbf{5.9}} & \textcolor{gold}{\textbf{0.8\%}} & 1.54 & 0.99 \\
TabPFN-3.5 & \textcolor{silver}{\textbf{2103${}_{-130,+245}$}} & \textcolor{silver}{\textbf{1.3}} & \textcolor{silver}{\textbf{1.4\%}} & 1.55 & 0.37 \\
AutoGluon 1.6 (noncommercial, 4h) & \textcolor{bronze}{\textbf{2056${}_{-141,+190}$}} & 0.6 & 2.1\% & 22.20 & 1.73 \\
TabFM+ & 2055${}_{-130,+186}$ & \textcolor{bronze}{\textbf{1.1}} & \textcolor{bronze}{\textbf{1.7\%}} & 7.28 & 7.19 \\
AutoGluon 1.6 (extreme, 4h) & 2002${}_{-107,+161}$ & 0.5 & 2.6\% & 19.36 & 1.63 \\
TabFM (D) & 1975${}_{-85,+161}$ & 0.5 & 2.8\% & 14.60 & 6.47 \\
TabPFN-3.5-Fast & 1912${}_{-94,+172}$ & 0.2 & 3.3\% & 0.66 & 0.13 \\
TabPFN-3-Thinking & 1907${}_{-130,+158}$ & 0.3 & 2.8\% & 37.75 & 2.43 \\
EXAONE-Tabular (D) & 1868${}_{-96,+131}$ & 0.2 & 3.9\% & 2.47 & 0.78 \\
TabPFN-3.5-Fast (1 est.) & 1816${}_{-106,+175}$ & 0.1 & 3.9\% & 0.20 & 0.05 \\
TabPFN-3 (D) & 1784${}_{-123,+205}$ & 0.0 & 3.6\% & 1.33 & 0.38 \\
TabPFN-2.6 (D) & 1723${}_{-47,+95}$ & 0.0 & 5.1\% & 8.52 & 0.70 \\
TabICLv2 (D) & 1672${}_{-134,+213}$ & 0.0 & 5.1\% & 0.64 & 0.11 \\
RealMLP (T+E) & 1636${}_{-56,+99}$ & 0.0 & 6.1\% & 3995.01 & 10.05 \\
TabDPT-Turbo (D) & 1594${}_{-88,+175}$ & 0.1 & 6.3\% & 0.49 & 0.18 \\
RealMLP (T) & 1540${}_{-68,+98}$ & 0.0 & 7.1\% & 3995.01 & 0.84 \\
CatBoost (T+E) & 1479${}_{-57,+78}$ & 0.0 & 9.0\% & 2229.21 & 0.38 \\
LightGBM (T+E) & 1472${}_{-68,+71}$ & 0.0 & 9.4\% & 700.19 & 9.32 \\
CatBoost (T) & 1456${}_{-58,+83}$ & 0.0 & 9.1\% & 2229.21 & 0.06 \\
LightGBM (T) & 1411${}_{-74,+75}$ & 0.0 & 10.0\% & 700.19 & 0.97 \\
XGBoost (T+E) & 1402${}_{-39,+49}$ & 0.0 & 10.0\% & 834.93 & 2.61 \\
TabSwift (D) & 1395${}_{-110,+146}$ & 0.0 & 8.3\% & 0.49 & 0.07 \\
XGBoost (T) & 1381${}_{-49,+58}$ & 0.0 & 10.1\% & 834.93 & 0.39 \\
CatBoost (D) & 1375${}_{-83,+83}$ & 0.0 & 10.6\% & 8.73 & 0.02 \\
RealMLP (D) & 1221${}_{-87,+102}$ & 0.0 & 11.5\% & 8.90 & 1.64 \\
LightGBM (D) & 1201${}_{-44,+29}$ & 0.0 & 12.4\% & 2.11 & 0.27 \\
XGBoost (D) & 1182${}_{-82,+83}$ & 0.0 & 13.0\% & 2.24 & 0.24 \\
\bottomrule
\end{tabular}

\end{table}

\subsubsection{Additional results on TabArena}
\begin{figure}
    \centering
    \includegraphics[width=\linewidth]{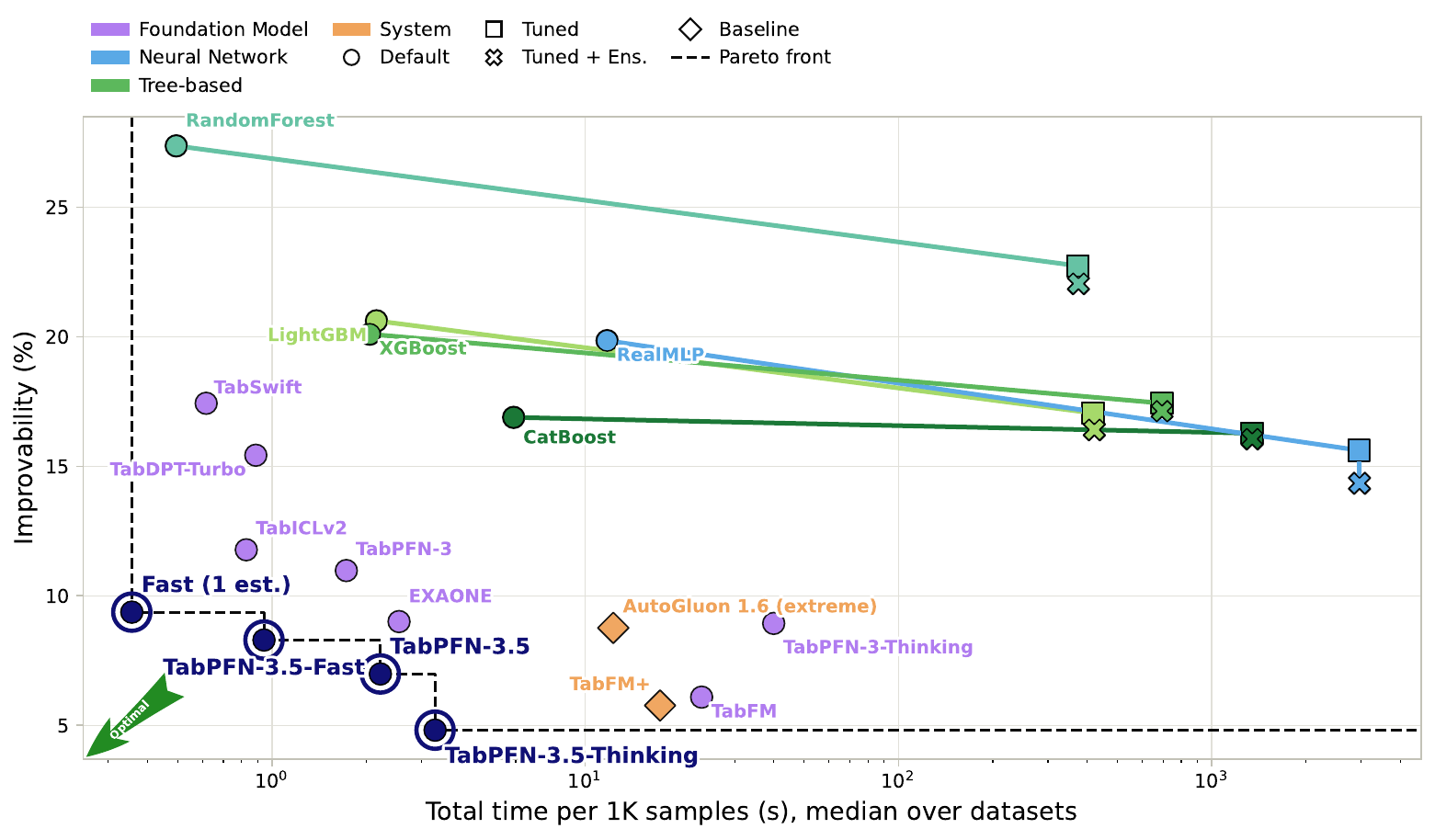}
    \caption{\textbf{TabArena improvability against total time.} Lower is better. Same methods and time axis as \Cref{fig:ta_pareto_elo}; \Cref{sec:app:tabarena_timing} describes the measurement.}
    \label{fig:ta_pareto_improv}
\end{figure}

\begin{figure}
    \centering
    \includegraphics[width=\linewidth]{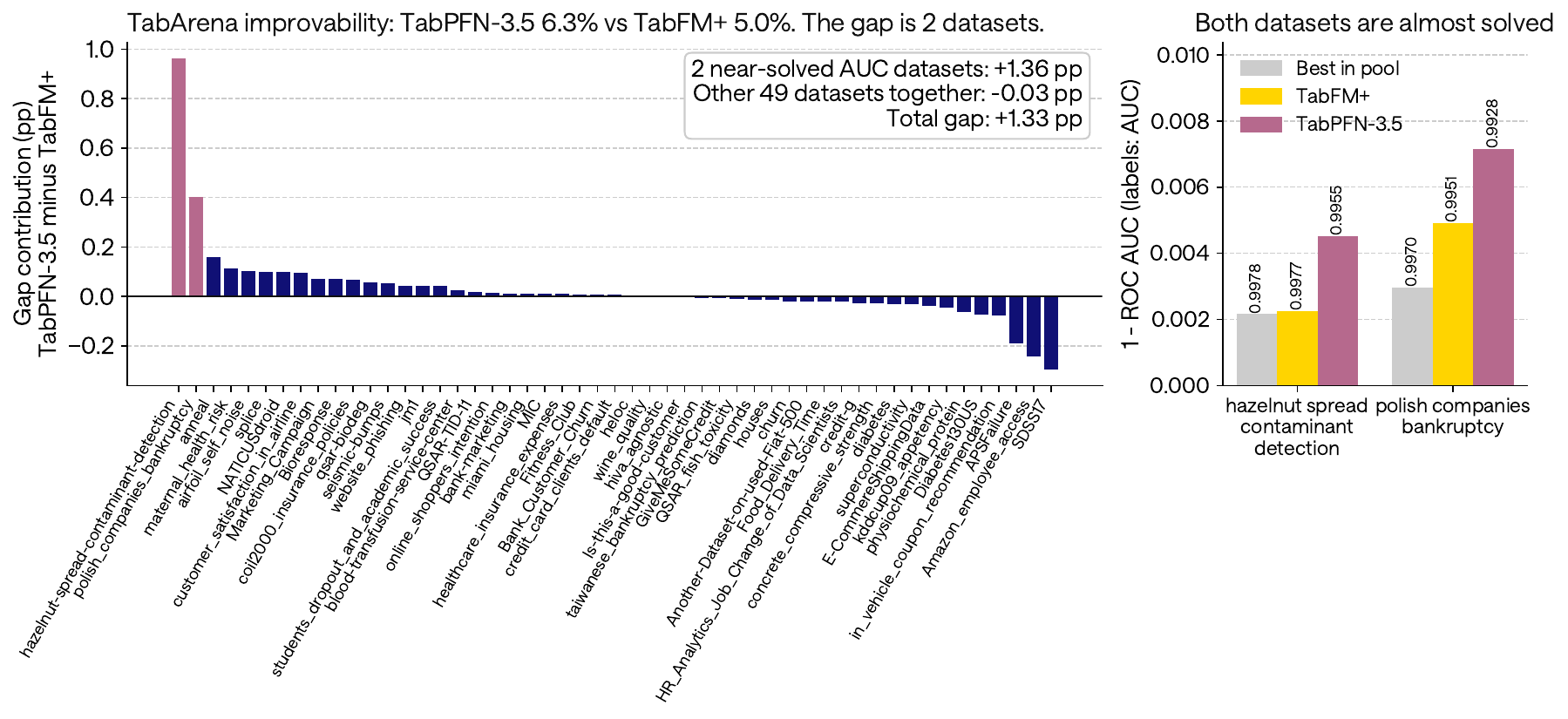}
    \caption{Improvability vs. TabFM+, we can see that the improvability gap in the final Pareto Frontier boils down to \texttt{hazelnut-spread-contaminant-detection} and \texttt{polish-companies-bankruptcy}, both of which are almost perfectly solved, with ROC AUC exceeding 0.997.}
    \label{fig:per_data_improv}
\end{figure}

\paragraph{What drives improvability differences?} \Cref{fig:ta_pareto_improv} shows the improvability as a function of total fit-and-predict time.
Although TabFM and TabFM+ achieve slightly higher total improvability than \ourmodel{}, the per-dataset comparison in \cref{fig:per_data_improv} shows that the remaining gap is dominated by two datasets: \texttt{hazelnut-spread-contaminant-detection} and \texttt{polish-companies-bankruptcy}. Both are already nearly perfectly solved, with ROC AUC above 0.997, which leads to improvability inflating the small absolute residual differences.

\subsubsection{How the time axis of the TabArena Pareto frontier is computed}
\label{sec:app:tabarena_timing}
\paragraph{Protocol.} All times in \Cref{fig:ta_pareto_elo} follow TabArena's default protocol \citep{erickson2025tabarena}: every method is fitted as an 8-fold bagging ensemble with a final refit on the full training set, in a fresh process per split, on one node with 24 CPU cores and one NVIDIA RTX Pro 6000 GPU. The fit time of a split is the wall-clock time of the whole bagged fit; the prediction time is the wall-clock time of the refitted model predicting the test third of the dataset. While the 8-fold bagging is not used for TFMs in general, as they are refitted on the full training set during inference, we keep it to make the comparison fairer to other non-TFM models, which could also be fitted without bagging (although with weaker performance). Models which are not default or tuning configurations of foundation models, neural networks or trees (\ourmodelthinking{}, \ourmodelthree{}-Thinking and systems like TabFM+ and AutoGluon~1.6) are instead called once per split, so their fit time might look a bit better than that of the bagged configurations; this is why TabFM+ shows a lower total time than TabFM in \Cref{fig:ta_pareto_elo}.

\paragraph{Normalization.} Per split, fit time is expressed per 1{,}000 training rows and prediction time per 1{,}000 test rows, averaged over folds per dataset and reported as the median over the 51 datasets, as in the TabArena leaderboard \citep{erickson2025tabarena}. The total time of a split is the sum of the two per-1{,}000-row rates, the definition TabArena's code uses for its combined time axis.

\paragraph{Re-timing of the foundation models.} The arena's wrappers for tabular foundation models carried overheads that are not part of running the model and were often larger than actual train times for small datasets, making it hard to compare different TFMs' speeds: the network was rebuilt from its checkpoint and copied to the GPU for each of the nine bagging children, the child models were saved with the full network weights inside, and the library import and the first forward pass of the process fell inside the timed fit. We removed these in the wrappers (one network per process shared across the children, no weights in the saved children, and an untimed warm-up on fixed synthetic data) and re-timed \ourmodel{}, \ourmodelfast{}, TabFM, TabFM+, EXAONE-Tabular, \ourmodelthree{}, TabICLv2, TabDPT-Turbo \citep{hosseinzadeh2026tabdpt} and TabSwift on the first repeat of every dataset (3 folds, 153 splits), with the Python environment on the node's local disk. Predictions are unchanged, so the Elo scores are those of the full 816-split evaluations. The tree-based and neural baselines use the arena's own measurements. \ourmodelthree{}-Thinking keeps the Elo and time of its published run from May 2026, which predates these fixes, so its time is an upper bound.

\subsection{Additional results on BeyondArena}
\label{sec:app:beyondarena}

\subsubsection{Radar view of the BeyondArena slices}
\label{sec:app:beyondarena:radar}
\begin{figure}[H]
    \centering
    \begin{subfigure}[t]{0.49\linewidth}
        \centering
        \includegraphics[width=\linewidth]{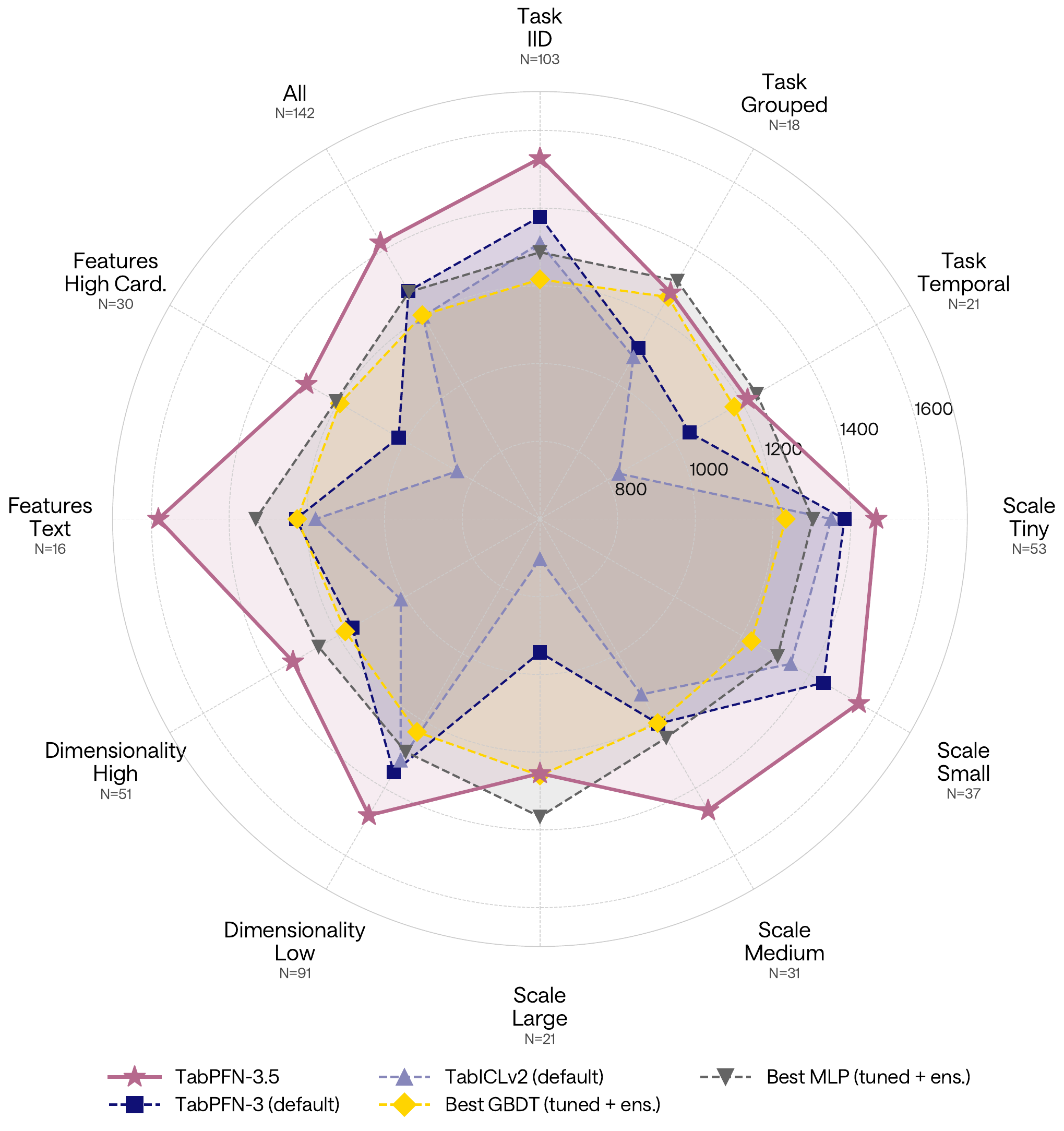}
        \caption{Full BeyondArena core.}
        \label{fig:ba-radar-core}
    \end{subfigure}
    \hfill
    \begin{subfigure}[t]{0.49\linewidth}
        \centering
        \includegraphics[width=\linewidth]{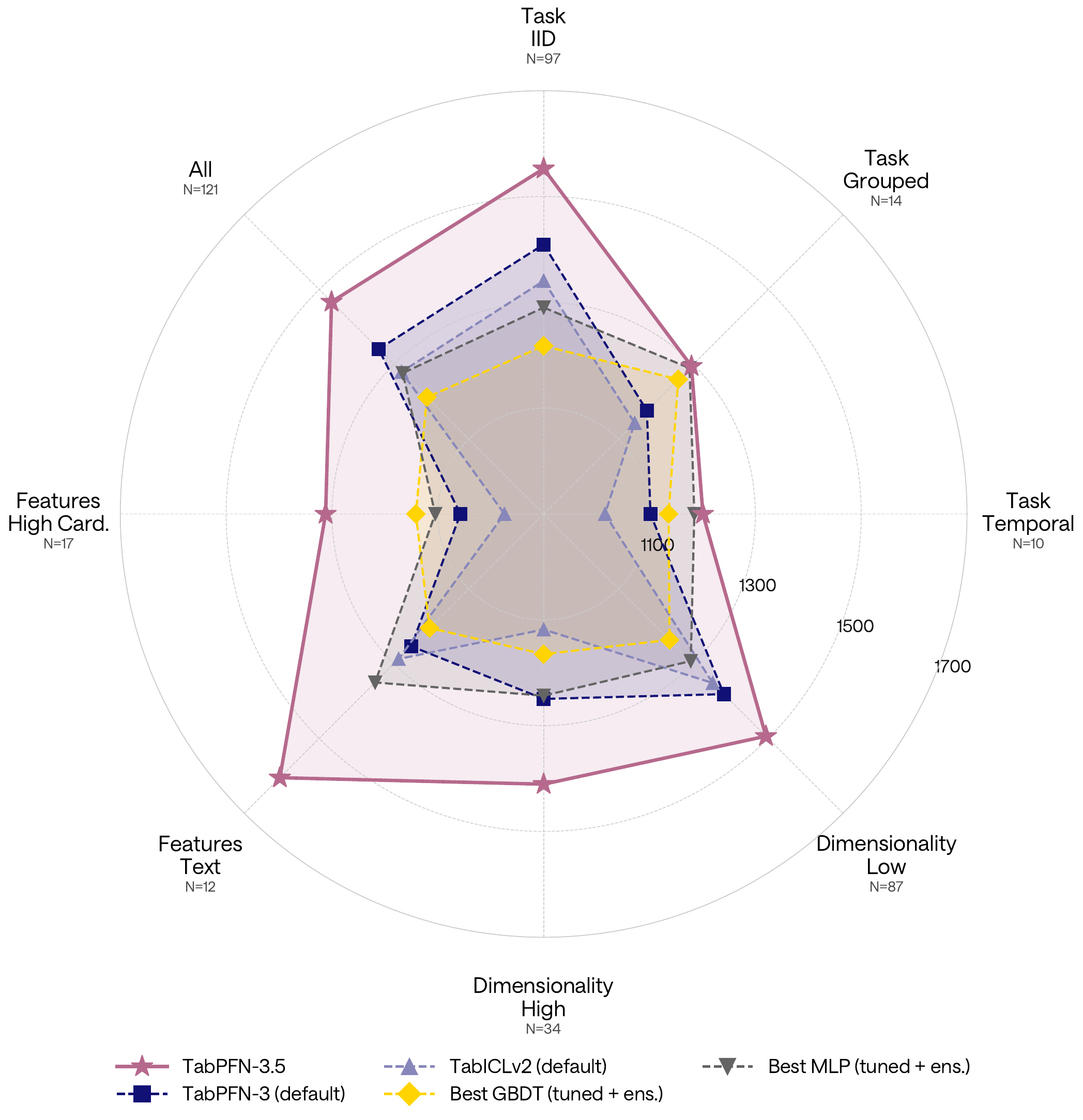}
        \caption{BeyondArena core without large tables.}
        \label{fig:ba-radar-nolarge}
    \end{subfigure}
    \caption{\textbf{BeyondArena Elo by data slice, radar view.} Same Elo fits as \Cref{fig:ba-core} (left) and \Cref{fig:ba-groupid-nolarge} (right): \ourmodel against \ourmodelthree, TabICLv2, and the strongest tuned and ensembled GBDT and MLP baselines. The group identifier is exposed to the TabPFN models on label-per-sample grouped datasets; see \Cref{sec:app:beyondarena:no_group_id}.}
    \label{fig:ba-radar}
\end{figure}

\Cref{fig:ba-radar} shows the Elo scores of \Cref{fig:ba-core} and \Cref{fig:ba-groupid-nolarge} as radar charts, with one axis per data slice. The numbers are the same as in the bar charts.

\subsubsection{Comparison to recent TFMs}
\label{sec:BA_TFMs}
\begin{figure}[t]
    \centering
    \includegraphics[width=\linewidth]{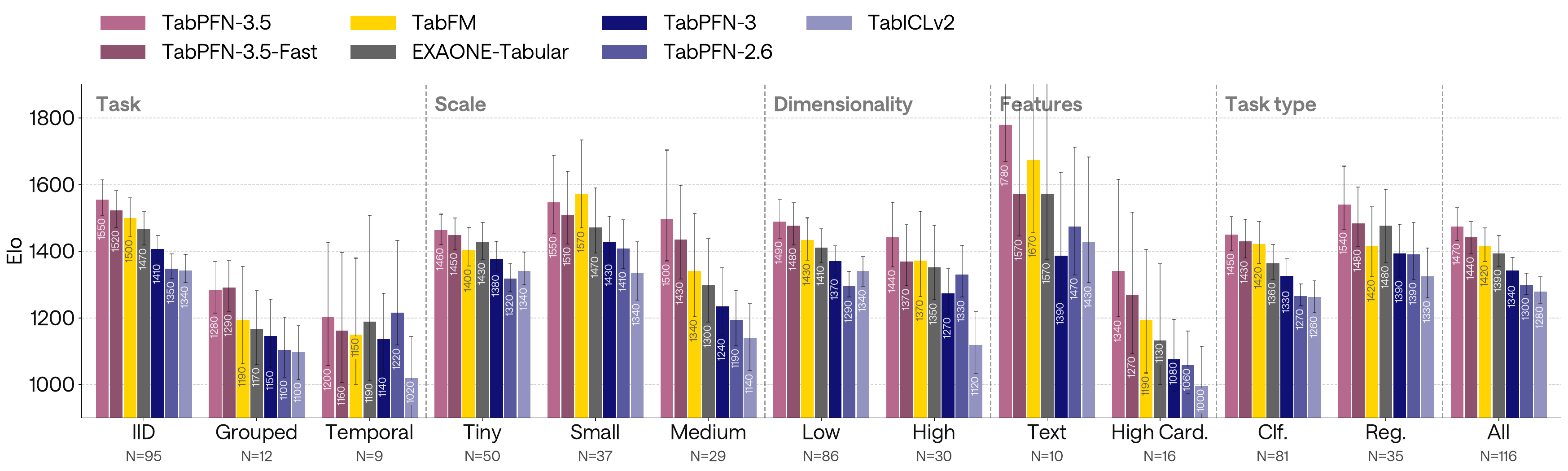}
    \caption{\textbf{Comparison with recent tabular foundation models on BeyondArena.} Elo scores on the 116 BeyondArena core datasets without large tables that \ourmodel, TabFM, and EXAONE-Tabular all support (see text), reported overall and by data slice. \ourmodel leads overall and across nearly every slice. The group identifier is exposed to the TabPFN models and to TabFM on label-per-sample grouped datasets, not to EXAONE-Tabular; see \Cref{sec:app:beyondarena:no_group_id}. The labels of the Elo scores are rounded to the nearest multiple of 10.}
    \label{fig:ba-3way}
\end{figure}

We additionally compare \ourmodel with TabFM~\citep{tabfm2026} and EXAONE-Tabular~\citep{eo2026exaonetabular} on the subset of BeyondArena that both models can run (\Cref{fig:ba-3way}). The subset is BeyondArena core without large tables (up to 100K rows), minus five datasets: \texttt{asp\_potassco\_classification}, \texttt{ljubljana\_primary\_tumor} and \texttt{micro\_mass} have more than ten classes, which TabFM does not support, and \texttt{sf\_permit\_time} (TabFM and EXAONE-Tabular) and \texttt{pva\_revenue\_prediction\_kddcup98} (TabFM) run out of memory on a 96\,GB GPU. This leaves 116 datasets, on which every method is scored on the same splits. \ourmodel leads both models overall and across every data slice except small datasets, where TabFM leads. Again, wide confidence intervals make some slice-level differences uncertain.

\subsubsection{Exposing the group identifier on label-per-sample grouped tasks}
  \label{sec:app:beyondarena:no_group_id}
   On grouped datasets, BeyondArena applies a shared preprocessing step which drops the group id before the table reaches a model.
   On \textit{label-per-group} tasks where every row of a group shares its label, BeyondArena also adds transductive feature aggregates within each group, which still encode which group a row belongs to.
   For \textit{label-per-sample} tasks, where labels vary within a group, no other grouped-specific preprocessing is done besides dropping the group id, so the group structure is not passed to the model.
   Not every method is affected in the same way: tuned and early-stopped baselines still exploit the group structure, because their validation splits are group-aware, whereas models which are fit on the full training set
  (TFMs) cannot exploit the group structure.

  To make the comparison fairer and closer to what practitioners can use in practice, our aim on label-per-sample tasks is to score every model in the configuration that serves it best. We therefore tried exposing the
  group key to every model family with several encodings (float, categorical, string), passing it raw or marking unseen test groups as missing. Simply passing the group id as a categorical variable helps every TabPFN model
  and we score \ourmodel{}, \ourmodelfast{}, \ourmodelthinking{} and \ourmodelthree{} with it. We score TabFM with the group id exposed as a float, unseen test groups marked as missing, which we find helps the most for this
  model. In our evaluations, no other model gains from passing the group id in any encoding, consistent with the results reported in the BeyondArena paper \citep{purucker2026beyond}, where only TabPFN-2.6 benefits from
  removing the grouped preprocessing for label-per-sample grouped datasets. For each model, we only evaluate whether passing the group id helps for the default model, not the tuned and ensembled version, so we cannot exclude that
  these versions would benefit from the group id (though they arguably already use the group-aware validation splits).

  On grouped data without large tables, when the group id is included as input, \ourmodel{} moves from 1240 to 1288 Elo (rank 6 to rank 1), \ourmodelthree{} from 1139 to 1176, and TabFM from 1167 to 1187. On grouped data
  including large tables, \ourmodel{} moves from 1187 to 1267 Elo (rank 9 to rank 4) and \ourmodelthree{} from 1078 to 1108. 

\subsection{Additional Results on STRABLE}
\label{sec:app-strable}

To evaluate performance on the 108 datasets of STRABLE (3 folds each), Elo ratings were computed jointly over a total of 200 methods. Specifically, we include only the 196 baselines from the original paper that successfully completed all 324 runs (108 datasets $\times$ 3 folds). On top of these, we added \ourmodel{} and \ourmodelfast{} with TF-IDF, \ourmodel{} with \textit{e5-small-v2}~\cite{wang2022text}, \ourmodelplus{}, and \ourmodelthinking{}. 

\begin{figure}[t]
    \centering
    \includegraphics[width=0.85\linewidth]{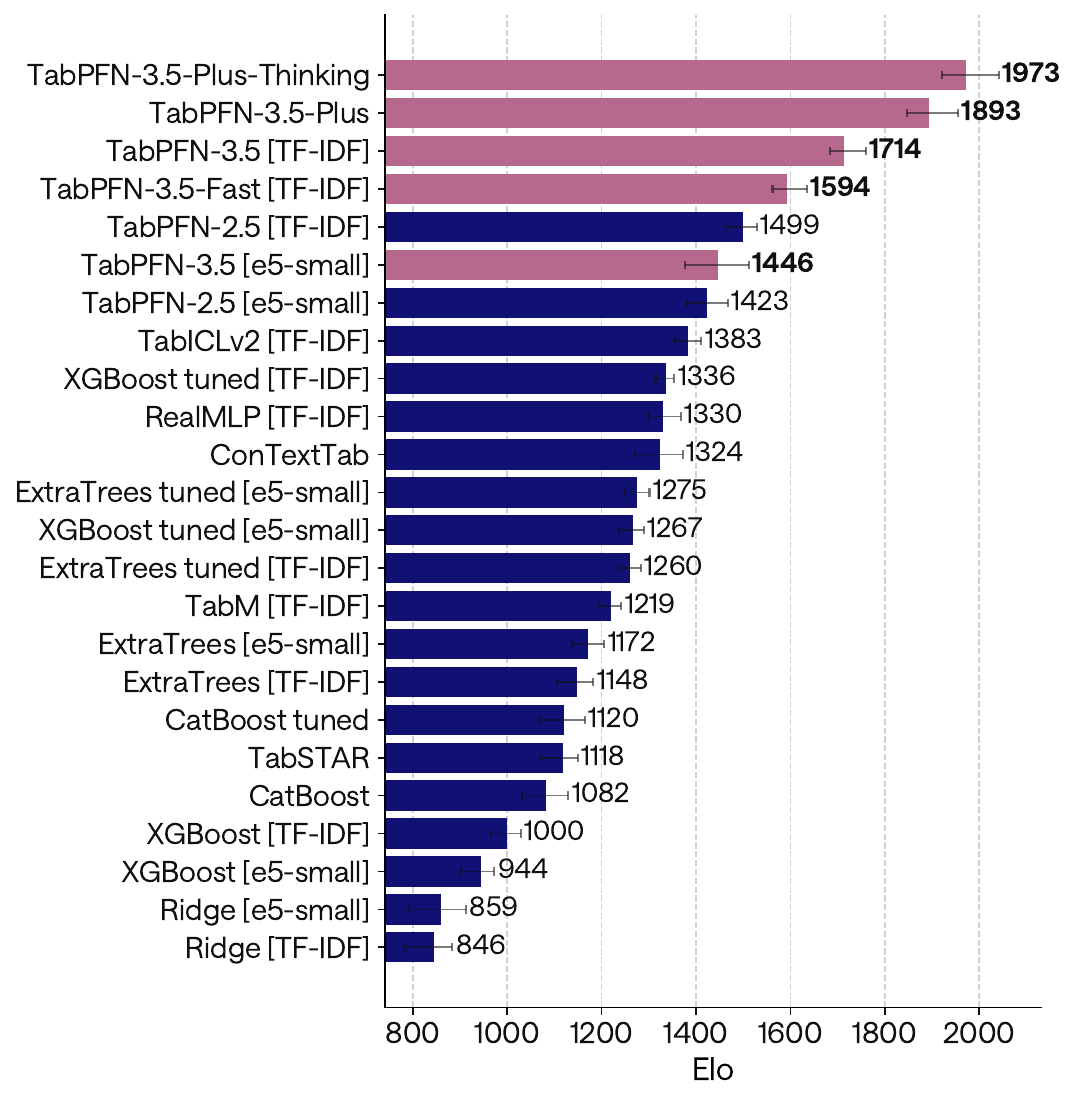}
    \caption{\textbf{Elo on STRABLE.} Elo ratings for a representative subset of methods, derived from a joint fit of all methods across all 108 datasets.}
    \label{fig:strable-elo-all}
\end{figure}

Figure~\ref{fig:strable-elo-all} displays a selected list of baselines, focusing specifically on models that are either end-to-end or utilize the stronger string encoding methods. Because the Elo ratings are derived from the comprehensive joint fit over all methods, the values remain identical to those presented in the main paper's figure, which simply filtered out a few of these pipelines for visual clarity.

Alongside our models, the 19 representative published pipelines shown include Ridge \cite{hoerl1970ridge}, ExtraTrees \cite{geurts2006extremely}, XGBoost \cite{chen2016xgboost}, RealMLP \cite{holzmuller2024realmlp}, TabM \cite{gorishniy2024tabm}, TabICLv2 \cite{qu2026tabiclv2}, and TabPFN-2.5 \cite{TabPFN-2.5}, all utilizing TF-IDF string encodings; four of these also feature variants with \textit{e5-small-v2} embeddings. Additionally, we include baselines that process raw text directly: CatBoost \cite{prokhorenkova2018catboost}, ConTextTab \cite{spinaci2026contexttab}, and TabSTAR \cite{arazi_tabstar_2025}.

When evaluating pipelines equipped with \textit{e5-small-v2} embeddings, the relative rankings remain consistent: \ourmodel{} narrowly outperforms TabPFN-2.5, albeit by a smaller margin than observed with TF-IDF. Crucially, \ourmodelthinking{} and \ourmodelplus{}, which process the raw text directly, retain the first and second positions on the overall leaderboard by a substantial margin. Finally, consistent with the official STRABLE benchmark findings, TF-IDF generally proves to be the stronger text representation. This aligns with the nature of the strings in the benchmark, which are mostly short entries rather than long-form free text.

\subsection{Additional Results on MulTaBench}
\label{sec:app-multabench}

\begin{figure}[t]
    \centering
    \includegraphics[width=0.85\linewidth]{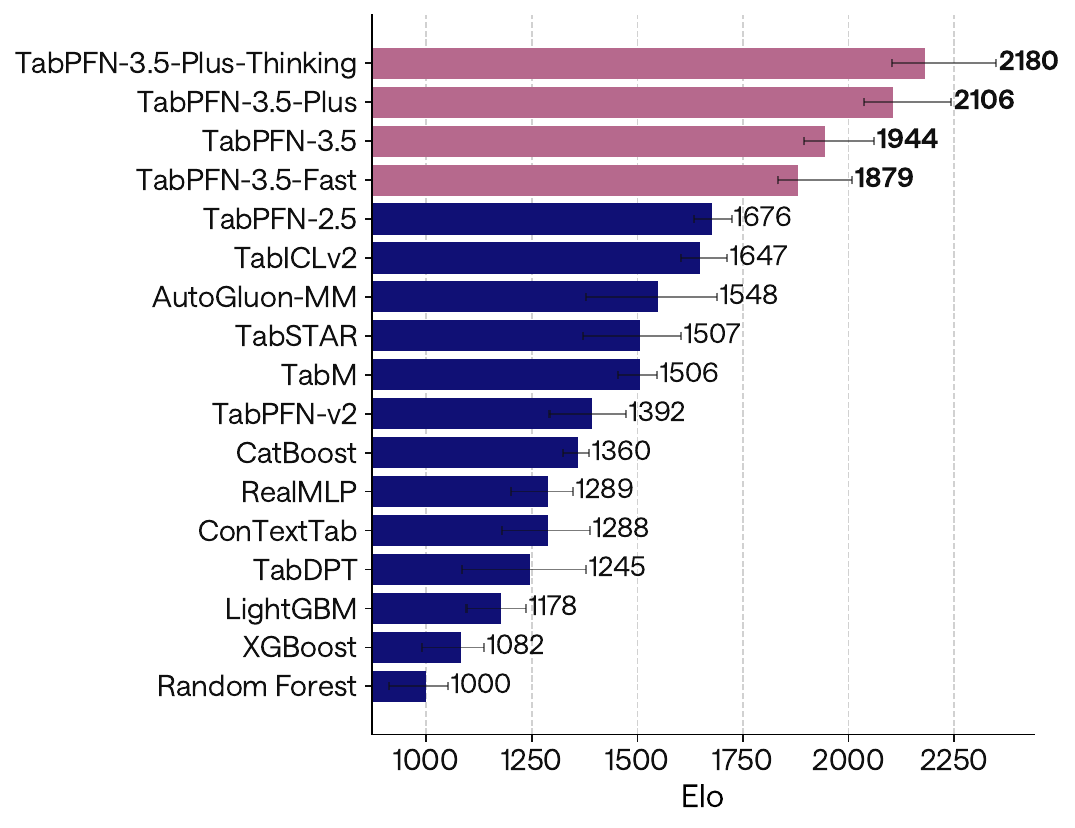}
    \caption{\textbf{Elo on MulTaBench text subset.} Elo ratings for all evaluated methods, computed jointly across the 20 text-focused datasets with 5 folds each. The \ourmodel{} family dominates the benchmark.}
    \label{fig:multabench-text}
\end{figure}

Figure~\ref{fig:multabench-text} displays the Elo rankings evaluated specifically on the subset of 20 text-focused datasets of MulTaBench (5 folds each). The Elo ratings are computed jointly over the 13 baseline models published in the official benchmark: TabICLv2 \cite{qu2026tabiclv2}, TabPFN-2.5 \cite{TabPFN-2.5}, TabPFN-v2 \cite{Hollmann2025tabpfnv2}, TabM \cite{gorishniy2024tabm}, TabDPT \cite{ma2024tabdpt}, CatBoost \cite{prokhorenkova2018catboost}, RealMLP \cite{holzmuller2024realmlp}, LightGBM \cite{lightgbm}, XGBoost \cite{chen2016xgboost}, and Random Forest \cite{breiman2001random}, all of which, like \ourmodel{}, utilize \textit{e5-small-v2}~\cite{wang2022text} embeddings, as well as AutoGluon-MM \cite{tang2024autogluon}, ConTextTab \cite{spinaci2026contexttab}, and TabSTAR \cite{arazi_tabstar_2025}, which process the raw text directly. In addition, we run our own models: \ourmodel{}, \ourmodelfast{}, \ourmodelplus{}, and \ourmodelthinking{}.

Focusing exclusively on these text datasets, the performance gap between our models and the baselines widens further compared to the full benchmark. \ourmodel{} and \ourmodelfast{}, despite relying on the same frozen \textit{e5-small-v2} features as the non-native baselines, extend their lead over prior tabular foundation models such as TabPFN-2.5 and TabICLv2. Crucially, \ourmodelplus{} and \ourmodelthinking{} achieve an even greater advantage, securing the first and second position by a substantial margin.

\subsection{TALENT}
\label{sec:talent}

TALENT \citep{talent_benchmark_jmlr} is a large collection of tabular datasets from many domains. Its base benchmark has 300 datasets spanning classification and regression tasks, and the authors later added extensions with larger and higher-dimensional datasets. Every dataset comes with a fixed split of 64\% training, 16\% validation and 20\% test rows. All methods are ranked on each dataset, and the average rank is reported with a 95\% bootstrap confidence interval over datasets. A method that has no result on some datasets gets those cells filled with the score of the KNN baseline; its bar is hatched and its label says how many cells were filled. The \ourmodelthree{} report \citep{grinsztajn2026tabpfn3technicalreport} describes these steps in detail.

We show two views of TALENT.

\begin{figure}[H]
    \centering
    \includegraphics[width=\linewidth]{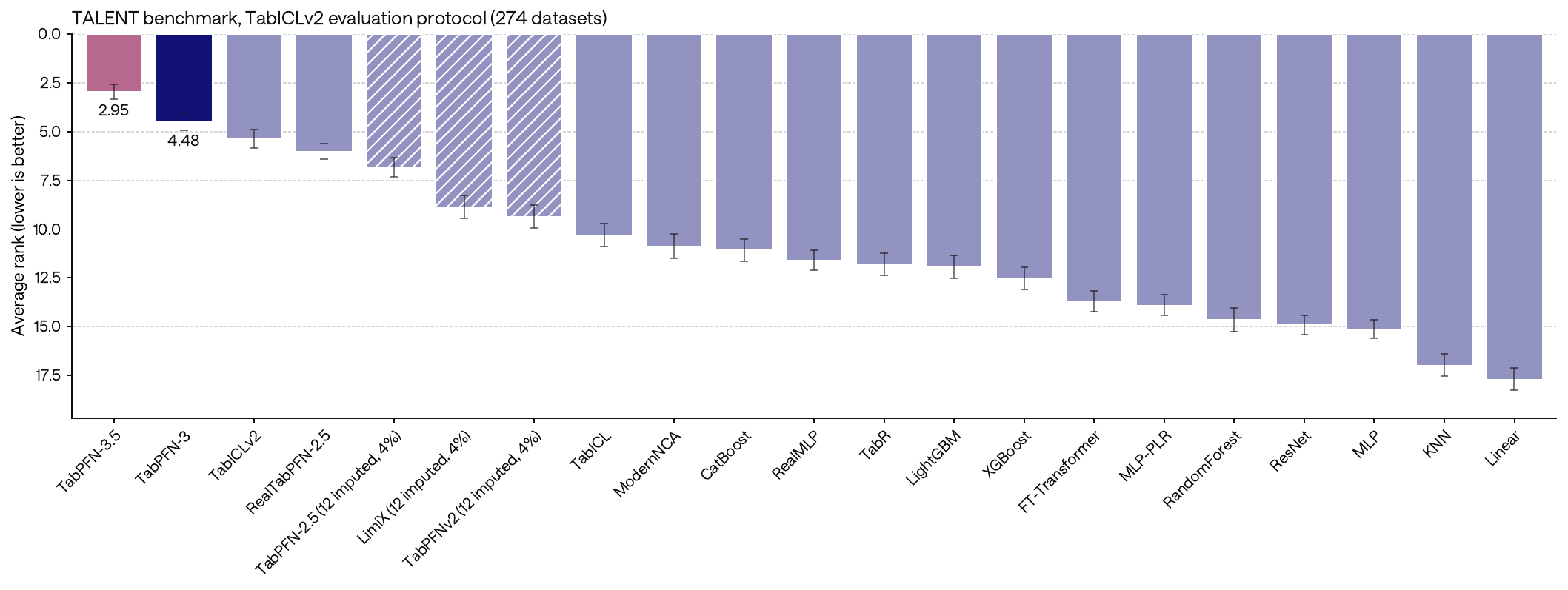}
    \caption{\textbf{Average rank on the TALENT benchmark, using the TabICLv2 evaluation protocol \citep{qu2026tabiclv2} (274 datasets).} Same datasets and baselines as the main TALENT figure of the \ourmodelthree{} report \citep{grinsztajn2026tabpfn3technicalreport}, with \ourmodel{} added. Lower is better; \ourmodel{} in plum, \ourmodelthree{} in dark blue. Error bars are 95\% bootstrap confidence intervals over datasets. Hatched bars have KNN-imputed cells, counted in the label.}
    \label{fig:talent-tabiclv2}
\end{figure}

\paragraph{TabICLv2 evaluation protocol.} \Cref{fig:talent-tabiclv2} follows the evaluation protocol of the TabICLv2 paper \citep{qu2026tabiclv2}: the 300 base datasets minus the 26 that were used during the development of TabPFN-2 and TabICLv2, 274 datasets in total, with the baselines the TabICLv2 authors ran. This is the same setup as the main TALENT figure of the \ourmodelthree{} report, with \ourmodel{} added. \ourmodel{} ranks first with an average rank of 2.95, ahead of \ourmodelthree{} at 4.48 and TabICLv2 at 5.35. The confidence intervals of \ourmodel{} and \ourmodelthree{} do not overlap.

\begin{figure}[H]
    \centering
    \includegraphics[width=\linewidth]{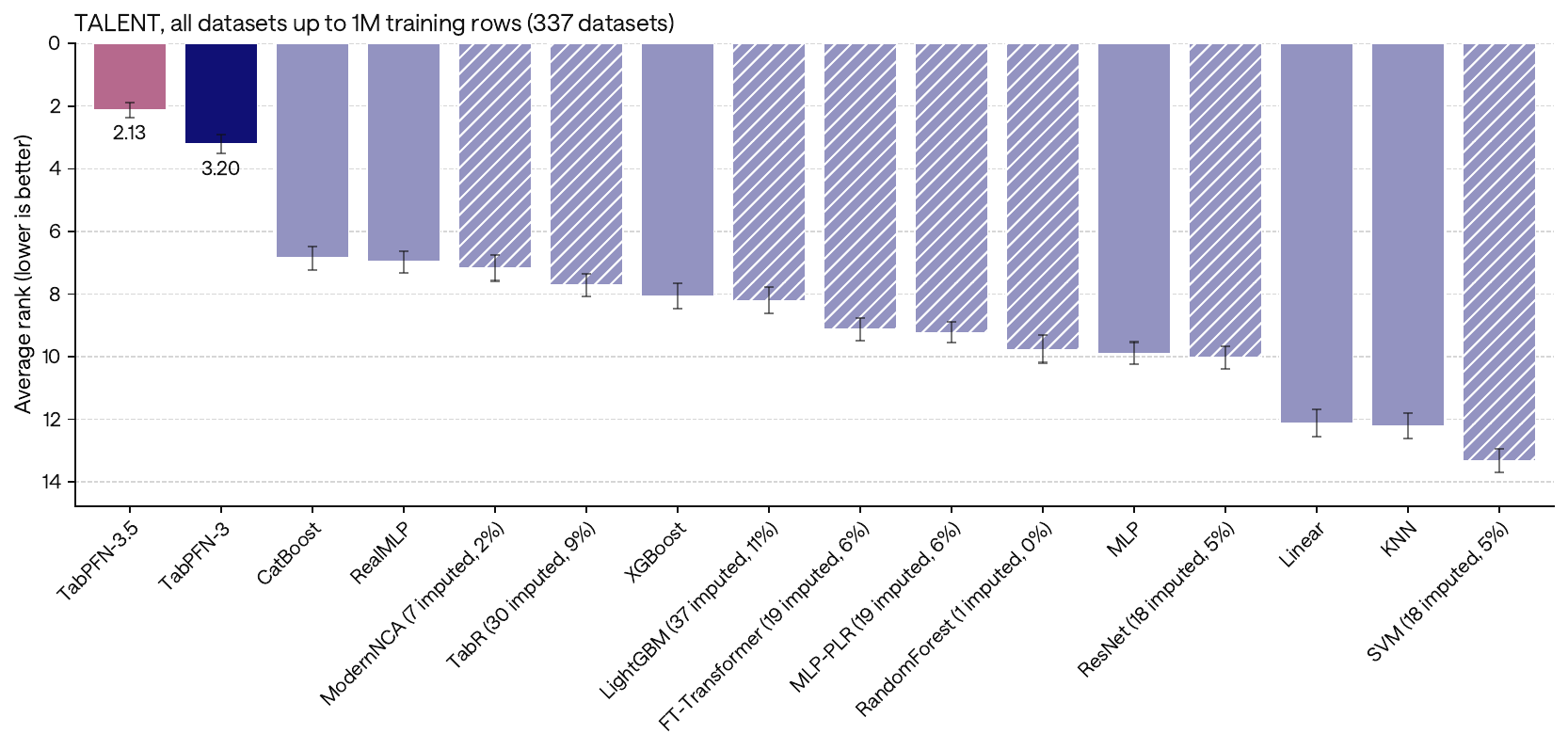}
    \caption{\textbf{Average rank on all TALENT datasets with up to one million training rows (337 datasets).} The 300 base datasets plus the large-data and high-dimensional extensions, against the baselines published by the TALENT authors \citep{talent_benchmark_jmlr}. Lower is better; \ourmodel{} in plum, \ourmodelthree{} in dark blue. Error bars are 95\% bootstrap confidence intervals over datasets. Hatched bars have KNN-imputed cells, counted in the label.}
    \label{fig:talent-full}
\end{figure}

\paragraph{All TALENT datasets up to one million training rows.} \Cref{fig:talent-full} uses every TALENT dataset with published baselines and at most one million training rows: the 300 base datasets plus the large-data and high-dimensional extensions, 337 datasets in total. The baselines are the results published by the TALENT authors. The TabICLv2 results used in \Cref{fig:talent-tabiclv2} do not cover the extensions, and the TALENT tables list earlier TabPFN versions only on a small part of the datasets, so this view compares against classical and deep tabular baselines only. \ourmodel{} again ranks first, with an average rank of 2.13 against 3.20 for \ourmodelthree{}.

\begin{figure}[H]
    \centering
    \includegraphics[width=\linewidth]{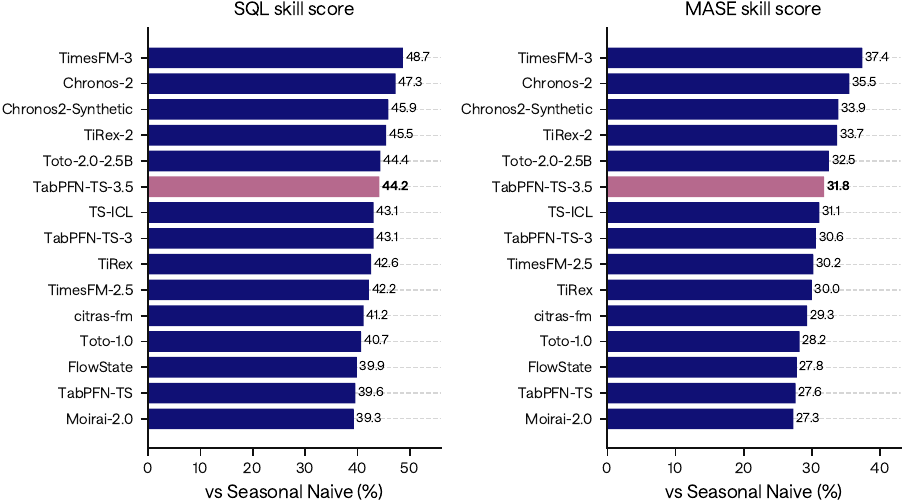}
    \caption{\textbf{fev-bench.} SQL (left) and MASE (right) skill score
    relative to Seasonal Naive on the 100 fev-bench tasks. Top 15 of 29 methods;
    each panel is sorted on its own metric. \textbf{TabPFN-TS-3.5 uses the default \ourmodel checkpoint}, whereas TabPFN-TS-3 uses a specialized time-series-specific checkpoint \citep{grinsztajn2026tabpfn3technicalreport}. Both are used inside our tabpfn-time-series wrapper.}
    \label{fig:fevbench-skill}
\end{figure}

\subsection{fev-bench}
\label{sec:fevbench}

fev-bench~\citep{shchur2025fev} is a forecasting benchmark of 100 tasks on
public series, scored as skill over Seasonal Naive with the SQL (quantile) and
MASE (point) metrics. We run \ourmodel{} through TabPFN-TS~\citep{hoo2024tabpfn_ts},
which casts each series as a regression over calendar and seasonal features,
with the inference config of the published TabPFN-TS-3 entry (context 32768, 12
AutoSeasonal periods), and score against the current upstream leaderboard field
of 29 methods. \Cref{fig:fevbench-skill} shows the result: \ourmodel{} places
sixth, 1.1 points above TabPFN-TS-3 at 2.4$\times$ its speed, behind TimesFM-3,
the two Chronos-2 variants, TiRex-2 and Toto-2.0. We emphasize that the \ourmodel{} checkpoint is the general
tabular model with no time-series finetuning, whereas TabPFN-TS-3 is a
time-series-specific checkpoint.

\subsection{ScoringBench}
\label{sec:scoringbench}

ScoringBench~\citep{landsgesell2026scoringbenchbenchmarkevaluatingtabular} evaluates the full predictive distribution of regression models with proper scoring rules rather than point error. It has 101 OpenML regression datasets, each subsampled to 3{,}000 rows and scored with 5-fold cross-validation. Its headline metric is the continuous ranked probability score (CRPS); it also reports interval scores, coverage, energy scores and density-based rules. All methods are ranked on each dataset, with folds averaged first, and the average rank over the 101 datasets is reported.

We ran \ourmodel{} and \ourmodelfast{} through the benchmark's own harness, with their default inference configuration and no tuning on the benchmark. The 51 baselines are the results the ScoringBench authors publish with the benchmark, including \ourmodelthree{}, TabICLv2 and EXAONE-Tabular; we did not re-run them. To check that our runs are comparable with these results, we re-ran the released \ourmodelthree{} under the same code: it reproduces its published CRPS to within $10^{-3}$ relative and lands on the same mean rank.

\begin{figure}[H]
    \centering
    \includegraphics[width=\linewidth]{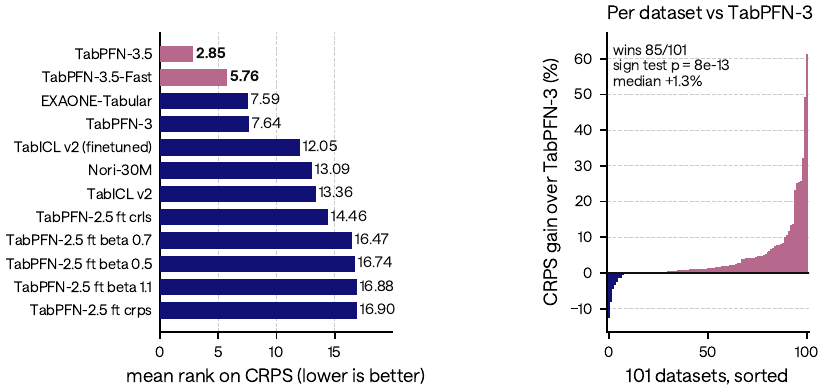}
    \caption{\textbf{ScoringBench.} Left: mean rank on CRPS across the 101 datasets for the top 12 of 53 methods; folds are averaged per dataset, methods ranked within each dataset, and ranks averaged. Lower is better; \ourmodel{} and \ourmodelfast{} in plum. Right: per-dataset relative CRPS improvement of \ourmodel{} over \ourmodelthree{}, sorted; positive bars are datasets \ourmodel{} wins (85 of 101, one-sided sign test $p = 8\times10^{-13}$, median $+1.3\%$). Published methods are ScoringBench's own committed results; \ourmodelthree{} re-run under the same code reproduces its published CRPS to within $10^{-3}$ relative.}
    \label{fig:scoringbench}
\end{figure}

\Cref{fig:scoringbench} shows the CRPS ranking. \ourmodel{} places first with a mean rank of 2.85 and \ourmodelfast{} second at 5.76; \ourmodel{} improves on \ourmodelthree{} (mean rank 7.64) on 85 of the 101 datasets.

\subsection{Details on the benchmark overview table}
\label{sec:app:family_winrates}

\Cref{tab:family_winrates_tiny_no_hindsight_named} summarises the seven tabular benchmarks of this report in
one table. This section states how each column is computed.

\paragraph{Family member.} Each benchmark is represented by the best \ourmodel{} family member evaluated
on it: \ourmodelthinking{} on TabArena, BeyondArena, STRABLE and MulTaBench (on STRABLE and MulTaBench
inside \ourmodelplus{}, which handles the raw text columns); TabPFN-Rel (3.5), the internal preview of the
TabPFN-Rel harness~\citep{hayler2026relarena} on \ourmodelplus{}, on \RelArena{}; and \ourmodel{} on TALENT and ScoringBench, where
no other family member was evaluated.

\paragraph{Rank.} The position of that member on the benchmark's headline ranking as reported in this
report: Elo on TabArena, BeyondArena, STRABLE, MulTaBench and \RelArena{}, mean rank on TALENT and
ScoringBench. The denominator counts every method on that ranking except the other \ourmodel{} family
members: the 88 entries of the TabArena field, every tuning variant counted separately (\Cref{sec:app:tabarena_leaderboard_tables});
the 28 pinned BeyondArena baselines; the 196 published STRABLE pipelines of the joint Elo fit
(\Cref{sec:app-strable}); the 13 published MulTaBench models; the 11 official \RelArena{} baselines; the
35 TALENT baselines published by the TALENT authors plus \ourmodelthree{}~\citep{grinsztajn2026tabpfn3technicalreport} (\Cref{sec:talent}); and the 51
published ScoringBench methods (\Cref{sec:scoringbench}).

\paragraph{Competitors.} The best other tabular foundation model and the best non-foundation model are the
highest-placed such methods on the same ranking. Systems that contain foundation models (AutoGluon~1.6~\citep{autogluon_tabular} on
TabArena, AutoGluon-MM~\citep{tang2024autogluon} on MulTaBench) are left out of both columns; \ourmodelthinking{} wins 79\% of
TabArena splits against AutoGluon~1.6 (noncommercial) and 82\% against AutoGluon~1.6 (extreme), and 77\%
of MulTaBench splits against AutoGluon-MM. On \RelArena{} the foundation-model column holds RT-PluRel~\citep{rt,plurel}, a
Relational Transformer fine-tuned per task from a PluRel-pretrained checkpoint, the closest to a foundation
model in that field, and the non-foundation column holds KurveRSC~\citep{madrigal2026kurversc}, a GraphReduce~\citep{madrigal_graphreduce} feature search with
CatBoost~\citep{prokhorenkova2018catboost}; both are system entries there. The 337-dataset TALENT view has no other foundation model, so its
TabICLv2~\citep{qu2026tabiclv2} entry is taken from the 274-dataset TabICLv2-protocol view (\Cref{fig:talent-tabiclv2}).

\paragraph{Win rate.} As on TabArena: every split (dataset and fold) is compared on its own, a tie counts
as half a win, the per-split results are averaged within a dataset and then over datasets, so every dataset
has the same weight. On \RelArena{} and TALENT a dataset has one split. TabArena has 51 datasets and 816
splits, BeyondArena 142 and 507 (the core view, group identifier exposed to the TabPFN models,
\Cref{sec:app:beyondarena:no_group_id}), STRABLE 108 and 324, MulTaBench 40 and 200, \RelArena{} 21 tasks,
TALENT 337 datasets and ScoringBench 101 datasets with 5 folds.

\end{document}